%% file: samplepaper.tex
\documentclass{article}

\usepackage{arxiv}

\usepackage[utf8]{inputenc}
\usepackage[T1]{fontenc}

\usepackage{amsmath, amssymb, amsfonts}
\usepackage{bm}

\usepackage{graphicx}
\usepackage{subcaption}
\usepackage{caption}
\usepackage{wrapfig}
\usepackage{adjustbox}

\usepackage{booktabs}
\usepackage{tabularx}
\usepackage{array}

\usepackage{algorithm}
\usepackage{algpseudocode}

\usepackage{url}
\usepackage{nicefrac}
\usepackage{microtype}
\usepackage{xcolor}
\usepackage{comment}
\usepackage{cancel}
\usepackage{placeins}

\usepackage[numbers]{natbib}

\usepackage{hyperref}
\hypersetup{
  colorlinks=true,
  linkcolor=blue,
  citecolor=blue,
  urlcolor=blue,
  filecolor=blue,
}

\newcommand{\maketitlesupplementary}{%
  \begin{center}
    {\LARGE Supplementary Material}\\[0.8em]
    {\large iStructTab: Structured Feature Sequencing for Multimodal Learning of Image and Tabular Data}
  \end{center}
}

\usepackage{graphicx}
\begin{document}
\makeatletter
\renewcommand{\refname}{References}
\makeatother
\title{iStructTab: Structured Feature Sequencing for Multimodal Learning of Image and Tabular Data
\thanks{
This paper has been accepted for presentation at the 28th International Conference on Pattern Recognition (ICPR 2026) in Lyon, France. 
Code: \url{https://github.com/zadid6pretam/iStructTab}. 
PyPI: \texttt{pip install istructtab}.
}
}
%
\author{
Al Zadid Sultan Bin Habib\textsuperscript{1},
Md Younus Ahamed\textsuperscript{1},
Prashnna Gyawali\textsuperscript{1},
Gianfranco Doretto\textsuperscript{2},
Donald A. Adjeroh\textsuperscript{1}
\\[0.6em]
\textsuperscript{1}Lane Department of Computer Science and Electrical Engineering,\\
West Virginia University, Morgantown, WV 26506, USA\\
\textsuperscript{2}Scientific Computing and Imaging Institute \& Department of Biomedical Informatics,\\
The University of Utah, Salt Lake City, UT 84112, USA\\[0.4em]
\texttt{\{ah00069,ma00087\}@mix.wvu.edu}\\
\texttt{\{prashnna.gyawali,donald.adjeroh\}@mail.wvu.edu},
\texttt{doretto@utah.edu}
}

\date{}
\maketitle
\bibpunct{[}{]}{,}{n}{}{,}
\begin{abstract}
Multimodal learning of images and tabular data is often impaired by ineffective representations, resulting in redundancy, dispersion, and generalization problems. To tackle this challenge, we introduce Graph-Enhanced Descriptor Sequencing (GEDS), a structured feature sequencing algorithm grounded in principles from the Column Permutation Problem (CPP). GEDS refines statistical descriptors of the features through similarity graph-based computations, systematically determining an effective feature sequencing. We incorporate GEDS within an order-aware efficient transformer framework, utilizing order-aware memory tokens that explicitly adhere to the derived feature sequencing via a dedicated loss function. Experimental results across multimodal benchmarks demonstrate that iStructTab effectively minimizes feature dispersion, improving predictive performance and robustness, and highlighting the significance of structured feature sequencing in multimodal learning.

\end{abstract}
\section{Introduction}
\label{intro}
Multimodal learning integrates heterogeneous sources such as images and tabular data, and is increasingly important in healthcare, remote sensing, and environmental modeling~\cite{bal}. Yet fusion remains difficult because tabular features are unordered, whereas images have strong spatial structure~\cite{ngiam,huang2020tabtransformer}. This structural mismatch can weaken cross-modal alignment, increase redundancy, and hurt generalization. Most tabular and multimodal methods address fusion through attention~\citep{saint}, contextual embeddings~\citep{huang2020tabtransformer}, ensemble regularization~\citep{grinsztajn2022tree}, contrastive pretraining, shared transformers, hierarchical attention, interaction modules, or temporal embeddings~\citep{hager2023best,du2024tip,stil,charms,miita,time}, but rarely consider feature sequencing. We revisit this overlooked factor by formulating multimodal feature ordering as a Column Permutation Problem (CPP)~\citep{cpp2,cpp3,lima2024delta,cpp4}, seeking a permutation that reduces inter-feature dispersion and improves structural coherence across modalities~\citep{cpp4,vinyals2015pointer}.

Feature ordering has a long history in pattern recognition and is central to Incremental Attribute Learning (IAL), where sequentially arriving features must be ranked before training~\citep{f1}. Unlike order-invariant set models~\citep{zaheer2017deep}, column order can affect redundancy, dependency capture, and prediction. Prior studies show that Fisher, correlation, and entropy-based rankings reduce interference and error over unordered baselines~\citep{f2,f4}, motivating task-aware sequencing~\citep{f3,f7}. Recent methods address permutation brittleness by enforcing order-agnostic representations~\citep{ack1,ack2}; TabICL~\citep{tabicl} averages predictions over multiple column permutations, while COPER~\citep{coper} uses a permutation-based correlation objective for image-table clustering. We instead treat feature ordering as a Column Permutation Problem (CPP): given \(X\in\mathbb{R}^{N\times m}\), find \(\pi\in S_m\) minimizing a task-relevant similarity or dispersion cost, where \(N\) and \(m\) denote samples and features. Since CPP is NP-hard~\citep{johnson1979computers}, most fusion methods avoid it and treat feature order as arbitrary. We address this gap with iStructTab, combining Graph-Enhanced Descriptor Sequencing (GEDS) with a sequence-aware efficient transformer~\citep{vaswani2017attention}. GEDS builds feature descriptors, refines them through graph similarity propagation~\citep{kipf2016semi}, and produces \(\pi_{\mathrm{GEDS}}\), an optimized multimodal feature sequence reflecting both statistical distinctiveness and relational structure.

We integrate this sequence into an Order-Aware Efficient Transformer with Memory Augmentation (OEMT), which processes features according to \( \pi_{\mathrm{GEDS}} \), preserves their positional structure, and incorporates learnable memory tokens for global reasoning \cite{Rae2020Compressive}. A dedicated sequencing loss is applied to align learned representations with the target permutation, encouraging consistent feature semantics and interpretability. Our key contributions are as follows:

\begin{itemize}
    \item We formulate multimodal fusion as a CPP and show how it impacts representation learning in hybrid tabular-image settings.
    \item We introduce GEDS, a simple yet effective descriptor sequencing algorithm based on graph-enhanced similarity propagation.
    \item We design OEMT that explicitly enforces feature sequencing through learnable memory tokens and sequencing-aware loss function.
    \item Through extensive experiments, we demonstrate that our GEDS-OEMT framework significantly improves classification accuracy and robustness across diverse multimodal benchmarks.
\end{itemize}

\section{Related Work}
\label{sec:formatting}
\textbf{A: Multimodal Learning with Tabular and Image Data.}
Recent image-tabular methods learn joint representations through contrastive alignment, transformer-based fusion, or attention-based interaction. MMCL~\citep{hager2023best} aligns modalities with contrastive pretraining and LaaF label tokens, while TIP~\citep{du2024tip} combines masked tabular reconstruction, cross-modal contrastive learning, and image-tabular matching in a unified transformer. DAFT~\citep{wolf2022daft} uses dense attentive fusion, and simpler CF/MF/IF baselines~\citep{du2024tip} merge embeddings by concatenation, pooling, or pairwise interaction. XTab~\citep{zhu2023xtab} benchmarks fusion strategies with ResNet features and linear/MLP heads. More recent systems include STiL~\citep{stil}, MIITA~\citep{miita}, CHARMS~\citep{charms}, and TIME~\citep{time}, which respectively use hierarchical attention, progressive tabular-to-visual integration, cross-hierarchical attention, and temporal embeddings. However, these methods largely treat fused descriptors as unordered vectors. iStructTab instead learns an explicit feature sequence, reducing dispersion and improving cross-modal alignment via structural priors.

\textbf{B: Deep Learning for Tabular Data.}
Deep tabular models increasingly rely on attention, embeddings, and self-supervision. TabNet~\citep{arik2021tabnet} selects features via sequential attentive masks; TabTransformer~\citep{huang2020tabtransformer} contextualizes categorical embeddings; and SAINT~\citep{saint} combines row/column attention with contrastive pretraining. FT-Transformer~\citep{b19}, VIME~\citep{yoon2020vime}, SCARF~\citep{bahri2021scarf}, and TANGOS~\citep{jeffarestangos} further use transformer backbones, denoising or contrastive objectives, and importance-aware attention regularization. TabPFN~\citep{hollmann2022tabpfn} and TabPFN V2~\citep{hollmann2025accurate} meta-train probabilistic function networks for rapid adaptation to small tabular tasks. However, these methods usually assume a fixed column order rather than optimizing feature permutations. TabSeq~\citep{habib2024tabseq} shows that explicit feature sequencing can affect both convergence and accuracy.

\textbf{C: Column Permutation and Combinatorial Optimization.}
Column permutation problems (CPPs) seek item orderings that minimize dispersion-style objectives. Lima et al.~\citep{lima2024delta} formalize this with a \(\Delta\)-evaluation function penalizing pairwise variance or dissimilarity. CPPs are closely related to permutation-based NP-hard problems such as TSP. Pointer Networks~\citep{vinyals2015pointer} learn such permutations through attention-based decoding, while GPN~\citep{yang2022graph} and PGN~\citep{velivckovic2020pointer} incorporate graph structure and adaptive routing. These works motivate our formulation of multimodal feature sequencing as a learnable CPP.

\textbf{D: Feature Sequencing and Structure-Aware Representation Learning.}
Feature order can strongly affect permutation-sensitive architectures~\citep{ack1, ack2, tabicl}. TabSeq~\citep{habib2024tabseq} learns task-driven feature sequences via clustering and feeds them to a transformer autoencoder, improving tabular prediction. Mambular~\citep{thielmann2024mambular} similarly shows that sequential column processing with Mamba layers is sensitive to positional arrangement. More broadly, structure-aware transformers~\citep{chen2022structure} and graph attention methods~\citep{dwivedi2021graph} inject topological information into attention patterns. Unlike MLP-style models that treat coordinates as largely exchangeable~\citep{zaheer2017deep}, these architectures depend on feature ordering or structure.

In iStructTab, we extend these ideas by using GEDS to explicitly model column-wise dispersion and multimodal similarity, producing a task-aware feature order prior to fusion. By enforcing this learned sequencing before the OEMT backbone, we inject a strong structural prior that is crucial for stable, robust multimodal deep learning.
\section{Methodology}
\label{method}
\begin{figure*}[t]
    \centering
    \includegraphics[width=\linewidth]{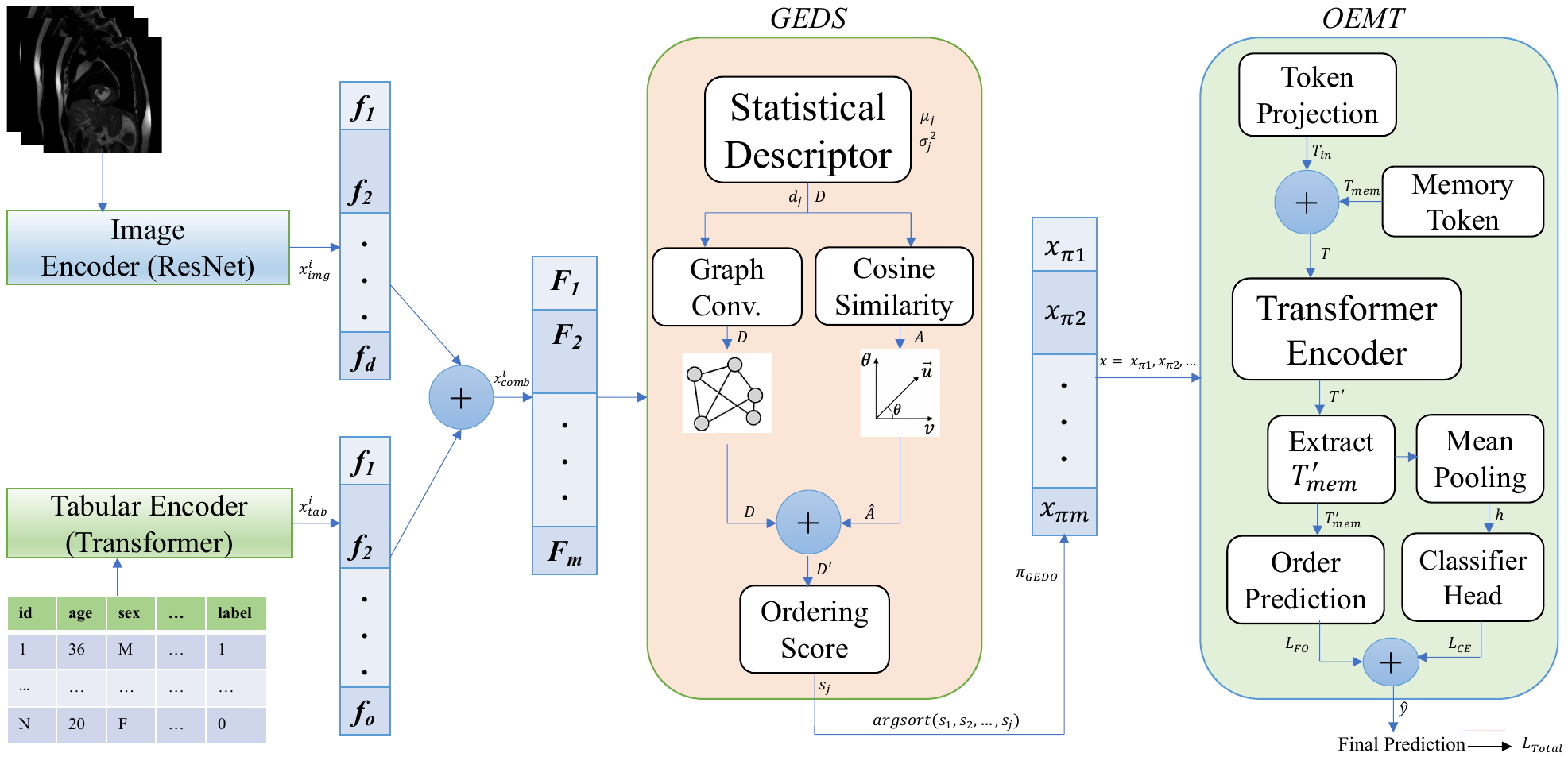}
    \caption{Overview of the proposed iStructTab architecture. We first extract modality-specific representations from image and tabular encoders. These are concatenated and passed to the GEDS block, which computes statistical descriptors and similarity-aware structural relationships via graph convolution and cosine similarity. A combined ordering score is used to generate $\pi_{\text{GEDS}}$ for feature sequencing. The reordered input is processed by the OEMT backbone to produce the final prediction.}
    \label{fig:gedo_oemt_architecture}
\end{figure*}
Figure \ref{fig:gedo_oemt_architecture} provides an overview of the proposed approach, iStructTab. We address the fusion of multimodal data (images and tabular features) by casting feature sequencing as a Column Permutation Problem (CPP). 
Here, given a dataset, we represent it as  
$X \in \mathbb{R}^{N \times m}
$.
This captures the input data matrix consisting of \(N\) samples and \(m\) features, where each row corresponds to a sample and each column corresponds to a feature.
Our objective is to determine a permutation \(\pi \in S_m\) that minimizes a dispersion cost:
\begin{equation}
\min_{\pi \in S_m} D(\pi) = \sum_{i<j} w_{ij}\ |\pi(i) - \pi(j)| \label{eq:CPP}
\end{equation}
\noindent Here, \(S_m\) denotes the symmetric group on \(m\) elements, i.e., the set of all permutations of \(m\) features. Equation \eqref{eq:CPP} defines the cost function \(D(\pi)\) which aggregates the pairwise dispersion between features, with the weights \(w_{ij}\) quantifying the dissimilarity between the $i$-th and $j$-th features (based on a given divergence metric or measure, for instance, variance differences).
\subsection{Graph-Enhanced Descriptor Sequencing (GEDS)}
For images, an encoder, or a deep model, (e.g., CNN  using ResNet~\citep{he2016deep} following TIP~\citep{du2024tip} and STiL~\citep{stil}) extracts compact feature vectors. Thus, here each image can be represented as a \(d\)-dimensional vector:  
$\mathbf{x}_{\text{img}}^i \in \mathbb{R}^d
$.
For tabular data, a Transformer encoder \(f_{\text{Transformer}}\) is applied to obtain encoded features and the tabular data is encoded into an \(p\)-dimensional feature vector, 
$\mathbf{x}_{\text{tab}}^i \in \mathbb{R}^{p}
$. 
These are then concatenated to form a unified feature vector as shown in  Eq. \ref{eq5}. The equation  describes the fusion process where the transformer-encoded tabular features and the CNN-extracted image features are combined to form a single vector of dimension \(p+d\). Stacking these for all \(N\) samples yields the unified feature matrix in Eq. \ref{eq6} where \(F\) contains the fused features for all samples; each column corresponds to one feature derived from the concatenation.

\begin{equation}
\label{eq5}
\mathbf{x}_{\text{comb}}^i = \text{concat}(\mathbf{x}_{\text{tab}}^i, \mathbf{x}_{\text{img}}^i) \in \mathbb{R}^{p+d}
\end{equation}

\begin{equation}
\label{eq6}
F \in \mathbb{R}^{N \times (p+d)}
\end{equation}

For each feature (i.e., each column of \(F\)), a simple statistical descriptor is computed. Each feature \(f_j\) is characterized by a descriptor. Eq. 
\ref{eq:descriptor} 
defines the descriptor \(d_j\) that comprises the mean and variance of feature \(f_j\).
\begin{equation}
d_j = \begin{bmatrix} \mu_j \\ \sigma_j^2 \end{bmatrix} \in \mathbb{R}^2 \label{eq:descriptor}
\end{equation}

where,
\begin{equation}
\mu_j = \frac{1}{N}\sum_{i=1}^{N} F_{ij}, \quad \sigma_j^2 = \frac{1}{N}\sum_{i=1}^{N} \left(F_{ij}-\mu_j\right)^2 \label{eq:stats}
\end{equation}

Eq. 
\ref{eq:stats} details how the mean \(\mu_j\) and variance \(\sigma_j^2\) are computed for each feature over all \(N\) samples. Let \( D \in \mathbb{R}^{m \times 2} \) be the matrix of descriptors, where $m=p+d$ is the total number of features. Each descriptor is normalized as 
shown in Eq. \ref{eq:normalize}, and each descriptor is scaled to have unit norm, which ensures that subsequent similarity calculations are not dominated by the magnitude of 
some  descriptors. The cosine similarity matrix is then computed as in  Eq. \ref{eq:cosine}. This computes the cosine similarity between every pair of normalized descriptors, thus quantifying the similarity between features and followed by symmetric normalization. 
The similarity matrix is first augmented with self-connections (via \(A+I\)) where $I$ is identity matrix, and $A$ is from Eq. \ref{eq:cosine}. The results is  then symmetrically normalized using the degree matrix \(D_{\mathrm{deg}}\). See  Eq.
\ref{eq:normAdj}.
A graph convolution refines the descriptors and in Eq. 
\ref{eq:gcn}, the refined descriptors \(D'\) are obtained by applying a graph convolution operation on \(D\), where \(W\) is a learnable weight matrix and \(\sigma\) is a non-linear activation function (e.g., ReLU).
\begin{equation}
d_j \leftarrow \frac{d_j}{\|d_j\|} \label{eq:normalize}
\end{equation}

\begin{equation}
A = D\,D^\top \label{eq:cosine}
\end{equation}

\begin{equation}
\begin{split}
\hat{A} &= D_{\mathrm{deg}}^{-1/2}(A+I)D_{\mathrm{deg}}^{-1/2} \\
\text{with } D_{\mathrm{deg}}(j,j) &= \sum_{k} (A+I)_{jk}
\end{split}
\label{eq:normAdj}
\end{equation}

\begin{equation}
D' = \sigma\bigl(\hat{A}\,D\,W\bigr) \label{eq:gcn}
\end{equation}

\begin{equation}
s_j = \|d'_j\|_2 \label{eq:score}
\end{equation}

An ordering score is then defined as in Eq. \eqref{eq:score}, which computes the Euclidean norm of the refined descriptor for each feature. This norm serves as a score reflecting the importance or distinctiveness of the feature. 
The initial sequence is given by Eq. \eqref{eq:gedo_order} which sorts the features in ascending order based on their scores, resulting in the permutation \(\pi_{\mathrm{GEDS}}\).
\begin{equation}
\pi_{\mathrm{GEDS}} = \operatorname{argsort}(s_1, s_2, \ldots, s_m) \label{eq:gedo_order}
\end{equation}

Algorithm \ref{alg:alg1} shows the detailed procedure.

\begin{algorithm}[t]
\caption{Graph-Enhanced Descriptor Sequencing (GEDS)}
\label{alg:alg1}
\begin{algorithmic}[1]
\Require \(X_{\rm tab}\in\mathbb{R}^{N\times p}\), \(\{X_{\rm img}^i\}_{i=1}^N\), encoders \(f_T,f_R\)
\State \(Z_{\rm tab}\gets f_T(X_{\rm tab}),\quad Z_{\rm img}\gets f_R(\{X_{\rm img}^i\}_{i=1}^N)\)
\State \(F\gets [Z_{\rm tab},Z_{\rm img}]\in\mathbb{R}^{N\times m}\), where \(m=p+d\)
\State \(d_j\gets [\mu(F_{:j}),\operatorname{var}(F_{:j})]^\top,\quad j=1,\ldots,m\)
\State \(D\gets [d_1,\ldots,d_m]^\top\in\mathbb{R}^{m\times 2}\), \quad \(D\gets \operatorname{row\_norm}(D)\)
\State \(A\gets DD^\top+I,\quad D_{\rm deg}\gets \operatorname{diag}(A\mathbf{1})\)
\State \(\hat{A}\gets D_{\rm deg}^{-1/2}AD_{\rm deg}^{-1/2}\)
\State \(D'\gets \sigma(\hat{A}DW)\)
\State \(s_j\gets \|D'_{j:}\|_2,\quad j=1,\ldots,m\)
\State \(\pi_{\rm GEDS}\gets \operatorname{argsort}(s_1,\ldots,s_m)\)
\State \Return \(\pi_{\rm GEDS}\)
\end{algorithmic}
\end{algorithm}

\subsection{Order-Aware Efficient Transformer with Memory Augmentation (OEMT)}
We introduce the Order-Aware Efficient Transformer with Memory Augmentation (OEMT), which not only learns powerful representations from structured inputs but also preserves and enforces a prescribed feature sequencing at every stage of its computation. We equip our backbone with two components: (i) an order-aware pooling that compresses $m$ ordered features into $k \!\ll\! m$ summary tokens, and (ii) a lightweight Linformer-based encoder operating on the pooled tokens plus $M$ memory tokens. A joint classification and feature sequencing loss then enforces the desired column permutation. Given an unordered feature set $\mathcal{F}=\{f_1,\dots,f_m\}$ and its optimal permutation $\pi$, we rearrange as per Eq.~\ref{eqo1} and project into $d_t$-dimensional tokens in Eq.~\ref{o2}, where $\mathbf{W}_p\in\mathbb{R}^{d_t\times m}$ and $\mathbf{b}_p\in\mathbb{R}^{d_t}$. In this work, the optimal permutation $\pi$ is given by $\pi_{\mathrm{GEDS}}$ from Algorithm~1. To reduce sequence length from $m$ to $k$, we learn a pooling matrix using an order-aware pooling in $\mathbf{W}_{\mathrm{pool}}\in\mathbb{R}^{k\times d_t}$ and compute $\mathbf{P}$ in Eq.~\ref{o3}. Since $\mathbf{T}_{\mathrm{in}}$ (Eq.~\ref{o4}) is ordered by $\pi$, the pooled tokens $\widetilde{\mathbf{T}}_{\mathrm{in}}$ preserve the exact permutation. We then prepend $M$ learnable memory tokens and form $\mathbf{T}$ in Eq.~\ref{o5}, where $\mathbf{T}_{\mathrm{mem}}\in\mathbb{R}^{M\times d_t}$. The resulting sequence $\mathbf{T}$ is processed by a Linformer encoder~\citep{wang2020linformer} as in Eq.~\ref{o6}. From the encoded tokens, we extract the memory outputs and compute $\mathbf{h}$ (Eq.~\ref{o7}) and the sequencing score vector \(\mathbf{s}'\) (Eq.~\ref{o8}). We train with a combined loss $\mathcal{L}_{\mathrm{total}}$ (Eq.~\ref{o9}) to obtain the final prediction where $\lambda_{\mathrm{FS}}$ is a weighting factor for the feature sequencing loss. 
\newline
\begin{equation}
\label{eqo1}
\mathbf{x} = [\,x_{\pi_1},\,x_{\pi_2},\,\dots,\,x_{\pi_m}\,]\;\in\;\mathbb{R}^m
\end{equation}
\begin{equation}
\label{o2}
\mathbf{T}_{\mathrm{in}}
= \mathrm{ReLU}\bigl(\mathbf{W}_p\,\mathbf{x} + \mathbf{b}_p\bigr)
\;\in\;\mathbb{R}^{m\times d_t}
\end{equation}
\begin{equation}
\label{o3}
\mathbf{P}
= \mathrm{softmax}\bigl(\mathbf{W}_{\mathrm{pool}}\,\mathbf{T}_{\mathrm{in}}^\top\bigr)
\;\in\;\mathbb{R}^{k\times m}
\end{equation}
\begin{equation}
\label{o4}
\widetilde{\mathbf{T}}_{\mathrm{in}}
= \mathbf{P}\,\mathbf{T}_{\mathrm{in}}
\;\in\;\mathbb{R}^{k\times d_t}
\end{equation}
\begin{equation}
\label{o5}
\mathbf{T}
= \bigl[\widetilde{\mathbf{T}}_{\mathrm{in}};\,\mathbf{T}_{\mathrm{mem}}\bigr]
\;\in\;\mathbb{R}^{(k+M)\times d_t}
\end{equation}
\begin{equation}
\label{o6}
\mathbf{T}'
= \mathrm{Encoder}\bigl(\mathbf{T}\bigr)
\;\in\;\mathbb{R}^{(k+M)\times d_t}
\end{equation}
\begin{equation}
\label{o7}
\mathbf{h} = \frac{1}{M}\sum_{i=1}^{M}\mathbf{T}'_{k+i} \in \mathbb{R}^{d_t}, 
\quad
\hat{y} = \mathrm{softmax}\bigl(\mathbf{W}_c\,\mathbf{h} + \mathbf{b}_c\bigr)
\end{equation}
\begin{equation}
\label{o8}
\mathbf{s}' = \mathbf{W}_o\,\mathrm{vec}\bigl(\mathbf{T}'_{k+1:k+M}\bigr) + \mathbf{b}_o \in \mathbb{R}^m,
\quad
\beta_{\pi(i)} = 1 - \frac{i-1}{m-1}
\end{equation}
\begin{equation}
\label{o9}
\mathcal{L}_{\mathrm{total}}
= \mathcal{L}_{\mathrm{CE}}(y,\hat{y})
+ \lambda_{\mathrm{FS}}
\bigl\|\mathbf{s}' - \boldsymbol{\beta}\bigr\|_2^2
\end{equation}
\centerline{
where $\mathbf{W}_c,\mathbf{W}_o$ and $\mathbf{b}_c$ are defined  in the paragraph below.}\\
\\
The matrix $\mathbf{W}_p\in\mathbb{R}^{d_t\times m}$ in Eq.~\ref{o2} is a learnable projection matrix that transforms the input features $\mathbf{x}$ into a $d_t$-dimensional token embedding space. The associated vector $\mathbf{b}_p\in\mathbb{R}^{d_t}$ is a bias vector for this projection operation. The matrix $\mathbf{W}_{\mathrm{pool}}\in\mathbb{R}^{k\times d_t}$ in Eq.~\ref{o3} 
is a learnable pooling matrix that reduces the number of tokens from $m$ (the original length of input tokens) to $k\ll m$ summary tokens. This pooling step enforces the desired feature sequencing structure by generating permutation-aware compressed tokens $\widetilde{\mathbf{T}}_{\mathrm{in}}$. The matrix $\mathbf{T}_{\mathrm{mem}}\in\mathbb{R}^{M\times d_t}$ in Eq.~\ref{o5} comprises $M$ learnable memory tokens appended to the pooled feature tokens. These tokens provide a global contextual representation that assists the model in capturing interactions across ordered features. In Eq.~\ref{o7}, $\mathbf{W}_c\in\mathbb{R}^{C\times d_t}$ and bias vector $\mathbf{b}_c\in\mathbb{R}^{C}$ are parameters of the final classification layer, where $C$ represents the total number of classes. The vector $\mathbf{h}\in\mathbb{R}^{d_t}$ is the average pooled memory token representation, subsequently mapped to class probabilities $\hat{y}$. On the other hand, $\mathbf{W}_o\in\mathbb{R}^{m\times(M d_t)}$ and the bias vector $\mathbf{b}_o\in\mathbb{R}^{m}$ (Eq.~\ref{o8}) constitute parameters used to predict the sequencing score vector $\mathbf{s}'\in\mathbb{R}^{m}$ from the memory-encoded token embeddings reshaped into a vector. The vector $\boldsymbol{\beta}\in\mathbb{R}^{m}$ explicitly defines the target sequence, with elements $\beta_{\pi(i)} = 1 - \frac{i-1}{m-1}$ linearly ranking features from most to least important according to the optimal permutation $\pi$. The combined loss function $\mathcal{L}_{\mathrm{total}}$ integrates the cross-entropy classification loss $\mathcal{L}_{\mathrm{CE}}(y, \hat{y})$ with a sequencing regularization term scaled by $\lambda_{\mathrm{FS}}$ in Eq.~\ref{o9}. This term penalizes deviations between the predicted ordering scores $\mathbf{s}'$ and the target sequencing vector $\boldsymbol{\beta}$. The overall loss thus simultaneously optimizes for accurate classification and faithful feature sequencing. Algorithm  \ref{alg:alg2} shows the end-to-end training procedure for {iStructTab}.
\begin{algorithm}[t]
\caption{iStructTab Training with GEDS + OEMT}
\label{alg:alg2}
\begin{algorithmic}[1]
\Require Minibatches \(\{(X_{\rm tab}^{(b)},X_{\rm img}^{(b)},y^{(b)})\}\); encoders \(f_T,f_R\); GEDS; OEMT; weight \(\lambda_{\rm FS}\)
\For{each epoch}
  \For{each minibatch \(b\)}
    \State \(Z_{\rm tab}\!\gets\! f_T(X_{\rm tab}^{(b)}),\quad Z_{\rm img}\!\gets\! f_R(X_{\rm img}^{(b)})\)
    \State \(F \gets [Z_{\rm tab},Z_{\rm img}] \in \mathbb{R}^{B\times m}\)
    \State \((\pi,s)\gets {\rm GEDS}(F),\quad F_\pi \gets F[:,\pi]\)
    \State \((\hat{y},s')\gets {\rm OEMT}(F_\pi)\)
    \State \(\beta_{\pi(i)} \gets 1-\frac{i-1}{m-1},\quad i=1,\ldots,m\)
    \State \(\mathcal{L}\gets \mathcal{L}_{\rm CE}(y^{(b)},\hat{y})+\lambda_{\rm FS}\|s'-\beta\|_2^2\)
    \State Update parameters by backpropagating \(\mathcal{L}\)
  \EndFor
\EndFor
\State \Return trained iStructTab
\end{algorithmic}
\end{algorithm}
\section{Experiments and Results}
\label{sec:res}
\begin{figure*}[htbp]
  \centering
  \setlength{\abovecaptionskip}{4pt}
  \setlength{\belowcaptionskip}{-4pt}

  \begin{subfigure}[t]{0.45\textwidth}
    \centering
    \includegraphics[width=\linewidth]{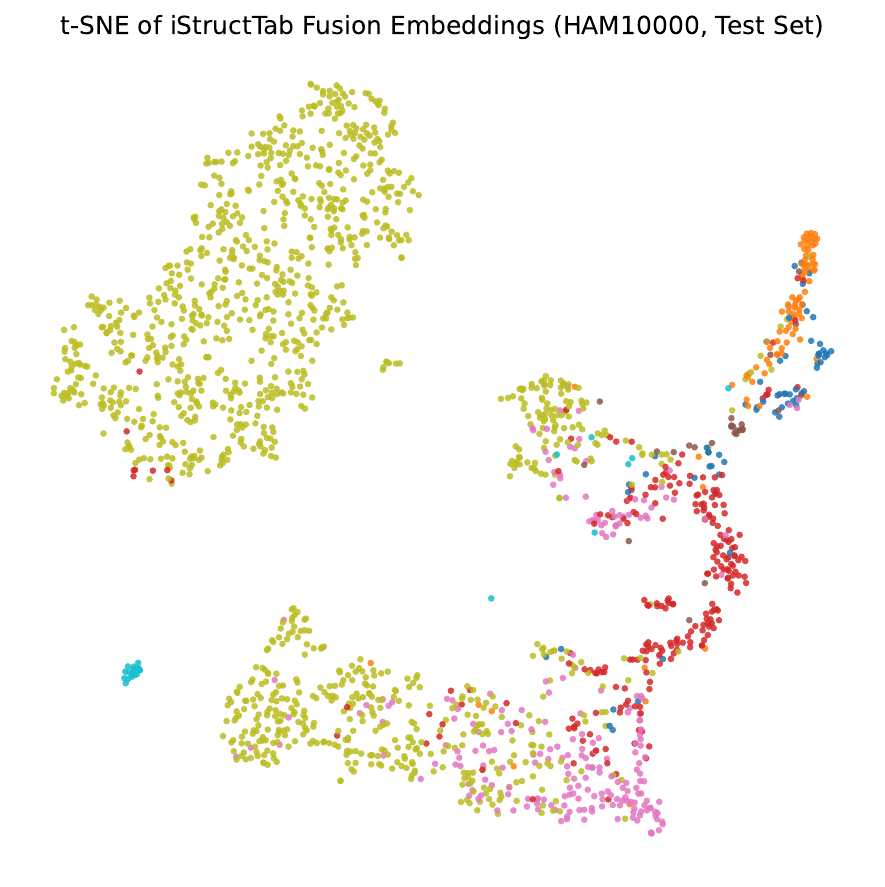}
    \caption{HAM10000}
    \label{fig:tsne_ham}
  \end{subfigure}\hfill
  \begin{subfigure}[t]{0.45\textwidth}
    \centering
    \includegraphics[width=\linewidth]{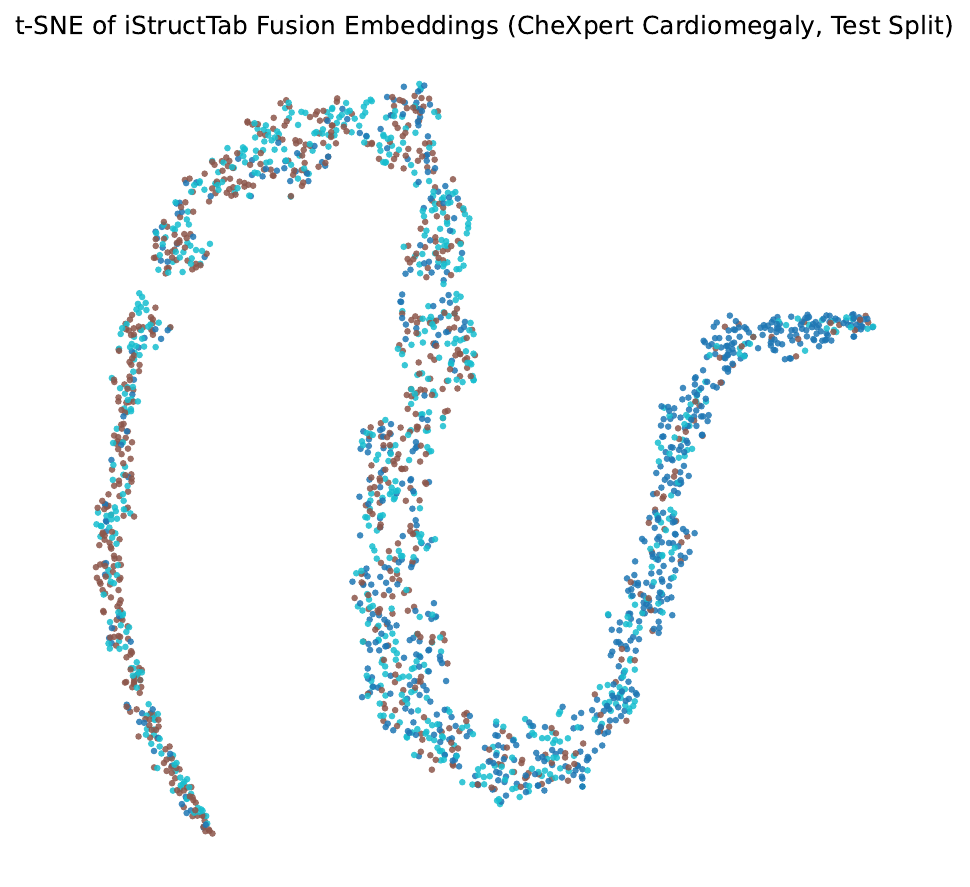}
    \caption{CheXpert (Cardiomegaly)}
    \label{fig:tsne_chexpert}
  \end{subfigure}

  \caption{t-SNE visualization of fused embeddings produced by iStructTab on two representative medical imaging-based multimodal (image+tabular) datasets. 
  Following STiL~\citep{stil}, we visualize the joint representation to assess emerging cluster structure
  and cross-modal alignment between image and tabular channels.}
  \label{fig:tsne_pair}
\end{figure*}

\begin{figure}[htbp]
  \centering
  \setlength{\abovecaptionskip}{2pt}
  \setlength{\belowcaptionskip}{0pt}

  \begin{subfigure}[t]{0.22\columnwidth}
    \centering
    \includegraphics[width=\linewidth,clip,trim=2pt 2pt 2pt 2pt]{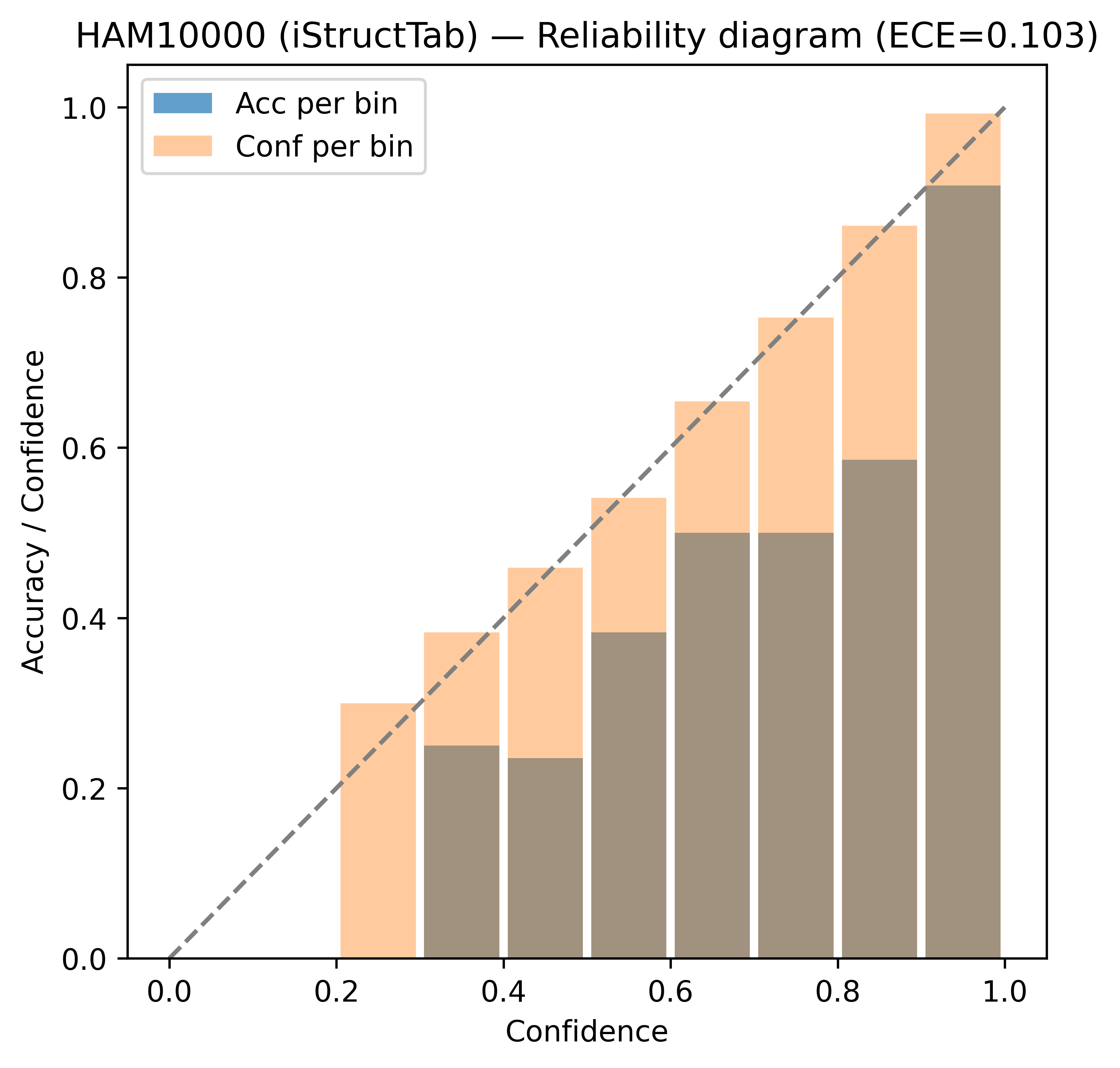}
    \caption{Reliability (ECE = 0.103)}
    \label{fig:ham_reliability}
  \end{subfigure}\hfill
  \begin{subfigure}[t]{0.27\columnwidth}
    \centering
    \includegraphics[width=\linewidth,clip,trim=2pt 2pt 2pt 2pt]{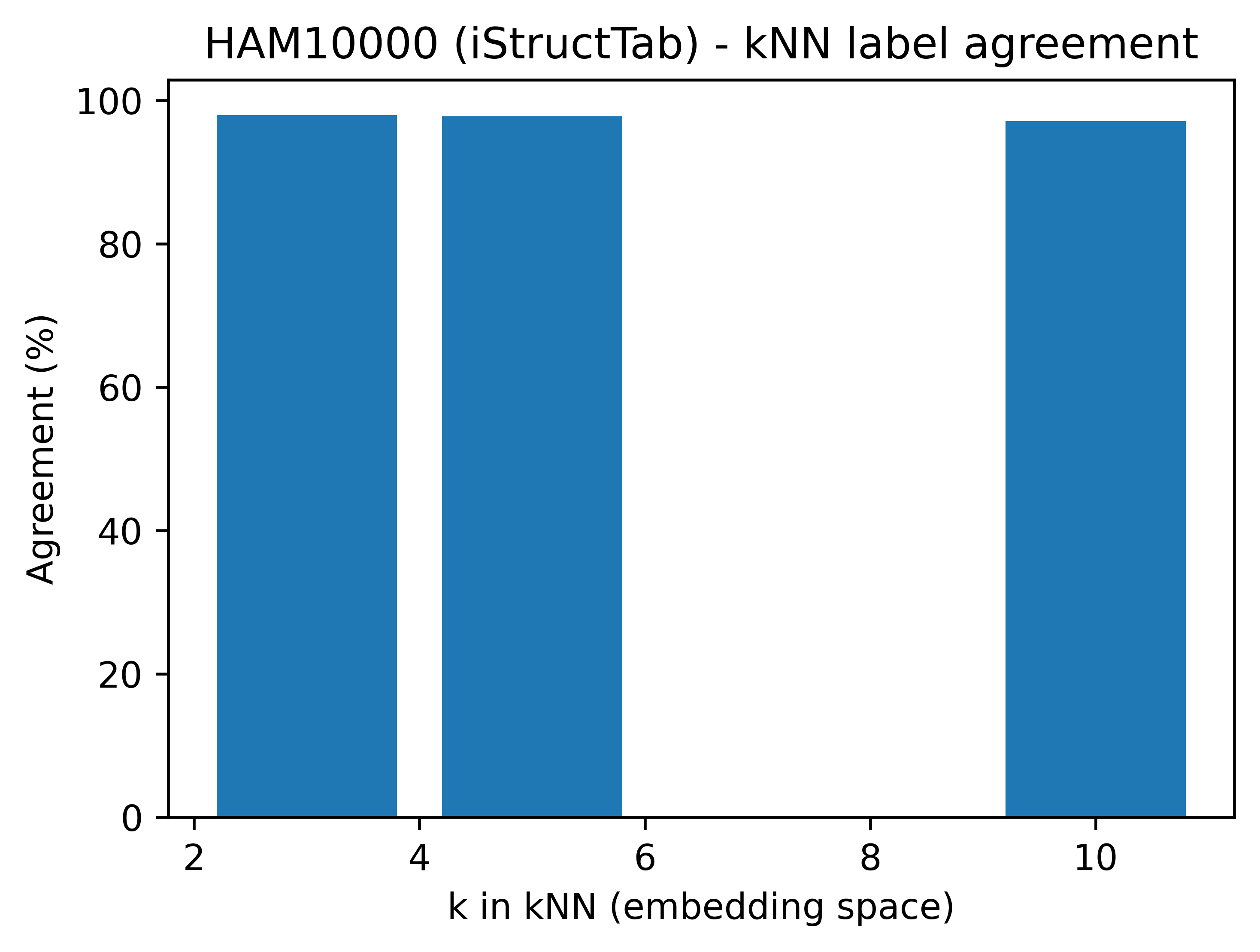}
    \caption{kNN label agreement}
    \label{fig:ham_knn}
  \end{subfigure}\hfill
  \begin{subfigure}[t]{0.31\columnwidth}
    \centering
    \includegraphics[width=\linewidth,clip,trim=2pt 2pt 2pt 2pt]{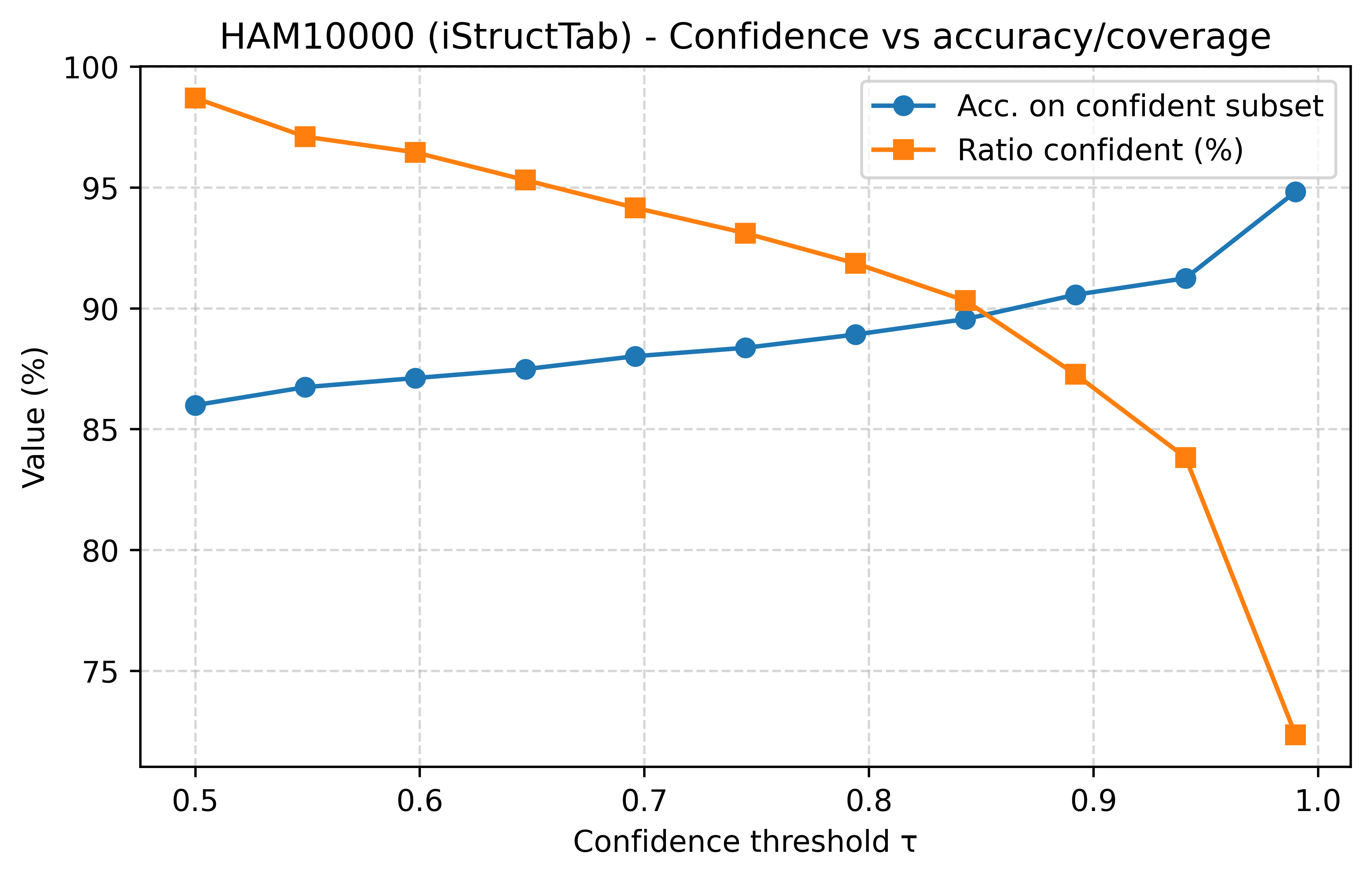}
    \caption{Confidence vs acc and coverage}
    \label{fig:ham_confidence}
  \end{subfigure}
  \vspace{2pt}
  \begin{subfigure}[t]{0.31\columnwidth}
    \centering
    \includegraphics[width=\linewidth,clip,trim=2pt 2pt 2pt 2pt]{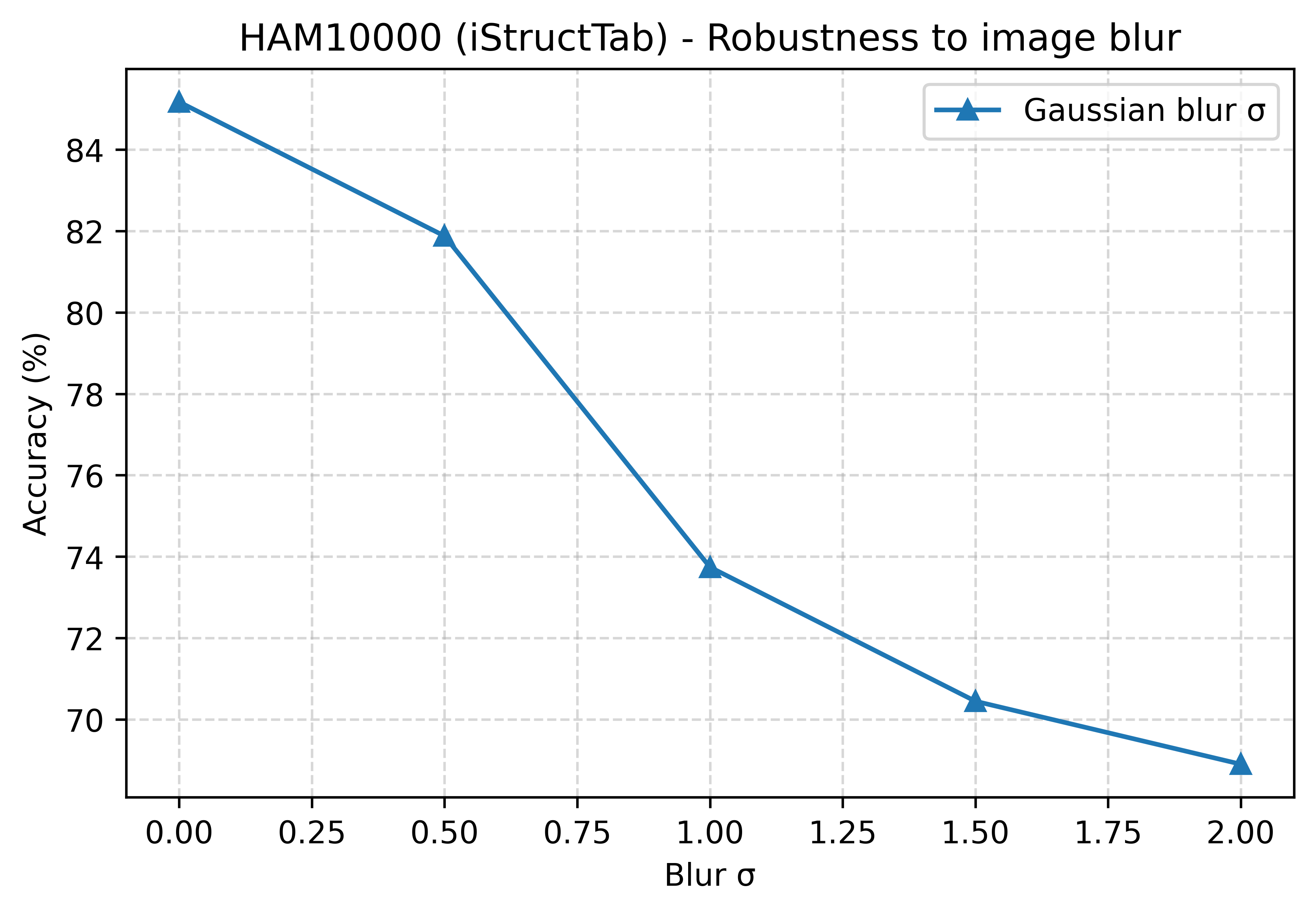}
    \caption{Robustness to image blur}
    \label{fig:ham_blur}
  \end{subfigure}\hfill
  \begin{subfigure}[t]{0.31\columnwidth}
    \centering
    \includegraphics[width=\linewidth,clip,trim=2pt 2pt 2pt 2pt]{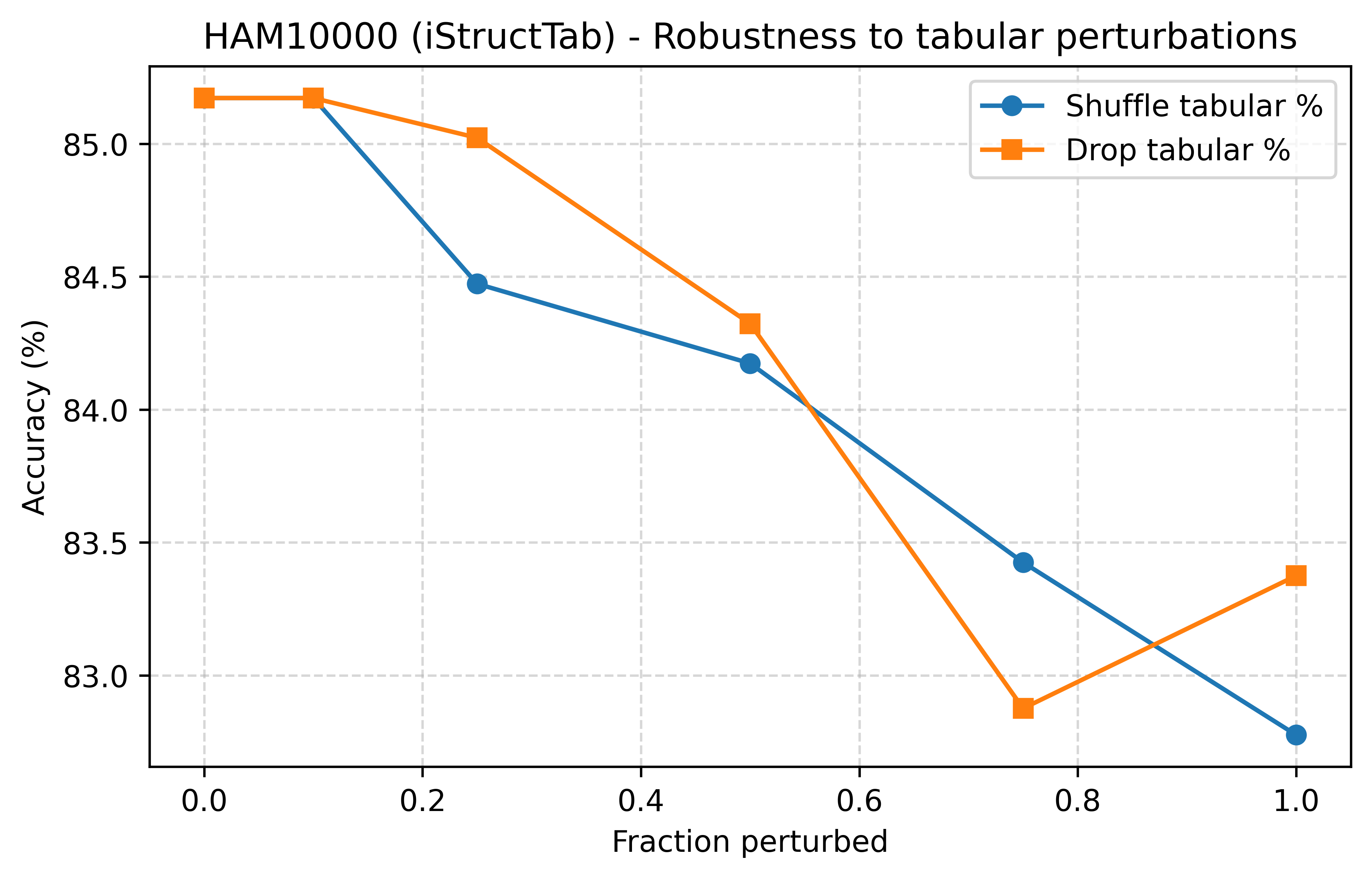}
    \caption{Robustness to tabular noise}
    \label{fig:ham_tabular}
  \end{subfigure}\hfill
  \begin{subfigure}[t]{0.31\columnwidth}
    \centering
    \includegraphics[width=\linewidth,clip,trim=2pt 2pt 2pt 2pt]{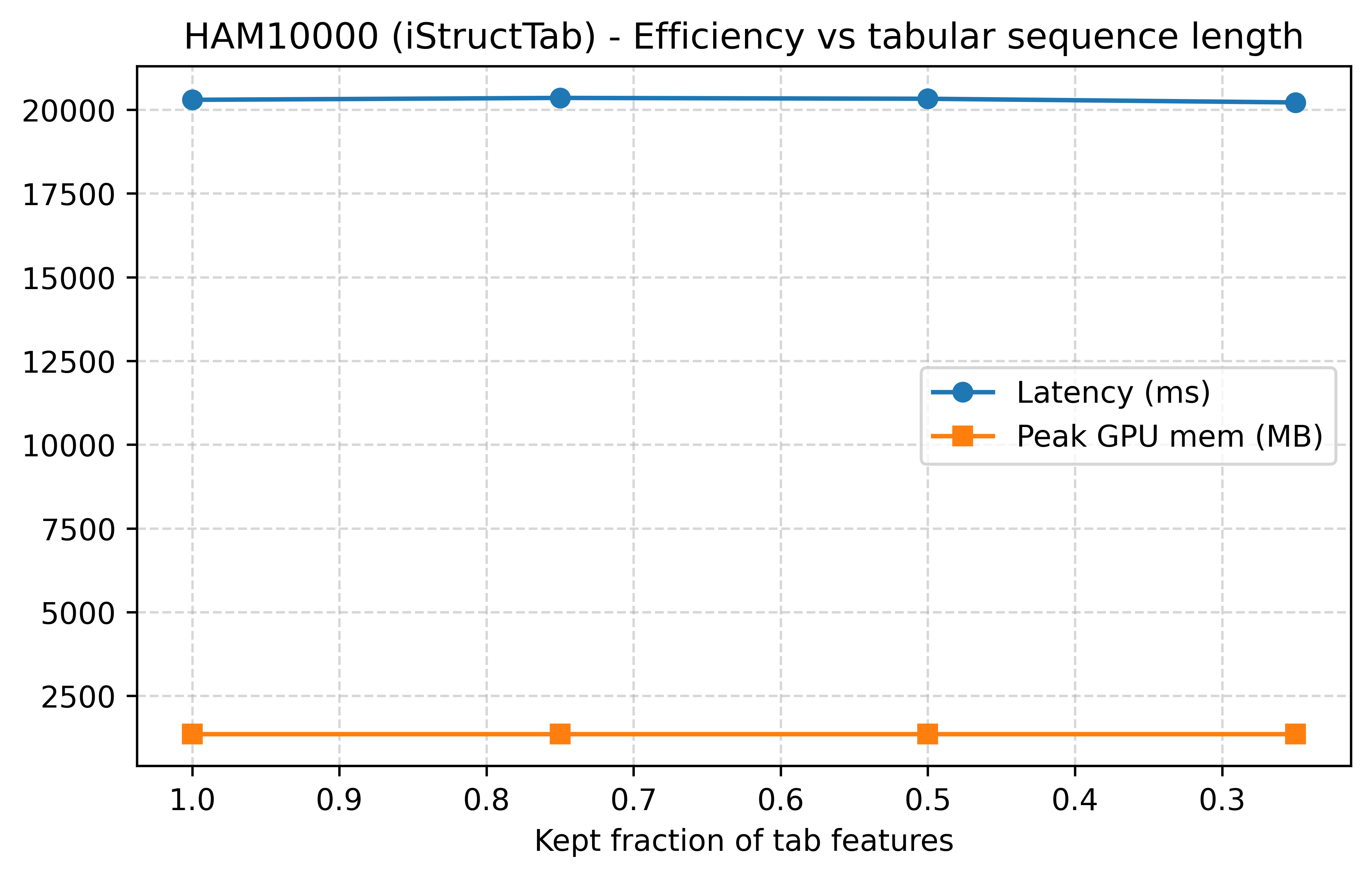}
    \caption{Efficiency vs fraction of tab features retained}
    \label{fig:ham_efficiency}
  \end{subfigure}

  \caption{Inference level ablations of iStructTab on HAM10000 in the style of STiL \citep{stil}: calibration (a), neighborhood consistency (b), confidence accuracy tradeoff (c), robustness to Gaussian blur (d) and tabular perturbations (e), and latency and memory versus tabular sequence length (f).}
  \label{fig:ham_ablation_onecol}
\end{figure}

\textbf{A: Datasets and Evaluation Metrics.}
We evaluate iStructTab on six image-tabular datasets. DVM~\citep{huang2022dvmcar} has 176{,}414 car images with 17 attributes and 283 classes; missing tabular values are imputed using CTGAN~\citep{ctgan}. Deep Lesion (DLes)~\citep{yan2018deep,dl} contains CT slices with lesion and patient metadata; after expanding coordinates and aligning pairs, we use 1{,}327 samples with 35 features. HAM10000 (HAM)~\citep{tschandl2018ham10000,codella2019skin,ham} provides 10{,}015 dermoscopic images with 4 metadata fields. CheXpert (CheX)~\citep{irvin2019chexpert,chex} is reduced to 9{,}999 frontal X-ray pairs with 17 metadata features for Cardiomegaly classification. Pokémon (Pok)~\citep{pokemon} includes 1{,}025 samples with 16 attributes for type prediction, and Pet Finder (Pet)~\citep{pet} has 14{,}652 pet images with 19 features for adoption-speed prediction. We use stratified 64/16/20 train/validation/test splits and report test accuracy, average rank and regret (Table~\ref{tab:cvpr_all}), noise robustness (Table~\ref{tab:ham_noise}), and efficiency (Fig.~\ref{fig:dvm_efficiency}). Significance is assessed with Friedman~\citep{fried}, Nemenyi~\citep{nemen}, and Wilcoxon-Holm~\citep{cd} tests, with CD diagrams in the Suppl. Mat.\newline
\textbf{B: Implementation Details.}
Image features are extracted with ResNet~\citep{he2016deep}, following STiL~\citep{stil} and TIP~\citep{du2024tip}. Tabular inputs are encoded by a transformer encoder and fused with image features through GEDS. We tune iStructTab with Optuna~\citep{akiba2019optuna} for 20 trials per dataset, following common tabular tuning practice~\citep{tabm,tabr,embed,b19}; baselines use recommended settings. Experiments use PyTorch~2.4.1+CUDA~12.1 with AMP, running tabular models on an RTX~2000 Ada 16\,GB workstation and multimodal models on TITAN RTX 24\,GB GPUs. The 20-trial budget ensures fair, compute-constrained tuning; pilot sweeps showed early validation saturation with marginal gains from extra trials. Tuned parameters are listed in the Suppl. Mat. Missing values are handled by median numerical imputation and categorical/text encoding, except DVM, where extensive missingness is imputed with CTGAN~\citep{ctgan}.
\newline
\textbf{C: Computational Complexity Analysis.}
iStructTab is dominated by GEDS sequencing and OEMT encoding. For fused features \(F\in\mathbb{R}^{N\times m}\), GEDS computes feature statistics in \(\mathcal{O}(Nm)\) and forms/refines the descriptor affinity graph in \(\mathcal{O}(m^2)\), giving \(\mathcal{O}(Nm+m^2)\). OEMT processes a pooled sequence of length \(L=k+M\), hidden size \(d_t\), and depth \(H\). With Linformer attention~\citep{wang2020linformer}, each layer costs \(\mathcal{O}(Lkd_t+Ld_t^2)\), instead of \(\mathcal{O}(L^2d_t)\) for full attention, where \(k\ll L\le m\). Thus, the per-batch cost is \(\mathcal{O}(Nm+m^2+H(Lkd_t+Ld_t^2))\). Although GEDS includes an \(\mathcal{O}(m^2)\) affinity step, \(m\) is the encoded fused dimension, not raw image pixels; in our benchmarks it is only hundreds to a few thousand, so runtime is dominated by the CNN/Transformer encoders. For larger \(m\), the affinity can be approximated using sparse \(k\)-NN graphs or low-rank similarity estimation. Layer-wise FLOP and memory details are in the Suppl. Mat., and empirical efficiency is shown in Fig.~\ref{fig:dvm_efficiency}. \newline
\indent \textbf{D: Comparative Results.}
 We consider three aspects: 
 \newline
\textbf{D1: Comparing w/ Multimodal and Standard Baselines.}
Across six benchmarks, iStructTab performs best overall (Table~\ref{tab:cvpr_all}), achieving the top Avg. Rank (1.50 $\pm$ 0.76) and lowest Avg. Regret (2.21 $\pm$ 4.59). Gains are strongest on multimodal-friendly datasets such as Pok and Pet, where aligning image cues with tabular context outperforms image-only and tabular-only baselines. STiL and TIP are strong multimodal competitors with lower regret than single-modality methods, but still trail iStructTab on most datasets, especially when metadata helps disambiguate visually similar or noisy samples. ViT is competitive on image-dominant tasks such as DVM, while tabular-only models show occasional gains but higher overall regret. Overall, the results suggest a consistent trend: multimodal fusion $>$ image-only $>$ tabular-only, with iStructTab benefiting most from alignment-aware image-tabular sequencing.\newline
\textbf{D2: Comparing in Noisy Scenario.}
iStructTab is robust to label noise (Table~\ref{tab:ham_noise}), outperforming prior methods at all noise rates and improving over CUFIT~\citep{yu2024curriculum} on average (80.42\% vs.\ 78.3\%). Gains are modest at 10-40\% noise but larger under severe corruption, reaching 76.42\% vs.\ 70.1\% at 60\% noise. This suggests that iStructTab preserves discriminative image-tabular signals better under noisy supervision.\newline
\begin{table*}[htbp]
\centering
\caption{Test accuracy (\%) across six datasets (higher is better). Baselines largely follow the STiL benchmark (CVPR~2025); we additionally include strong tabular-only baselines (LGBM, CatBoost, TabM, TabSeq) and an image-only ViT baseline, and evaluate all methods on our 
test datasets. Avg. Rank is computed per dataset by ranking all 16 methods together (descending) with the standard ``average tie'' rule; we report mean $\pm$ std across datasets (lower is better). Avg. Regret (↓) is the mean gap to the per-dataset best score across the six datasets (mean $\pm$ std). Here, 
\dag denotes results taken from \citep{stil} and * denotes results taken from \citep{miita}.}
\footnotesize
\setlength{\tabcolsep}{6pt}
\renewcommand{\arraystretch}{1.02}
\setlength{\arrayrulewidth}{0.4pt}
\begin{tabular}{|l|c|c|r|r|r|r|r|r|c|c|}
\hline
\multicolumn{3}{|c|}{} & \multicolumn{6}{c|}{\textit{Dataset statistics}} & \multicolumn{2}{c|}{} \\
\cline{4-9}
\multicolumn{3}{|l|}{\textit{Images}}          & 176414 & 10015 & 1327 & 1025 & 9999 & 14652 & \multicolumn{2}{c|}{} \\
\multicolumn{3}{|l|}{\textit{Tabular Features}} & 17     & 4     & 35   & 16   & 17   & 19    & \multicolumn{2}{c|}{} \\
\multicolumn{3}{|l|}{\textit{Classes}}          & 283    & 7     & 3    & 18   & 3    & 5     & \multicolumn{2}{c|}{} \\
\hline
\multicolumn{1}{|c|}{Model} & \multicolumn{2}{c|}{Modality} & \multicolumn{1}{c|}{DVM} & \multicolumn{1}{c|}{HAM} & \multicolumn{1}{c|}{DLes} & \multicolumn{1}{c|}{Pok} & \multicolumn{1}{c|}{CheX} & \multicolumn{1}{c|}{Pet} & \multicolumn{1}{c|}{Avg. Rank $\downarrow$} & \multicolumn{1}{c|}{Avg. Regret $\downarrow$} \\
\cline{2-3}
\multicolumn{1}{|c|}{} & \multicolumn{1}{c|}{I} & \multicolumn{1}{c|}{T} &  &  &  &  &  &  &  &  \\
\hline
\multicolumn{11}{|c|}{\textit{Tabular Classifiers}} \\
\hline
LGBM \citep{lgbm}                  &  & \checkmark & 26.89 & 71.80 & 76.71 & 45.73 & 52.25 & 39.70 & 8.67 $\pm$ 3.14  & 33.45 $\pm$ 22.03 \\
CatBoost \citep{catboost}          &  & \checkmark & 27.30 & 72.05 & 77.18 & 51.83 & 53.19 & 39.40 & 7.67 $\pm$ 2.69  & 32.13 $\pm$ 22.65 \\
TabSeq \citep{habib2024tabseq}     &  & \checkmark & 26.66 & 70.14 & 83.46 & 15.61 & 51.10 & 37.33 & 11.08 $\pm$ 4.25 & 38.24 $\pm$ 23.99 \\
SCARF \citep{bahri2021scarf}       &  & \checkmark & 64.47\dag & 69.40 & 83.08 & 19.02 & 51.35 & 33.03 & 10.33 $\pm$ 3.73 & 32.23 $\pm$ 18.44 \\
SAINT \citep{saint}                &  & \checkmark & 83.36\dag & 70.14 & 69.17 & 13.17 & 35.50 & 27.29 & 13.00 $\pm$ 3.50 & 35.85 $\pm$ 20.49 \\
TabM \citep{tabm}                  &  & \checkmark & 27.09 & 70.84 & \textbf{84.59} & 18.05 & 51.90 & 38.01 & 9.08 $\pm$ 4.23  & 37.21 $\pm$ 24.00 \\
\hline
\multicolumn{11}{|c|}{\textit{Image Classifiers}} \\
\hline
ResNet50 \citep{he2016deep}        & \checkmark &  & 32.07\dag & 82.13 & 69.17 & 31.22 & 48.35 & 35.52 & 10.25 $\pm$ 3.16 & 35.88 $\pm$ 21.46 \\
ViT \citep{ViT}                    & \checkmark &  & 88.00* & 82.33 & 69.17 & 35.61 & 48.70 & 34.94 & 8.58 $\pm$ 3.83  & 25.83 $\pm$ 17.41 \\
SimCLR \citep{chen2020simple}      & \checkmark &  & 51.44\dag & 82.38 & 64.66 & 63.40* & 56.15 & 31.76 & 8.42 $\pm$ 4.87  & 27.33 $\pm$ 20.11 \\
BYOL \citep{grill2020bootstrap}    & \checkmark &  & 47.49\dag & 80.78 & 56.39 & 24.39 & 51.25 & 34.77 & 10.83 $\pm$ 2.97 & 36.45 $\pm$ 16.62 \\
\hline
\multicolumn{11}{|c|}{\textit{Multimodal Classifiers}} \\
\hline
DAFT \citep{wolf2022daft}          & \checkmark & \checkmark & 74.22\dag & 74.88 & 77.07 & 18.05 & 67.75 & 45.45 & 7.42 $\pm$ 2.71  & 26.06 $\pm$ 15.64 \\
Interact Fuse \citep{duanmu2020prediction} & \checkmark & \checkmark & 78.58\dag & 84.87 & 70.68 & 11.71 & 50.90 & 52.44 & 8.33 $\pm$ 4.71  & 27.43 $\pm$ 18.16 \\
MMCL \citep{hager2023best}         & \checkmark & \checkmark & 85.79\dag & 70.09 & 67.34 & 64.60* & 49.00 & 29.72 & 10.67 $\pm$ 5.19 & 24.54 $\pm$ 18.47 \\
TIP \citep{du2024tip}              & \checkmark & \checkmark & 98.27\dag & 70.39 & 69.17 & 63.40* & 87.50 & 83.86 & 5.83 $\pm$ 4.21  & 6.86 $\pm$ 6.01 \\
STiL \citep{stil}                  & \checkmark & \checkmark & 99.27\dag & 78.48 & 81.35 & 27.32 & \textbf{88.60} & 87.68 & 4.33 $\pm$ 2.92  & 8.51 $\pm$ 14.72 \\
\textbf{iStructTab (ours)}         & \checkmark & \checkmark & \textbf{99.29} & \textbf{85.23} & 83.75 & \textbf{68.29} & 76.15 & \textbf{87.75} & \textbf{1.50 $\pm$ 0.76} & \textbf{2.21 $\pm$ 4.59} \\
\hline
\end{tabular}
\label{tab:cvpr_all}
\end{table*}
\begin{figure*}[htbp]
  \centering
  \setlength{\tabcolsep}{2pt} 
  \renewcommand{\arraystretch}{0} 
  \begin{tabular}{cccc}
    \includegraphics[width=0.24\linewidth]{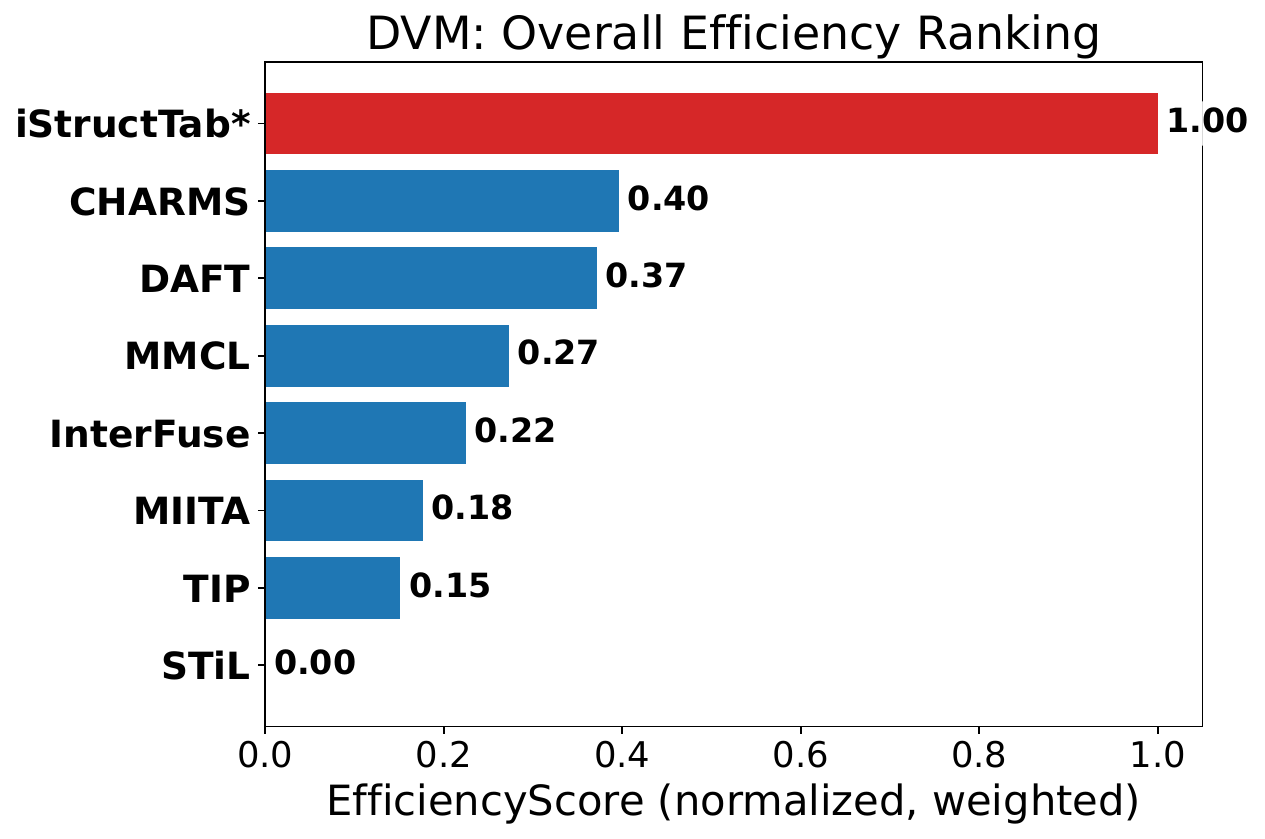} &
    \includegraphics[width=0.22\linewidth]{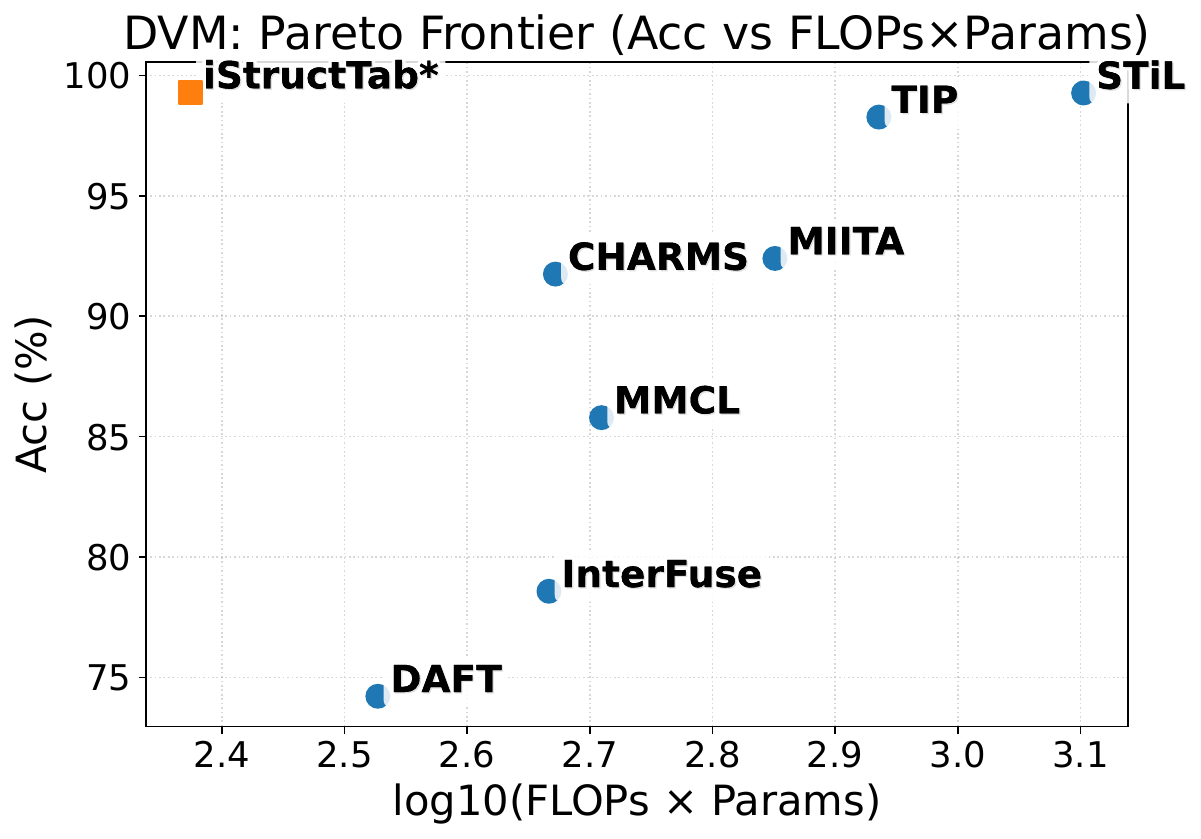} &
    \includegraphics[width=0.22\linewidth]{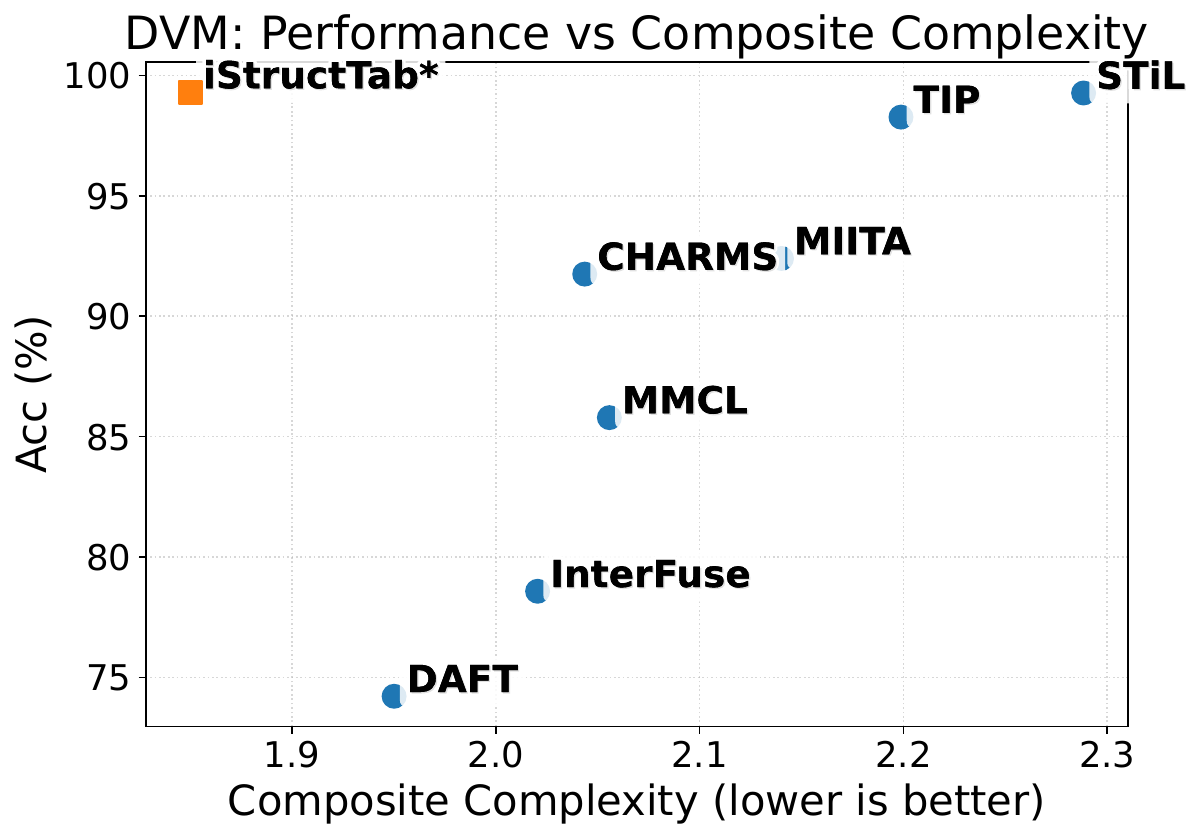} &
    \includegraphics[width=0.24\linewidth]{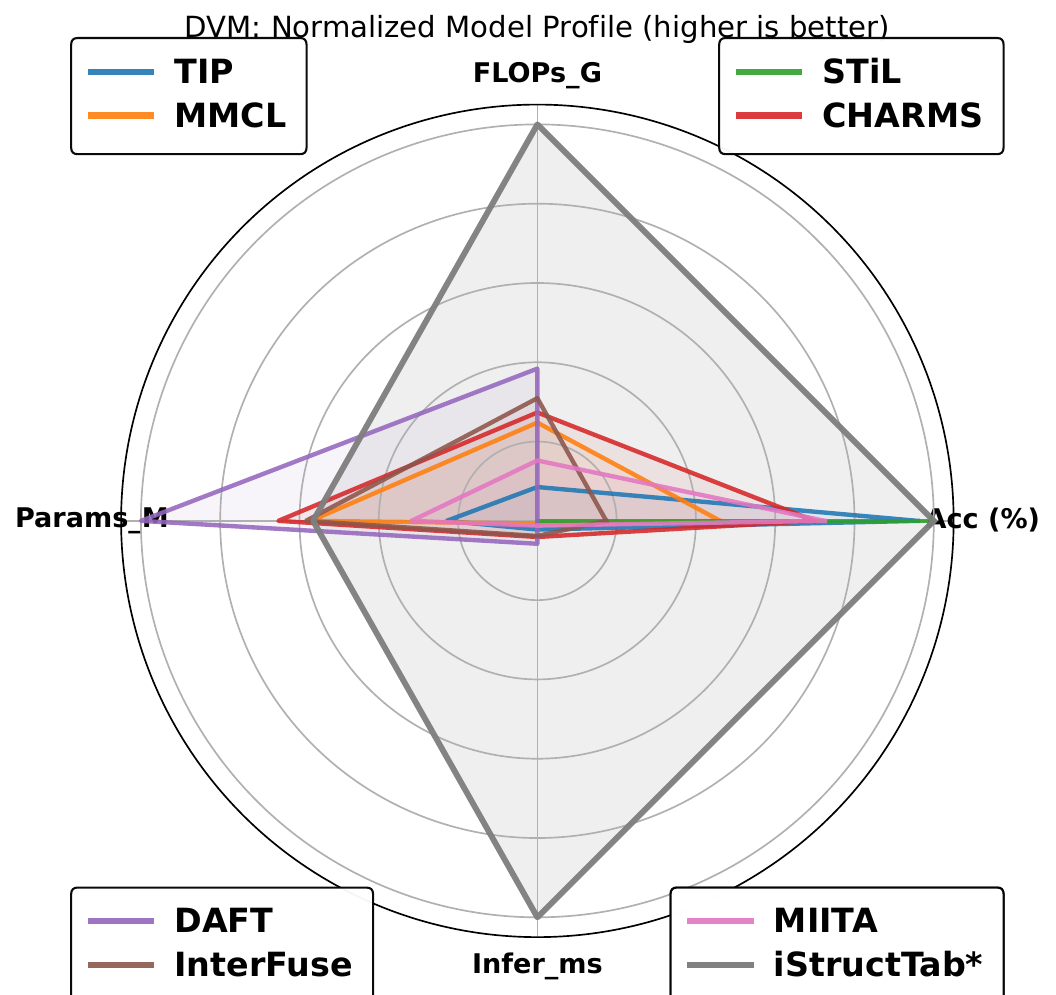}
    \\
     [2pt]
    (a) Efficiency ranking &
    (b) Pareto front &
    (c) Acc vs composite complexity &
    (d) Radar profile
  \end{tabular}
  \caption{
  Comparison of multimodal image-tabular methods on the DVM dataset:
  (a) Weighted efficiency ranking combining performance and resource usage;
  (b) Pareto frontier of accuracy vs.~complexity (\( \mathrm{FLOPs}\times\mathrm{Params} \));
  (c) Trade-off between accuracy and composite complexity;
  (d) Normalized radar chart summarizing accuracy, computation, and inference latency.
  The reported accuracies are directly taken from the respective papers; for the TIME model, accuracy is unavailable (\texttt{NaN}).
  Peak memory usage, performance vs.~FLOPs, performance vs.~parameters, and parameter–FLOP relationships are included in the Supplementary Material.
  }
\label{fig:dvm_efficiency}
\end{figure*}
\begin{table}[t]
\centering
\scriptsize
\setlength{\tabcolsep}{4pt}
\begin{minipage}{0.45\linewidth}
    \centering
    \caption{Test accuracy (\%) under label noise on HAM. Baselines from CUFIT \cite{yu2024curriculum} (NeurIPS~2024); we add \textbf{iStructTab (Ours)} as an extra row. Bold indicates the best per column.}
    \label{tab:ham_noise}
    \begin{tabular}{@{}lccccc@{}}
    \toprule
    \textbf{Method/ noise level } & \textbf{0.1} & \textbf{0.2} & \textbf{0.4} & \textbf{0.6} & \textbf{Mean} \\
    \midrule
    Full-training \citep{yu2024curriculum}        & 66.5 & 62.6 & 56.1 & 59.9 & 61.3 \\
    Linear probing \citep{yu2024curriculum}       & 75.6 & 75.3 & 71.0 & 61.9 & 71.0 \\
    Rein \citep{wei2024stronger}                  & 78.6 & 72.1 & 54.9 & 37.8 & 60.8 \\
    Co-teaching \cite{han2018co}                  & 81.5 & 79.1 & 74.3 & 67.3 & 75.5 \\
    JoCor \citep{wei2020combating}                & 81.1 & 79.4 & 73.9 & 67.1 & 75.4 \\
    CoDis \citep{xia2023combating}                & 81.9 & 80.1 & 74.1 & 66.1 & 75.5 \\
    CUFIT \citep{yu2024curriculum}                & 82.6 & 81.5 & 79.1 & 70.1 & 78.3 \\
    \textbf{iStructTab (Ours)}                    & \textbf{83.92} & \textbf{81.92} & \textbf{79.42} & \textbf{76.42} & \textbf{80.42} \\
    \bottomrule
    \end{tabular}
\end{minipage}
\hfill
\begin{minipage}{0.48\linewidth}
    \centering
    \caption{Ablation on iStructTab components on the DVM dataset under different configurations (t*=tuning).}
    \label{tab:ablation}
    \vspace{2pt}
    \begin{tabular}{@{}lcc@{}}
    \toprule
    \textbf{Variant} & \textbf{w/o t*} & \textbf{w/ t*} \\ 
    \midrule
    Concat Fuse (no sequencing, no memory)         & 90.42 & 89.60 \\
    iStructTab (ResNet-50 + GEDS)                 & \textbf{96.85} & \textbf{99.29} \\
    iStructTab (ResNeXt-50 + GEDS)                & 91.88 & 93.33 \\
    iStructTab w/o memory tokens                  & 92.84 & 87.60 \\
    iStructTab w/o sequencing-loss term           & 96.85 & 96.78 \\
    iStructTab w/o feature sequencing             & 80.80 & 82.98 \\
    \shortstack[l]{iStructTab no sequencing\\w/ memory loss} & 13.91 & 16.86 \\
    \bottomrule
    \end{tabular}
\end{minipage}
\end{table}
\textbf{D3: Comparing Computational Efficiency.}
We compare the efficiency of multimodal image-tabular models~\citep{du2024tip,hager2023best,stil,charms,time,wolf2022daft,miita,duanmu2020prediction} on DVM by jointly measuring accuracy and resource use. For each method, we compute an efficiency score \(\mathrm{Eff}_i\) from min-max normalized GFLOPs, parameters, latency, and composite complexity, using weights \(\mathbf{w}=(0.40,0.25,0.20,0.15)\) to emphasize computation. Fig.~\ref{fig:dvm_efficiency} shows that iStructTab achieves the best accuracy-efficiency trade-off: it has near-maximum accuracy with the lowest FLOPs, parameters, and latency, and lies in the low-complexity/high-accuracy Pareto region. Its overall score \(\mathrm{Eff}_i=1.00\) is more than twice the next-best method, CHARMS (\(0.40\)), showing that GEDS+OEMT improves both accuracy and efficiency.\newline
\textbf{E: Ablation on Model Components.}
Table~\ref{tab:ablation} shows that the full iStructTab model (ResNet-50 + GEDS + OEMT) performs best on DVM, reaching 96.85\% without tuning and 99.29\% with tuning, outperforming simple concatenation by roughly 6-10 points. Replacing ResNet-50 with ResNeXt-50 reduces accuracy, suggesting that gains come mainly from GEDS sequencing and OEMT rather than a stronger image backbone. Removing memory tokens causes a large drop, especially with tuning (99.29\% to 87.60\%), confirming their role in capturing global context. Removing the sequencing loss gives a smaller decline (99.29\% to 96.78\%), indicating useful but secondary regularization. In contrast, removing feature sequencing collapses performance to about 81-83\%, while applying memory-based sequencing loss without an ordering signal fails severely (13.91-16.86\%). Overall, GEDS ordering and memory-based OEMT jointly drive performance, whereas absent or misaligned sequencing is detrimental. Additional analyses on representation quality, calibration, and robustness are provided in the Supplementary Material.\newline
\textbf{F. Representation Quality via t-SNE.}  
Following the STiL~\citep{stil} protocol of CVPR 2025, we inspect the fused image-tabular embeddings produced by iStructTab using t-SNE~\citep{ts1}; the resulting projections for HAM10000 and CheXpert (Cardiomegaly) are shown in Fig.~\ref{fig:tsne_pair}. In both cases we visualize the joint representation after fusion (rather than raw features), so the geometry reflects how iStructTab organizes examples in a shared space across modalities. While t-SNE is only a qualitative tool and primarily preserves local neighborhoods, it is still informative for assessing whether the model learns coherent manifolds and whether cross-modal signals are aligned~\citep{ts2}.
\begin{table}[t]
\scriptsize
\centering
\caption{Ablation summary on accuracy of {iStructTab} using HAM10000. Tab Shuffle / Tab Drop denote the fraction of shuffled or dropped tabular features; accuracies and kNN-Agree@5 are in \%. Expected Calibration Error (ECE) = 0.103.}
\label{tab:ham_istructtab_ablation}
\begin{tabular}{@{}ccccc@{}}
\toprule
\textbf{Frac.} &
\textbf{Tab Shuffle $\downarrow$} &
\textbf{Tab Drop $\downarrow$} &
\textbf{Img Blur $\sigma$} &
\textbf{kNN-Agree@5} \\
\midrule
0.00 & 85.17 & 85.17 & 0.0 & 97.79 \\
0.10 & 85.17 & 85.17 & 0.5 & 97.79 \\
0.25 & 84.47 & 85.02 & 1.0 & 97.79 \\
0.50 & 84.17 & 84.32 & 1.5 & 97.79 \\
0.75 & 83.42 & 82.88 & 2.0 & 97.79 \\
1.00 & 82.78 & 83.37 & -  & 97.79 \\
\bottomrule
\end{tabular}
\end{table}
On HAM10000 (Fig.~\ref{fig:tsne_ham}) we observe large, smooth manifolds with clear contiguous regions and relatively dense local neighborhoods, despite the seven-way lesion taxonomy. Points concentrate along a few dominant structures rather than being scattered uniformly across the plane. This suggests that iStructTab learns a representation in which semantically related cases (lesions with similar appearance and metadata) occupy nearby regions and that the fused feature space has relatively low intrinsic dimensionality for downstream classification. The lack of severe fragmentation or many isolated “clouds” is consistent with a representation that can be separated by comparatively simple decision boundaries, aligning with the strong HAM10000 performance reported in the main tables. In contrast, the CheXpert Cardiomegaly projection (Fig.~\ref{fig:tsne_chexpert}) exhibits a more elongated, continuum-like geometry: embeddings lie along a curved manifold where negative, uncertain, and positive cases interleave rather than forming sharply separated clusters. This pattern matches the underlying clinical semantics, where cardiomegaly severity is closer to a spectrum and label noise (e.g., “uncertain” labels) is substantial. The fused representation appears to organize patients along this severity axis, but with weaker local cohesion between discrete label bins, indicating that decision boundaries must account for both uncertainty in the labels and subtler cross-modal cues. Together, these qualitative views corroborate our quantitative results: HAM10000 emerges as a setting where iStructTab achieves especially strong cluster structure and cross-modal alignment, motivating its use as the primary target for our inference-only ablations and robustness studies, whereas CheXpert provides a noisier, spectrum-style benchmark that stresses the model’s ability to handle ambiguous and weakly supervised labels.\newline
\indent \textbf{G. Inference Level Ablation on Calibration and Robustness.}
\textbf{G. Inference-Level Ablation on Calibration and Robustness.}
Following the STiL~\citep{stil} protocol, we evaluate iStructTab inference reliability on HAM10000 through calibration, neighborhood consistency, and robustness to visual/tabular perturbations. Table~\ref{tab:ham_istructtab_ablation} shows that accuracy degrades smoothly under tabular shuffling/dropping and Gaussian blur, while fused-space kNN label agreement remains high (\(\sim\)98\%) and ECE is low at 0.103~\citep{cali1}. Fig.~\ref{fig:ham_ablation_onecol} further shows that confidence tracks accuracy, noise causes graceful degradation, and latency/GPU memory remain nearly flat as the effective tabular sequence length varies~\citep{kn2,kn1}. Overall, iStructTab provides calibrated predictions, stable embeddings, and robust inference without retraining.
\section{Conclusion}
\label{con}
We introduced iStructTab, a multimodal architecture that treats image-tabular fusion as feature sequencing over fused representations. By formulating sequencing as a CPP, GEDS learns a structured, data-driven feature permutation, while OEMT exploits this order through memory-augmented pooling and a sequencing loss. Across six benchmarks, iStructTab improves over concatenation and strong state-of-the-art baselines. DVM and HAM10000 ablations confirm that GEDS ordering and OEMT memory tokens are key to accuracy, calibration, perturbation robustness, and OOD stability. Complexity analysis shows that iStructTab remains scalable through efficient token projection and Linformer-style encoding. Future work will study trade-offs among embedding dimension, pooling length, runtime, and performance, and extend iStructTab to larger-scale pretraining and additional modalities.
\section*{Acknowledgement}
This work was supported in part by the US National Science Foundation under Awards \#1920920, \#2125872, and \#2223793, and by the International Association for Pattern Recognition (IAPR) through ICPR 2026 registration support.
\bibliographystyle{splncs04}
\bibliography{ref}
\input{supple}
\end{document}

%% file: supple.tex
\clearpage
\maketitlesupplementary

\setcounter{section}{0}
\renewcommand{\thesection}{A}
\renewcommand{\thesubsection}{A.\arabic{subsection}}
\setcounter{figure}{0}\renewcommand{\thefigure}{A.\arabic{figure}}
\setcounter{table}{0}\renewcommand{\thetable}{A.\arabic{table}}


This supplementary document supports our main paper \textit{iStructTab: Structured Feature Sequencing for Multimodal Learning of Image and Tabular Data} (Submitted to the 28\textsuperscript{th} International Conference on Pattern Recognition (ICPR) 2026). Specifically, it includes:
\begin{itemize}
    \item iStructTab Hyperparameters in Sec.~\ref{ab2}
    \item Baseline Configurations in Sec.~\ref{ab3}
    \item More on Computational Efficiency Comparison in Sec.~\ref{ab4}
    \item Statistical Significance Analysis in Sec.~\ref{ab5}
    \item Data Efficiency Analysis in Sec.~\ref{ab6}
    \item Sanity and Stress Diagnostics in Sec.~\ref{ab9}
    \item Additional Reliability and Interpretability Diagnostics in Sec.~\ref{ab10}
    \item Theory-Inspired Representation Diagnostics in Sec.~\ref{ab11}
    \item Turing-Style Human-Model Evaluation in Sec.~\ref{ab12}
    \item OOD and Local Sensitivity Diagnostics in Sec.~\ref{ab13}
    \item Deployment-Oriented Triage Diagnostics in Sec.~\ref{ab14}
\end{itemize}

\section{iStructTab Hyperparameters}
\label{ab2}
The configurations in Table~\ref{tab:istructtab_optuna} summarize the Optuna~\citep{akiba2019optuna} search outcomes used for all reported iStructTab experiments on the multimodal benchmarks. For each dataset, we jointly tune the OEMT backbone capacity (token embedding size $d_{\text{model}}$, number of OEMT heads and layers, and OEMT projection rank $k_{\text{OEMT}}$), the downstream Linformer projection rank $k_{\text{Lin}}$, and the optimizer hyperparameters (learning rate, weight decay, batch size), together with the feature-sequencing loss weight $\lambda_{\mathrm{FS}}$. Higher-resolution or more class-imbalanced medical benchmarks such as CheXpert, DeepLesion, and HAM10000 tend to favor moderately wide embeddings with shallow OEMT stacks, whereas lighter benchmarks such as Pet and Pok\'emon work well with smaller $d_{\text{model}}$ and fewer layers. The DVM Car dataset, which has a richer set of visual prototypes, benefits from a larger OEMT embedding and a higher projection rank. For most datasets, $\lambda_{\mathrm{FS}}$ lies between roughly $0.05$ and $0.18$, while DeepLesion and DVM prefer much smaller values on the order of $10^{-5}$, indicating that the sequencing-consistency loss again acts as a secondary regularizer that stabilizes the learned feature orders without overwhelming the primary cross-entropy objective.
\begin{table*}[htbp]
  \caption{
    Optuna-tuned hyperparameters for iStructTab on each multimodal benchmark.
    $d_{\text{model}}$ is the OEMT token embedding dimension, ``OEMT heads'' and ``OEMT depth'' denote the number of OEMT attention heads and layers, $k_{\text{OEMT}}$ is the OEMT projection rank, and $k_{\text{Lin}}$ is the downstream Linformer projection rank.
    ``LR'' is the learning rate, ``WD'' the weight decay, ``BS'' the batch size, and $\lambda_{\mathrm{FS}}$ the weight of the feature-sequencing loss.
  }
  \centering
  \small
  \setlength{\tabcolsep}{6pt}
  \begin{tabular}{lccccccccc}
    \toprule
    Dataset &
    $d_{\text{model}}$ &
    OEMT heads &
    OEMT depth &
    $k_{\text{OEMT}}$ &
    $k_{\text{Lin}}$ &
    LR &
    WD &
    BS &
    $\lambda_{\mathrm{FS}}$ \\
    \midrule
    DVM    & 512 & 2 & 3 & 256 & 64  & $6.09\times 10^{-5}$ & $8.93\times 10^{-4}$ & 32  & $9.22\times 10^{-5}$ \\
    Pok\'emon  & 128 & 8 & 2 & 128 & 32  & $2.41\times 10^{-4}$ & $1.33\times 10^{-5}$ & 32 & 0.120 \\
    CheXpert   & 192 & 2 & 3 &  32 & 64  & $1.14\times 10^{-4}$ & $1.09\times 10^{-4}$ & 16 & 0.178 \\
    DeepLesion & 192 & 4 & 1 &  32 & 16  & $1.53\times 10^{-4}$ & $1.31\times 10^{-6}$ & 16 & $1.22\times 10^{-5}$ \\
    Pet        & 128 & 2 & 2 &  64 & 16  & $1.71\times 10^{-4}$ & $1.87\times 10^{-5}$ & 32 & 0.0576 \\
    HAM10000   & 192 & 2 & 1 &  32 & 64  & $1.91\times 10^{-4}$ & $7.23\times 10^{-5}$ & 32 & 0.175 \\
    \bottomrule
  \end{tabular}
  \label{tab:istructtab_optuna}
\end{table*}
\renewcommand{\thesection}{B}
\renewcommand{\thesubsection}{B.\arabic{subsection}}
\setcounter{figure}{0}\renewcommand{\thefigure}{B.\arabic{figure}}
\setcounter{table}{0}\renewcommand{\thetable}{B.\arabic{table}}
\section{Baseline Configurations}
\label{ab3}
\subsection{Tabular Classifiers}
We summarize the configuration of all tabular classifier baselines in Table~\ref{tab:tabular_classifiers}.  Gradient-boosted decision-tree methods (LightGBM, CatBoost) are run with standard multiclass settings and moderately expressive yet conventional hyperparameters. In contrast, deep tabular models (TabSeq, TabM, SAINT, and SCARF) are trained with fixed architectures and conservative optimization schedules on identical 64\%/16\%/20\% train/validation/test splits.  For the deep baselines, we employ early stopping on validation performance and avoid dataset-specific hyperparameter tuning beyond a single, shared configuration per model family.  This design keeps the comparison focused on intrinsic modeling differences, ensuring that the reported gains of iStructTab over these baselines are not artifacts of unequal tuning effort.
\begin{table*}[t]
\caption{
  Hyperparameters and training setup for the tabular classifier baselines.
  }
  \centering
  \scriptsize
  \setlength{\tabcolsep}{3pt}
  \begin{tabular}{p{0.13\textwidth}p{0.20\textwidth}p{0.63\textwidth}}
    \toprule
    Model & Implementation & Hyperparameters / training setup \\
    \midrule
    LightGBM (LGBM) &
    \textsc{LGBMClassifier} (LightGBM\citep{lgbm}) &
    Multiclass objective with $K$ classes; $n_{\text{estimators}} = 500$; learning rate $= 0.05$; 
    no maximum depth ($\text{max\_depth}=-1$); subsample $=0.9$; column subsample $=0.9$; 
    random seed $=42$. \\[3pt]
    
    CatBoost &
    \textsc{CatBoostClassifier} (CatBoost\citep{catboost}) &
    Multiclass loss; depth $=8$; $1000$ boosting iterations; learning rate $=0.05$; 
    $L_2$ regularization $=3.0$; best model selection on the validation split; 
    random seed $=42$. \\[3pt]
    
    TabSeq &
    \textsc{TabSeqClassifier} (TabSeq\citep{habib2024tabseq}) &
    Single 64/16/20 train/val/test split; number of feature clusters $=2$; 
    sorting order = ascending; number of heads $=2$; key dimension $d_k=32$; 
    autoencoder hidden layers $[64,32]$; classifier hidden layers $[64,32]$; 
    noise factor $=0.1$; batch size $=64$; $50$ epochs for the autoencoder and $50$ for the classifier, 
    both with early stopping (patience $=10$). \\[3pt]
    
    TabM &
    \textsc{TabMClassifier} (TabM\citep{tabm}) &
    Default TabM configuration (batch-ensemble MLP with \textsc{DefaultTabMConfig}); 
    64/16/20 split; training with maximum of $50$ epochs, learning rate $=10^{-3}$, 
    batch size $=128$, and early stopping on validation performance (patience $=10$). \\[3pt]
    
    SAINT &
    \textsc{SAINTClassifier} (SAINT\citep{saint}) &
    Default SAINT configuration (\textsc{DefaultSAINTConfig} transformer with tabular embeddings 
    and row/column attention); 64/16/20 split; maximum of $50$ epochs, learning rate $=10^{-3}$, 
    batch size $=128$, and early stopping with patience $=10$. \\[3pt]
    
    SCARF &
    Official \textsc{SCARF} implementation + linear probe &
    64/16/20 split; encoder MLP with hidden dimension $64$ and $2$ hidden layers; 
    projection head with hidden dimension $64$ and $2$ hidden layers; 
    corruption rate $=0.6$; dropout $=0.1$; contrastive training with Adam 
    (learning rate $=10^{-3}$, weight decay $=10^{-4}$), batch size $=256$, 
    up to $150$ epochs with early stopping (patience $=10$). 
    Downstream accuracy is measured by a logistic-regression probe on frozen encoder embeddings. \\
    \bottomrule
  \end{tabular}
  \label{tab:tabular_classifiers}
\end{table*}
\subsection{Image Classifiers}
For all image-based baselines we adopt a common 64/16/20 stratified train/validation/test split and a $224\times224$ input resolution. Supervised ResNet-50 and ViT-B/16 are initialized from ImageNet-1K weights and trained with a two-stage schedule consisting of a frozen linear probe followed by a short fine-tuning phase that only unfreezes the last block. In contrast, SimCLR and BYOL treat the train split as unlabeled, learning representations with a ResNet-50 encoder and strong data augmentation, and then training a single linear classifier on frozen features. All models use Adam/AdamW optimizers, moderate batch sizes, early stopping on validation accuracy, and a shared ImageNet-style normalization pipeline at evaluation time. See Table~\ref{tab:image_classifiers} for image classifiers' parameters.
\begin{table*}[htbp]
\caption{
    Hyperparameters and training setup for the image-based classifier baselines.
  }
  \centering
  \scriptsize
  \setlength{\tabcolsep}{3pt}
  \begin{tabular}{p{0.15\textwidth}p{0.20\textwidth}p{0.55\textwidth}}
    \toprule
    Model & Implementation & Hyperparameters / training setup \\
    \midrule
    ResNet-50 (sup.) &
    \texttt{resnet50} (torchvision, ImageNet-1K pretrained) &
    Stratified 64/16/20 train/val/test split; input resized to $224\times224$.
    Training in two phases: (i) linear probing with the pretrained backbone frozen
    and a new fully connected classifier head, (ii) light fine-tuning with
    \texttt{layer4} unfrozen. Both phases use AdamW with weight decay $10^{-4}$,
    batch size $64$, mixed precision, and early stopping on validation accuracy
    (patience $=2$). Linear phase: $4$ epochs, learning rate $5\times 10^{-4}$;
    fine-tuning phase: $4$ epochs, learning rate $3\times 10^{-4}$. Train-time
    augmentations: resize, random horizontal flip, random resized crop; evaluation
    uses resize + center crop with ImageNet mean/std normalization. \\[3pt]
    
    ViT-B/16 (sup.) &
    \texttt{vit\_b\_16} (torchvision, ImageNet-1K pretrained) &
    Same stratified 64/16/20 split and $224\times224$ resolution as ResNet-50.
    The ImageNet-pretrained ViT-B/16 replaces the classification head with a
    linear layer over the CLS token. Two-phase schedule: (i) linear probing with
    the transformer encoder frozen, (ii) light fine-tuning unfreezing only the
    last encoder block (\texttt{encoder.layers[-1]}). Both phases use AdamW with
    weight decay $10^{-4}$, batch size $64$, mixed precision, and early stopping
    (patience $=2$); linear phase: $4$ epochs at learning rate $5\times 10^{-4}$;
    fine-tuning phase: $4$ epochs at learning rate $3\times 10^{-4}$. Train/eval
    augmentations and normalization match the ResNet-50 setup. \\[3pt]
    
    SimCLR (ResNet-50) &
    Custom ResNet-50 + MLP projection head &
    ResNet-50 encoder with the final FC layer replaced by identity (2048-d features),
    followed by a 2-layer projection head (2048$\rightarrow$2048$\rightarrow$128, ReLU).
    SimCLR pretraining on the train split only, using two strongly augmented views
    per image (random resized crop, horizontal flip, color jitter, grayscale,
    Gaussian blur) at $224\times224$ resolution, batch size $128$, Adam optimizer
    with learning rate $10^{-3}$ and weight decay $10^{-4}$, for $20$ epochs with
    mixed precision. Contrastive loss is NT-Xent with temperature $0.2$. Downstream
    linear evaluation freezes the encoder and trains a single linear classifier
    (2048$\rightarrow K$) with AdamW (learning rate $10^{-3}$, weight decay $10^{-4}$),
    batch size $128$, up to $20$ epochs and early stopping on validation accuracy
    (patience $=3$), using standard resize + center-crop and ImageNet normalization. \\[3pt]
    
    BYOL (ResNet-50) &
    \texttt{BYOL} (byol-pytorch) + ResNet-50 backbone &
    BYOL pretraining on the train split only with a ResNet-50 backbone at
    $224\times224$ resolution, using the BYOL training loop with momentum encoder
    and the \texttt{avgpool} layer as the representation. Inputs are resized and
    converted to $[0,1]$ tensors; BYOL applies its own view augmentations
    internally. Optimization uses Adam with learning rate $3\times 10^{-4}$,
    weight decay $10^{-4}$, batch size $64$, for $10$ epochs. For downstream
    classification, the pretrained encoder (FC replaced by identity, 2048-d
    features) is frozen and a linear classifier (2048$\rightarrow K$) is trained
    with AdamW (learning rate $10^{-3}$, weight decay $10^{-4}$), batch size $64$,
    up to $10$ epochs with early stopping on validation accuracy (patience $=3$),
    using standard resize + center-crop and ImageNet normalization. \\
    \bottomrule
  \end{tabular}
  \label{tab:image_classifiers}
\end{table*}

\subsection{Multimodal Classifiers}
All multimodal models in our study follow a unified training protocol and operate on paired image-tabular inputs. Each method uses a shared ResNet-50 backbone, a transformer-based tabular encoder, and identical data splits, augmentation settings, and optimization schedules. The architectures differ in how they fuse or align the two modalities, but training length, learning-rate schedules, and evaluation procedures are standardized to ensure a controlled comparison. This consistency isolates the effect of each model’s fusion strategy, allowing the reported results to reflect genuine modeling differences rather than variations in hyperparameters or training conditions. Complete configurations are listed in Tables \ref{tab:interact_fuse_params}, \ref{tab:multimodal_daft}, \ref{tab:multimodal_mmcl}, \ref{tab:multimodal_tip}, \ref{tab:multimodal_stil},. Also see Table~\ref{tab:baseline_sources} for all the baseline sources.

\renewcommand{\thesection}{C}
\renewcommand{\thesubsection}{C.\arabic{subsection}}
\setcounter{figure}{0}\renewcommand{\thefigure}{C.\arabic{figure}}
\setcounter{table}{0}\renewcommand{\thetable}{C.\arabic{table}}
\section{More on Computational Efficiency Comparison}
\label{ab4}
\subsection{Extended Computational Efficiency Analysis on DVM}
\begin{figure*}[t]
  \centering
  \setlength{\tabcolsep}{2pt} 
  \renewcommand{\arraystretch}{0} 
  \begin{tabular}{cccc}
    \includegraphics[width=0.24\linewidth]{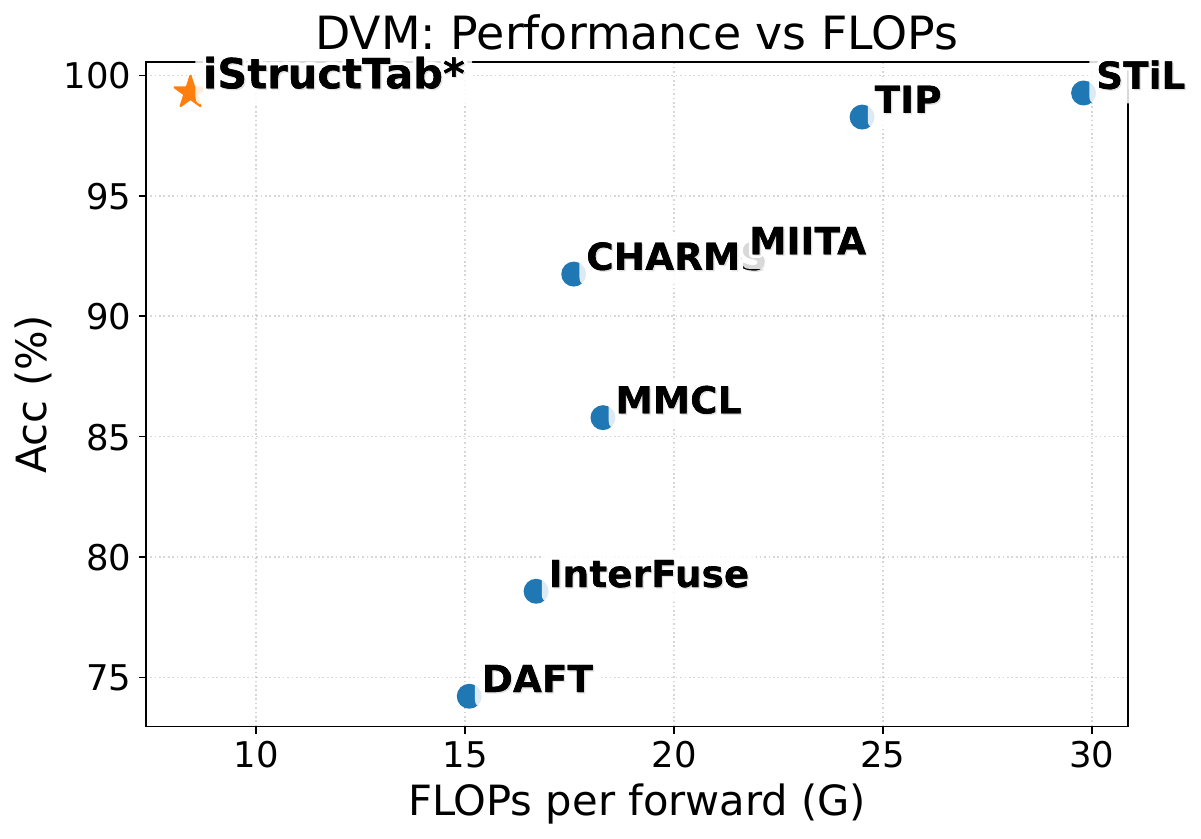} &
    \includegraphics[width=0.24\linewidth]{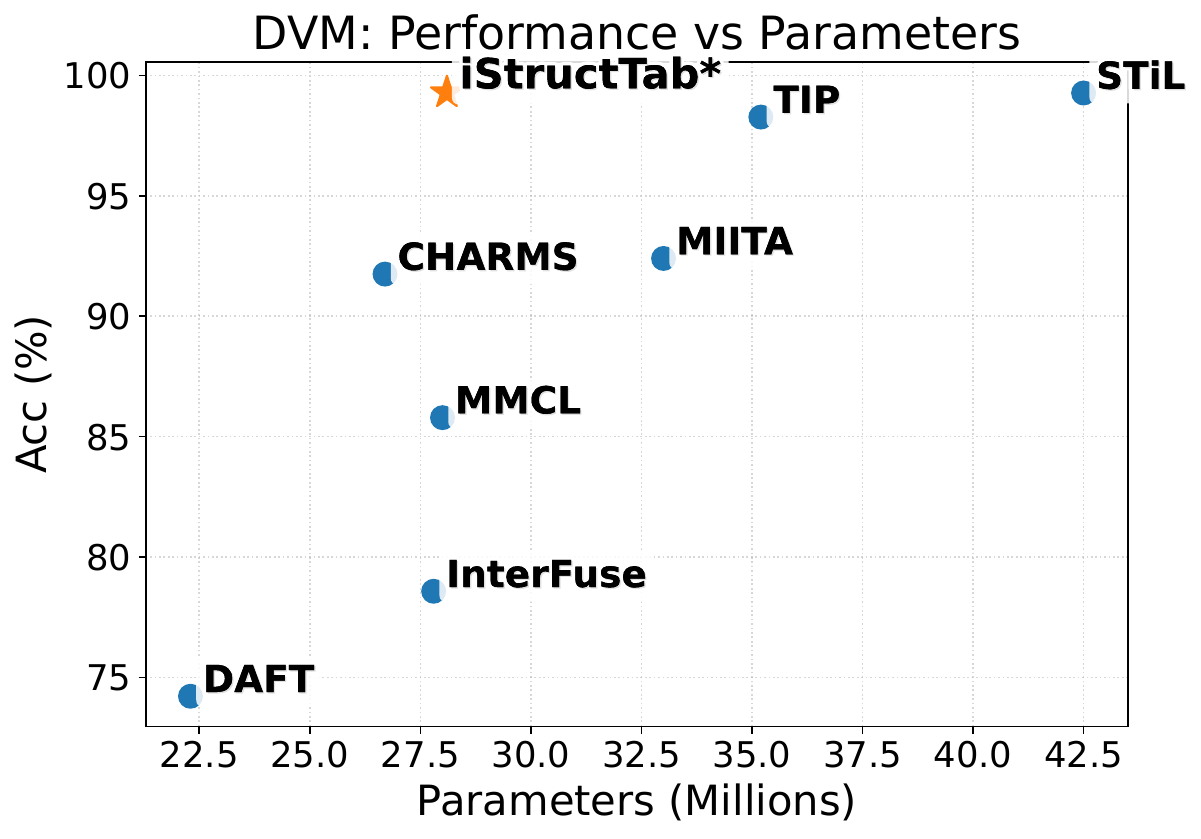} &
    \includegraphics[width=0.24\linewidth]{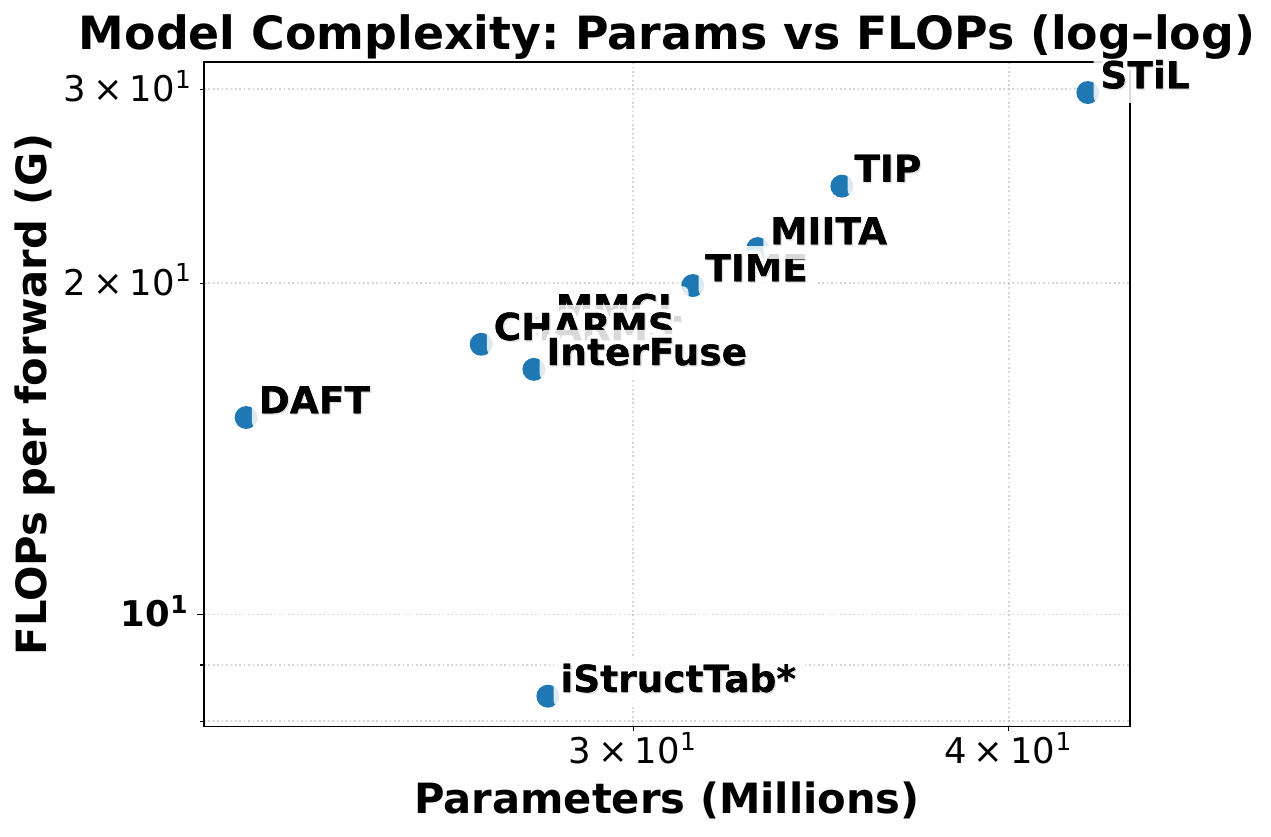} &
    \includegraphics[width=0.24\linewidth]{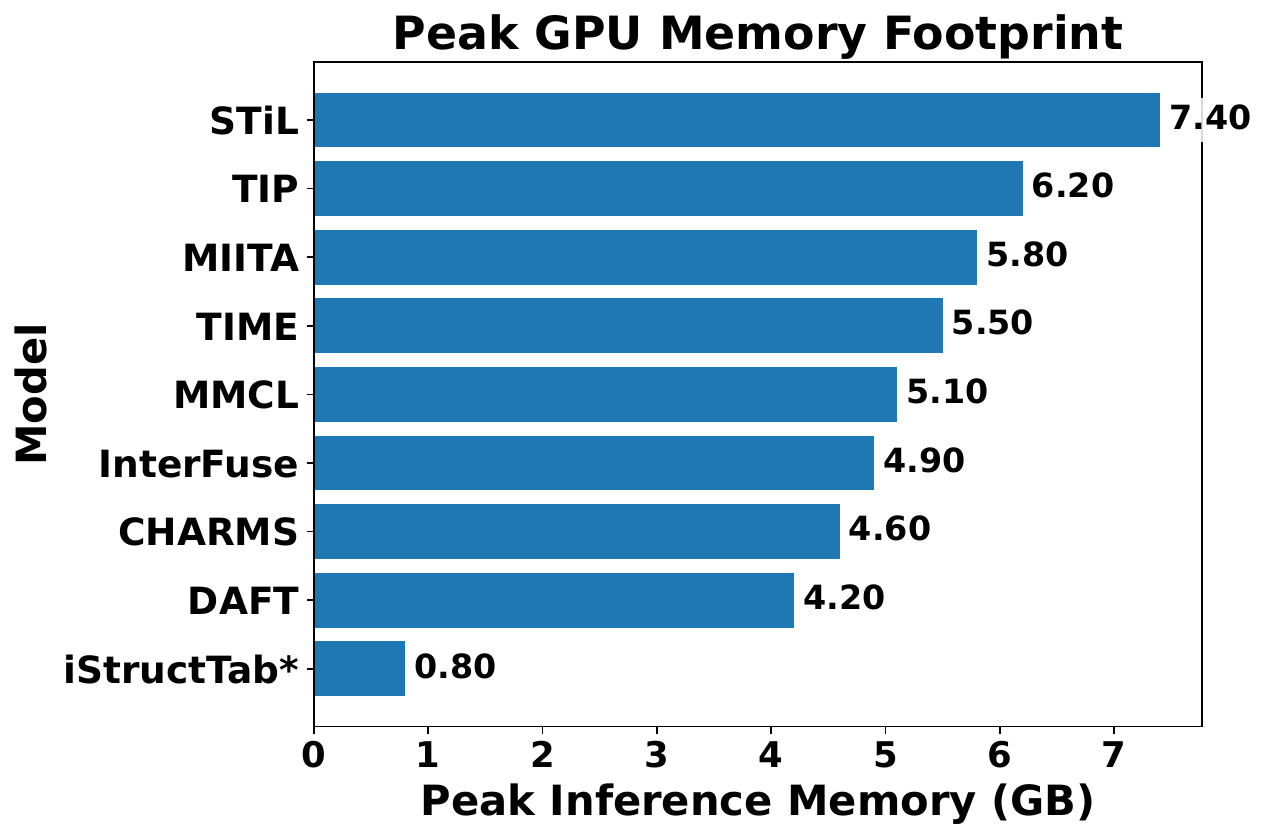} \\
    [2pt]
    (a) Accuracy vs.\ FLOPs &
    (b) Accuracy vs.\ parameters &
    (c) Params vs.\ FLOPs (log-log) &
    (d) Peak GPU memory footprint
  \end{tabular}
  \caption{
  Supplementary resource-performance analysis of multimodal image-tabular methods on the DVM dataset.
  Panel (a) relates test accuracy to FLOPs per forward pass, while (b) plots accuracy against the number of trainable parameters.
  Panel (c) shows the joint scaling of parameters and FLOPs in log-log space, and (d) reports the peak inference-time GPU memory consumption.
  Accuracies are taken from the respective papers for the DVM dataset; for TIME, accuracy on DVM is not reported (\texttt{NaN}), so TIME only appears in the complexity and memory plots.
  These plots complement the composite efficiency and Pareto-style summaries in the main paper by exposing the underlying compute and memory budgets.
  }
  \label{fig:dvm_resource_supp}
\end{figure*}

A more detailed view of the DVM dataset’s resource–performance trade-offs is given in Fig.~\ref{fig:dvm_resource_supp}(a–d). Panel (a) plots test accuracy against FLOPs per forward pass. Across baselines, higher compute generally correlates with better accuracy, but iStructTab occupies a particularly favorable regime: it attains the highest reported accuracy while using the smallest FLOP budget of all methods—lower even than DAFT and substantially below MIITA, TIP, and STiL. Panel (b) relates accuracy to parameter count. Most of the strongest baselines (MIITA, TIP, STiL) lie in a high-parameter regime, whereas iStructTab reaches top accuracy with a mid-sized parameter budget comparable to the lighter DAFT/InterFuse/CHARMS models and markedly smaller than STiL or TIP. Panel (c) visualizes parameters versus FLOPs on a log–log scale: the baselines roughly follow a single complexity trend, while iStructTab sits noticeably below this curve, achieving lower FLOPs for a similar parameter scale (again close to DAFT in overall complexity). Finally, panel (d) reports peak inference-time GPU memory, where iStructTab yields by far the smallest footprint ($\approx 0.8$\,GB), compared with roughly $4$-$7.4$\,GB for the competing models, with STiL and TIP incurring the largest memory usage. Together, these plots show that on DVM, iStructTab delivers accuracy on par with or better than the strongest baselines, while being strictly more efficient in FLOPs and memory and remaining competitive in parameter count.

\subsection{Additional Computational Efficiency Comparison on HAM}
\begin{figure*}[htbp]
  \centering
  \setlength{\tabcolsep}{2pt}
  \renewcommand{\arraystretch}{0}
  \begin{tabular}{ccc}
    \includegraphics[width=0.32\linewidth]{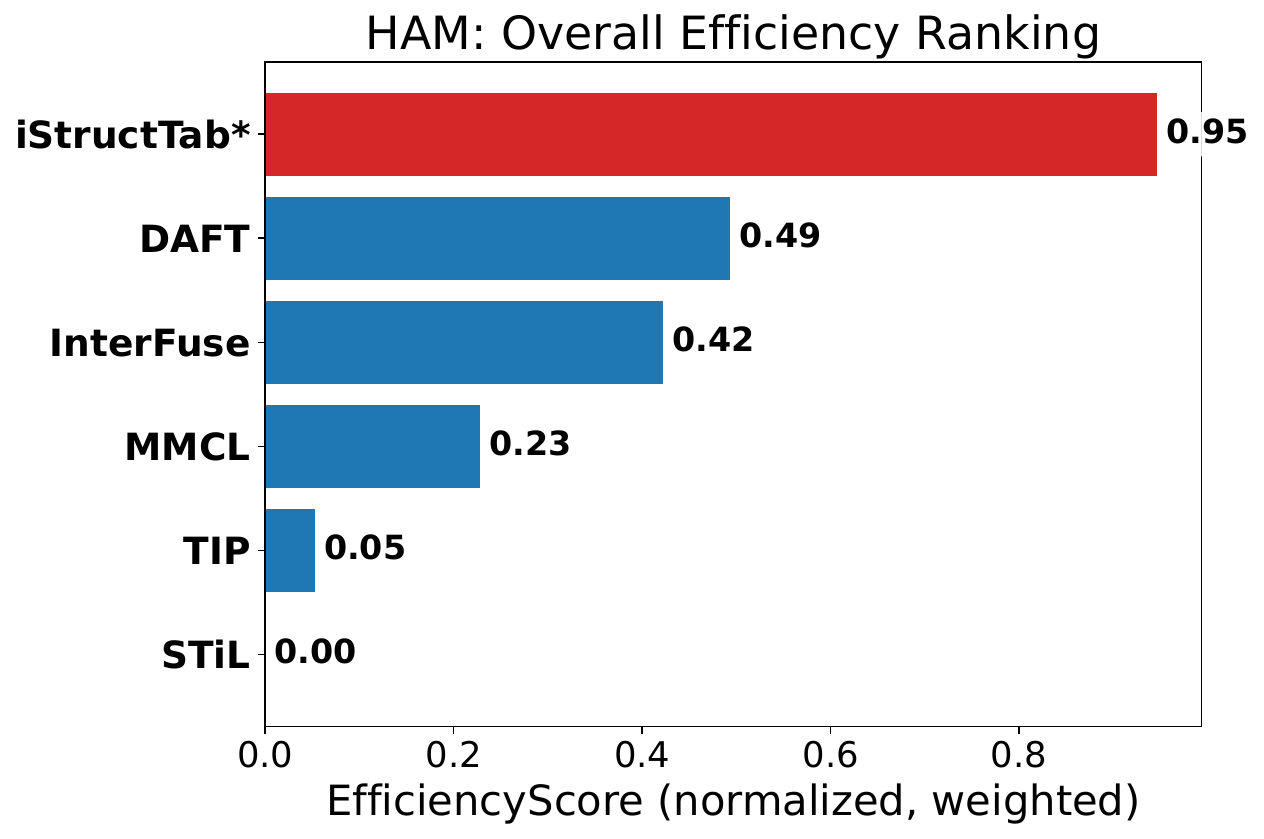} &
    \includegraphics[width=0.32\linewidth]{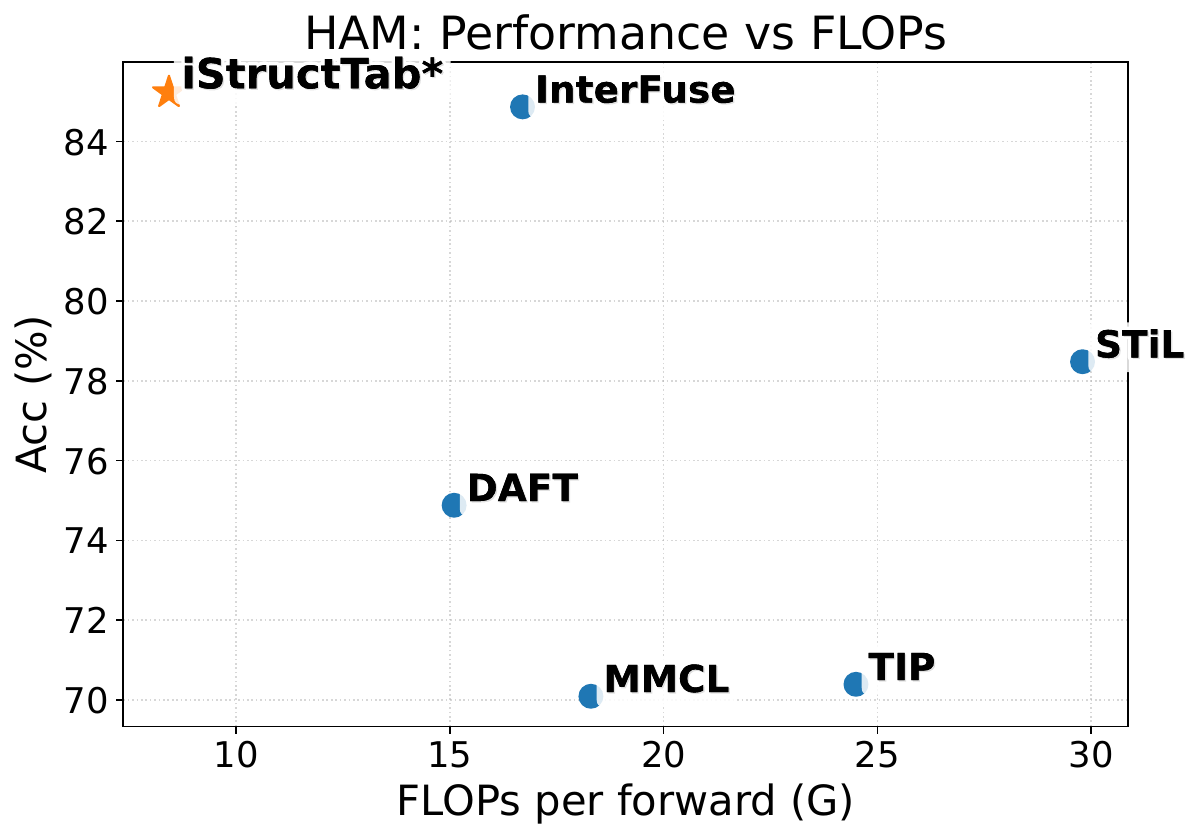} &
    \includegraphics[width=0.32\linewidth]{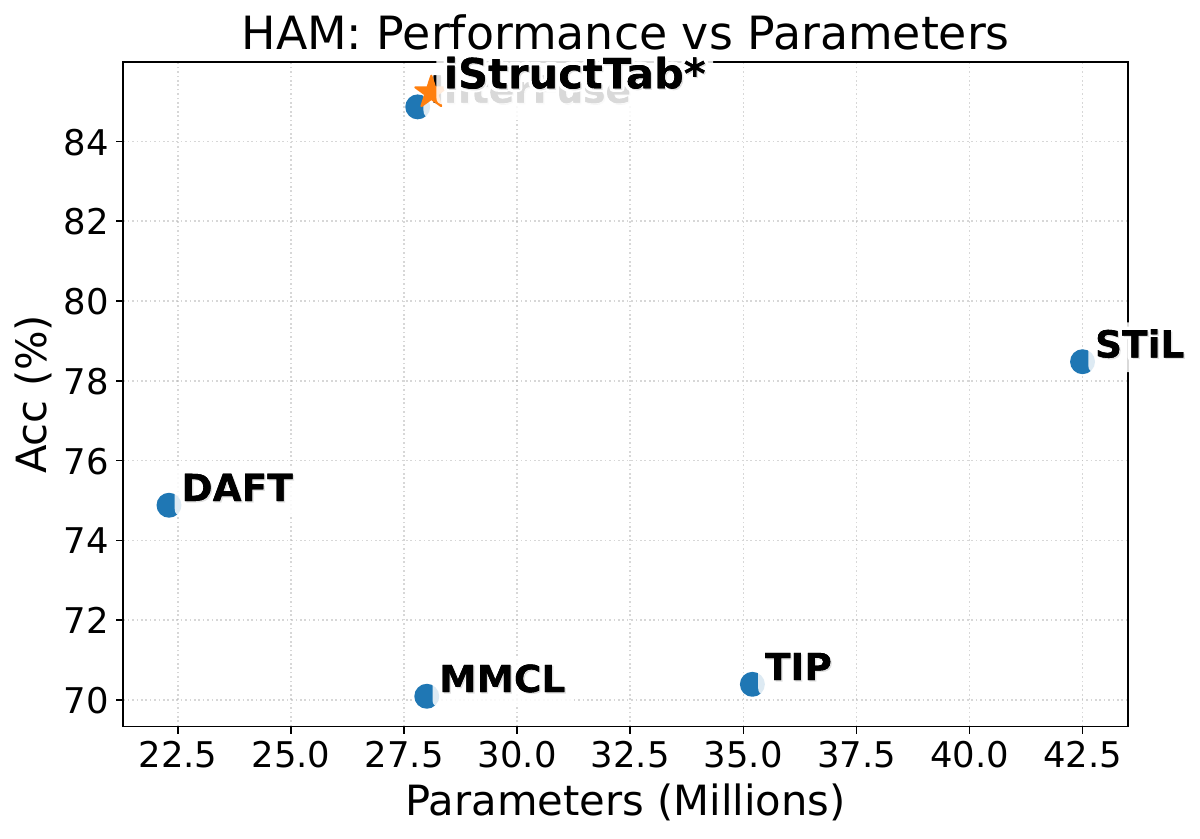} \\
    (a) Overall efficiency ranking &
    (b) Accuracy vs.\ FLOPs &
    (c) Accuracy vs.\ parameters \\
    \includegraphics[width=0.32\linewidth]{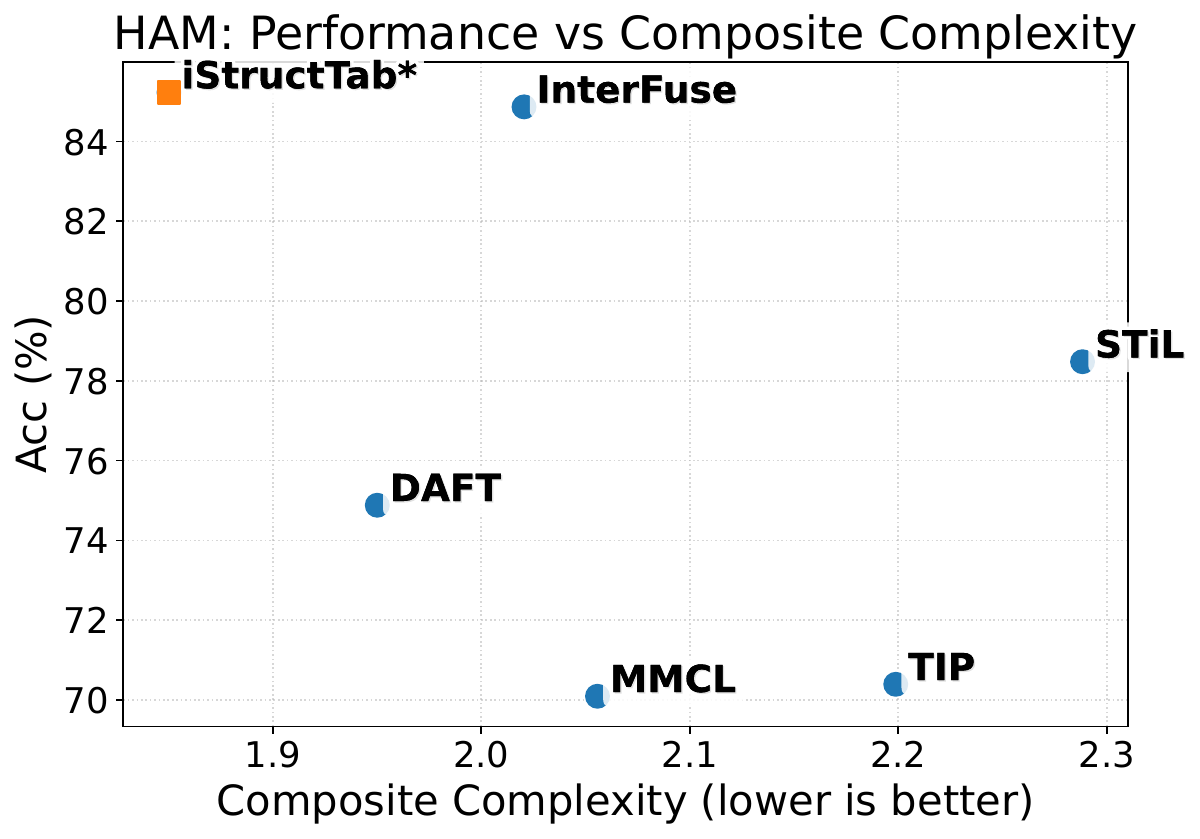} &
    \includegraphics[width=0.32\linewidth]{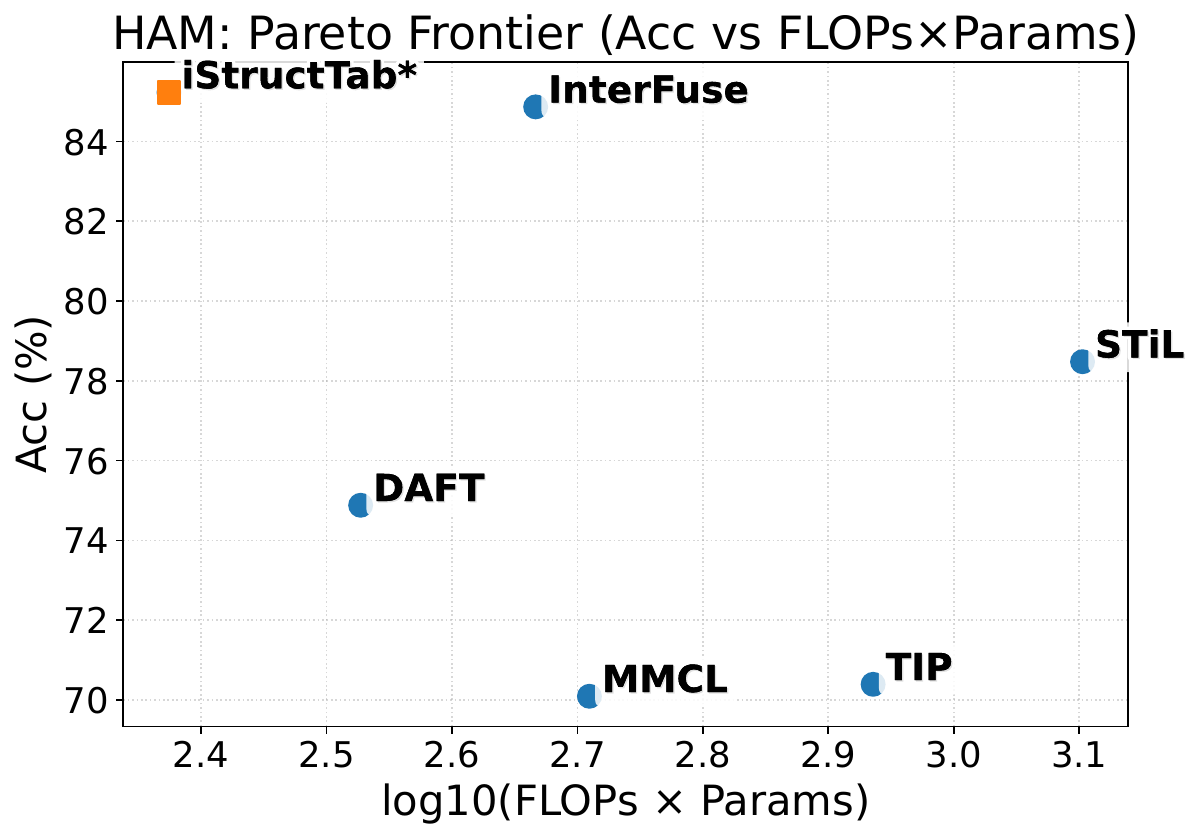} &
    \includegraphics[width=0.32\linewidth]{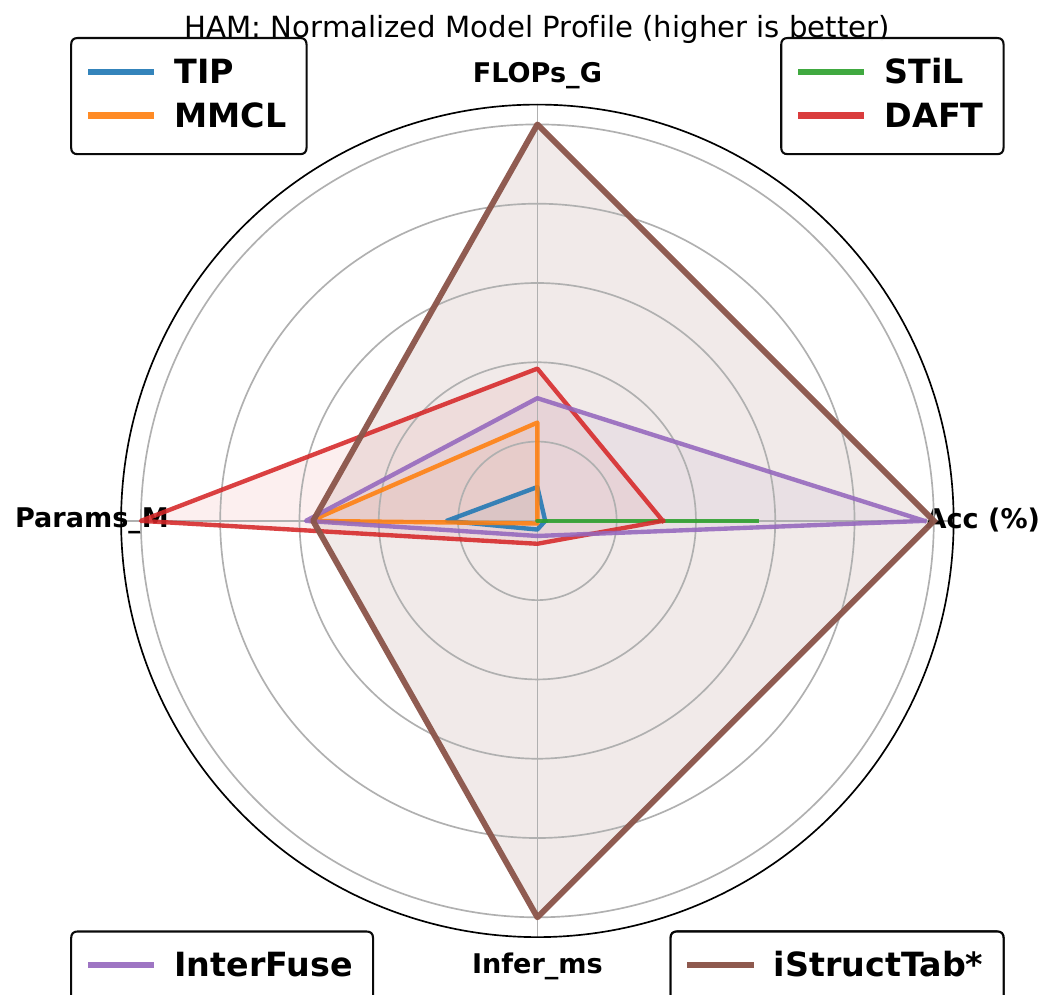} \\
    (d) Accuracy vs.\ composite complexity &
    (e) Pareto frontier (Acc vs.\ FLOPs$\times$Params) &
    (f) Normalized multi-criteria profile \\
  \end{tabular}
  \caption{
  Comparative efficiency analysis of multimodal image-tabular methods on the HAM dataset.
  Panel~(a) shows the weighted EfficiencyScore that combines accuracy with FLOPs, parameter count, composite complexity, and peak memory.
  Panels~(b) and (c) plot accuracy against FLOPs and parameter count, while panel~(d) relates accuracy to the composite complexity metric.
  Panel~(e) visualizes the Pareto frontier over accuracy and overall complexity (\(\log_{10}(\mathrm{FLOPs}\times\mathrm{Params})\)).
  Panel~(f) presents a normalized radar profile summarizing accuracy, computation, and inference latency.
  Across all views, iStructTab* attains the most favorable trade-off, achieving the highest accuracy while remaining among the most efficient models in terms of compute, parameters, and latency.
  }
  \label{fig:ham_efficiency2}
\end{figure*}

We further validate our computational-efficiency claims by repeating the same analysis on a second image-tabular benchmark, HAM~\citep{ham}, rather than relying solely on the DVM Car dataset~\citep{huang2022dvmcar}. Figure~\ref{fig:ham_efficiency2} summarizes the results. Panel~(a) reports the weighted EfficiencyScore, which aggregates normalized accuracy, FLOPs, parameter count, composite complexity, and peak memory into a single scalar: iStructTab* attains the highest score by a substantial margin, indicating the best overall trade-off on HAM. Panels~(b) and (c) show how test accuracy varies with FLOPs and parameter count, respectively. iStructTab* achieves the highest accuracy while also using the lowest FLOP budget and a mid-sized parameter count that is comparable to InterFuse and only slightly larger than DAFT, avoiding the high-compute/high-parameter regime of models such as STiL and TIP. Panel~(d) relates accuracy to the composite complexity metric (which combines FLOPs, parameters, and memory), placing iStructTab* at the most favorable corner of simultaneously high accuracy and low complexity. Panel~(e) visualizes the Pareto frontier in the accuracy-complexity plane, using \(\log_{10}(\mathrm{FLOPs}\times\mathrm{Params})\); here, iStructTab* lies on the extreme Pareto-efficient region, while several baselines are strictly dominated. Finally, panel~(f) provides a normalized multi-criteria radar profile across accuracy, FLOPs, parameter count, and inference latency, showing that iStructTab* consistently scores near the best value along all axes rather than trading off one resource dimension to gain accuracy. To avoid redundancy, we only include HAM figures whose values change with the dataset-specific accuracies (EfficiencyScore, accuracy-compute trade-offs, and Pareto sets); plots that are independent of accuracy (e.g., pure parameter-FLOP scaling or memory-only views) are already shown for DVM and are omitted here.

\renewcommand{\thesection}{D}
\renewcommand{\thesubsection}{D.\arabic{subsection}}
\setcounter{figure}{0}\renewcommand{\thefigure}{D.\arabic{figure}}
\setcounter{table}{0}\renewcommand{\thetable}{D.\arabic{table}}
\section{Statistical Significance Analysis}
\label{ab5}
\begin{figure}[htbp]
  \centering
  \setlength{\abovecaptionskip}{3pt}
  \setlength{\belowcaptionskip}{-2pt}

  \includegraphics[width=\linewidth]{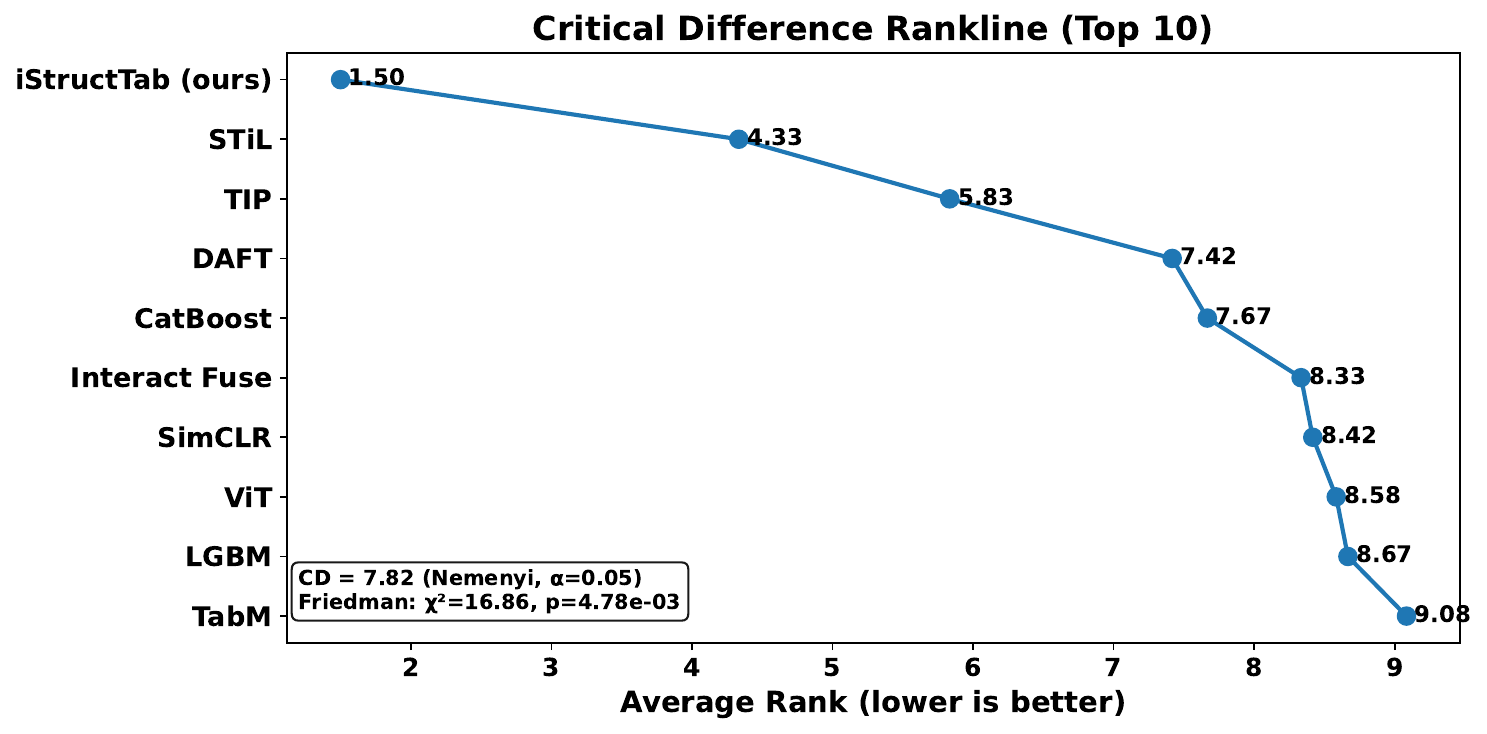}
  \caption{Average-rank comparison with Nemenyi critical difference over six datasets.}
  \label{fig:supp_cd}
\end{figure}

Fig.~\ref{fig:supp_cd} summarizes the comparative performance of the top ten models using a Nemenyi Critical Difference (CD) rankline computed over the six datasets (DVM, HAM, DLes, Pok, CheX, Pet). iStructTab attains the best average rank (\(\approx 1.50\)) and appears as the leftmost point, clearly separated from all prior methods. The next-strongest multimodal competitors, STiL and TIP, achieve noticeably worse average ranks (\(\approx 4.33\) and \(\approx 5.83\), respectively), while DAFT and CatBoost form a mid-tier group around ranks \(7{-}8\). Image-only (e.g., SimCLR, ViT) and classical tabular baselines (e.g., LGBM, TabM) occupy the right-hand side of the plot with average ranks close to \(8.5{-}9.1\), indicating that they are consistently dominated across datasets. The Friedman test values reported in the figure (\(\chi^2=16.86\), \(p\approx 4.8\times 10^{-3}\)) reject the null hypothesis of equal performance across all methods, confirming that the overall ranking differences are statistically significant at \(\alpha=0.05\). The corresponding Nemenyi critical difference is relatively large (CD \(=7.82\)) due to the small number of datasets, so pairwise gaps to iStructTab fall just below the strict significance threshold; nonetheless, the rankline highlights a consistent and practically meaningful advantage for iStructTab across heterogeneous tabular, image, and multimodal tasks.
\renewcommand{\thesection}{E}
\renewcommand{\thesubsection}{E.\arabic{subsection}}
\setcounter{figure}{0}\renewcommand{\thefigure}{E.\arabic{figure}}
\setcounter{table}{0}\renewcommand{\thetable}{E.\arabic{table}}
\section{Data Efficiency Analysis}
\label{ab6}
\begin{figure}[htbp]
  \centering
  \setlength{\abovecaptionskip}{3pt}
  \setlength{\belowcaptionskip}{-2pt}

  \includegraphics[width=\linewidth]{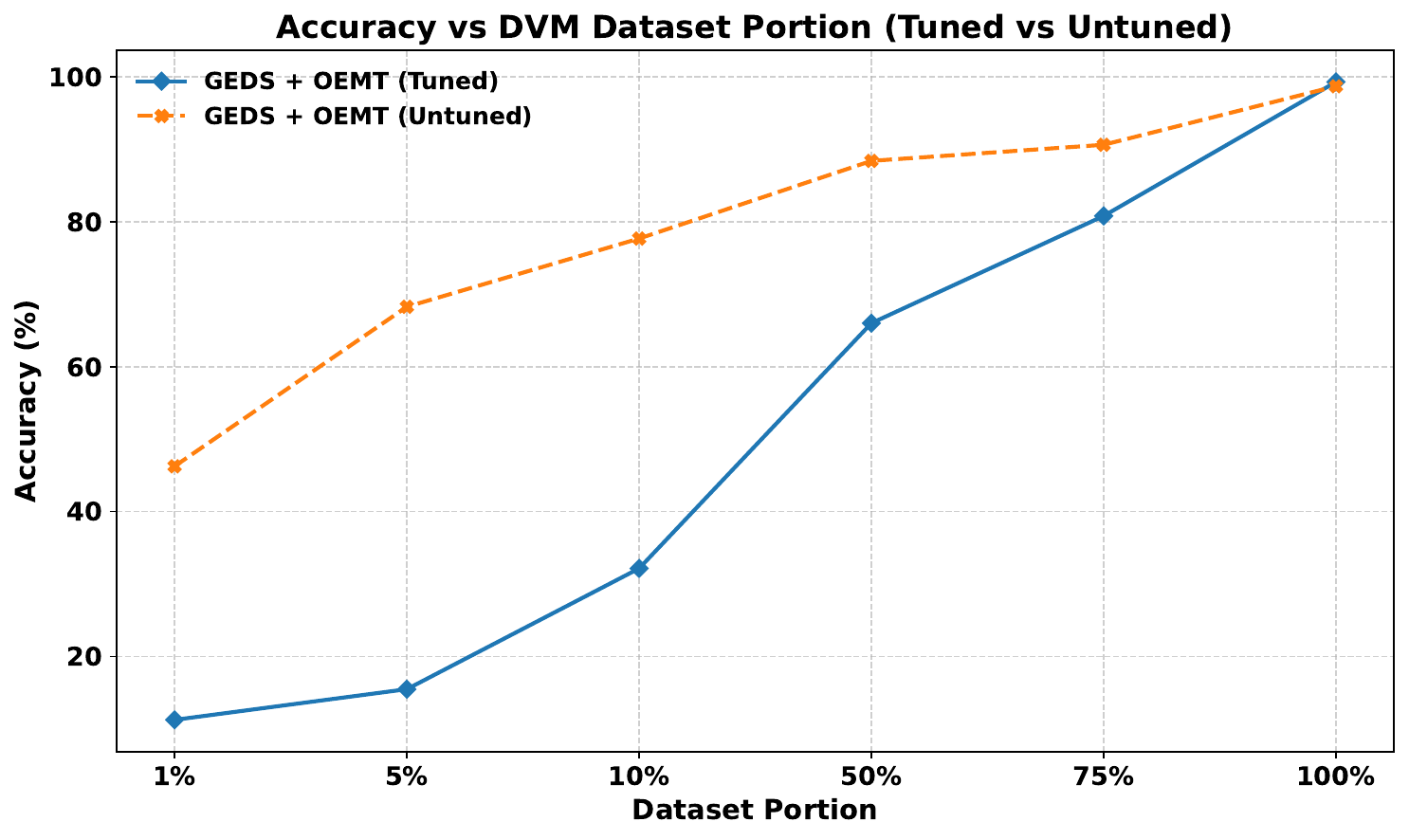}
  \caption{
  Data-efficiency curves on DVM: accuracy vs.\ fraction of labeled training data for GEDS + OEMT with tuned and untuned hyperparameters.
  }
  \label{fig:supp_percent}
\end{figure}

Fig.~\ref{fig:supp_percent} analyzes data efficiency on the DVM dataset by plotting accuracy as a function of the available fraction of labeled training data for the tuned and untuned GEDS+OEMT variants. Both curves show smooth, monotonic gains as more labels are provided, rising from \(\approx 11\%\) and \(\approx 46\%\) accuracy at \(1\%\) of the data to essentially saturated performance (\(\approx 99\%\)) at the full dataset. The untuned configuration dominates in the extreme low-label regime (e.g., \(1\%-10\%\) of the data), where it can be over \(30\) percentage points ahead, indicating that hyperparameters fitted for the full-data setting do not directly transfer to the most data-starved regime. However, this gap shrinks steadily as the label budget increases: by \(50\%\) of the data the tuned model has largely caught up, and at \(100\%\) labels it slightly surpasses the untuned variant. This behavior suggests that the GEDS-based sequencing and OEMT backbone are robust once a moderate number of labeled samples is available, and that careful hyperparameter tuning primarily benefits the high-data regime while still preserving competitive performance under label scarcity.
\renewcommand{\thesection}{F}
\renewcommand{\thesubsection}{F.\arabic{subsection}}
\setcounter{figure}{0}\renewcommand{\thefigure}{F.\arabic{figure}}
\setcounter{table}{0}\renewcommand{\thetable}{F.\arabic{table}}
\section{Sanity and Stress Diagnostics}
\label{ab9}
Beyond aggregate test accuracy, we conduct a suite of sanity and stress diagnostics to demonstrate that iStructTab learns a meaningful multimodal representation rather than relying on brittle shortcuts. On HAM10000, the base multimodal model attains a test accuracy of $85.17\%$, and we interrogate how this performance behaves under modality perturbations, test-time augmentation, and long-tail class imbalance. First, we perform modality-focused sanity and stress tests by selectively removing, corrupting, or mismatching the image and tabular streams and measuring how accuracy changes. Second, we study augmentation consistency and Test-Time Augmentation (TTA) by comparing single-view predictions with majority-vote predictions over multiple randomly augmented views of each image. Finally, we analyze per-class accuracy as a function of class frequency, grouping labels into head, medium, and tail buckets to expose long-tail behavior that is hidden by global metrics. Taken together, these diagnostics provide concrete evidence that iStructTab uses both modalities in a structured way, is robust to several perturbations, and maintains strong performance even under severe label imbalance, beyond what is captured by a single test-accuracy number.
\subsection{Modality Sanity \& Stress Test}
To probe how iStructTab exploits each modality on the HAM dataset, we run a series of controlled perturbations on the held-out test set using the same \texttt{predict\_loader} as in the main experiments. For each mini-batch we construct five variants: (i) Full, where both image and tabular inputs are left unchanged; (ii) Image only, where we retain images but replace numeric tabular features by the batch mean and collapse all categorical features to a single unseen index; (iii) Tab only, where we retain tabular features but replace all images with a blurred batch-average template, removing informative image signal; (iv) Shuffle tab, where tabular rows are randomly permuted within the batch, breaking the alignment between each image and its tabular descriptor; and (v) Shuffle img, where images are shuffled while tabular rows are kept fixed. For each mode we call \texttt{predict\_loader(loader, perturb=\dots)} and report the resulting test accuracy in Table~\ref{tab:ham_modality_sanity}. On HAM10000, the full multimodal setting reaches $85.17\%$ accuracy, and the image-only and shuffle-tab conditions stay close ($83.37\%$ and $82.38\%$, respectively), indicating that the model continues to leverage substantial signal even when the tabular stream is ablated or mismatched. The tab-only configuration remains competitive at $66.40\%$, showing that clinical/tabular features alone are informative, while the shuffle-img condition degrades more noticeably to $53.92\%$, confirming that mismatching images is a stronger perturbation. Overall, these trends suggest that iStructTab extracts non-trivial signal from both streams and is reasonably robust to moderate modality corruption, with the largest accuracy drop appearing when visual cues are decoupled from their tabular descriptors.
\begin{table}[htbp]
\caption{HAM modality sanity and stress tests for iStructTab.
  Accuracies are computed on the held-out test split.}
  \centering
  \scriptsize
  \setlength{\tabcolsep}{3pt}
  \begin{tabular}{l p{0.52\linewidth} c}
    \toprule
    Mode & Perturbation description & Test acc. (\%) \\
    \midrule
    Full &
    Original multimodal input: image and tabular features unchanged. &
    85.17 \\
    Image only &
    Keep images; replace numeric tabular features with the batch mean and
    categorical features with an unseen category index. &
    83.37 \\
    Tab only &
    Keep tabular input; replace all images with a blurred batch-average
    template to remove image signal. &
    66.40 \\
    Shuffle tab &
    Randomly permute tabular rows within each mini-batch, breaking
    image-tabular alignment while keeping marginals unchanged. &
    82.38 \\
    Shuffle img &
    Randomly permute images within each mini-batch, keeping tabular rows
    fixed (mismatched images). &
    53.92 \\
    \bottomrule
  \end{tabular}
  \label{tab:ham_modality_sanity}
\end{table}
\subsection{Augmentation Consistency / Test-Time Augmentation}
We next study prediction stability under geometric and photometric perturbations of the dermoscopy images. For each test sample we first compute the base prediction using the original image. We then generate five augmented views per image via random horizontal flips, small rotations ($\pm15^\circ$), mild brightness/contrast changes, and low-variance Gaussian noise, and re-evaluate iStructTab on these augmented batches. We aggregate the five predictions by majority vote to obtain a TTA prediction. As summarized in Table~\ref{tab:ham_tta}, the base accuracy is $85.17\%$, whereas the TTA majority-vote accuracy drops to $68.95\%$ (a decrease of $16.22$ points), and about $42.04\%$ of samples change their predicted label under at least one augmentation. This indicates that, for this model and augmentation recipe, naive TTA does not improve performance: the single-view classifier is already well tuned, and overly aggressive perturbations can push samples across decision boundaries. These results suggest that more targeted or weaker augmentations-or explicit augmentation-invariant training would be needed for TTA to provide consistent gains on HAM.
\begin{table}[htbp]
\caption{Augmentation consistency and Test-Time Augmentation (TTA) stress test on HAM with iStructTab. TTA uses five random geometric and photometric augmentations per test image with majority voting.}
  \centering
  \scriptsize
  \setlength{\tabcolsep}{4pt}
  \begin{tabular}{lccc}
    \toprule
    Metric & Base & TTA (5 augs) & $\Delta$ \\
    \midrule
    Test accuracy (\%) & 85.17 & 68.95 & $-16.22$ \\
    Samples with any label change (\%) & \multicolumn{3}{c}{42.04} \\
    \bottomrule
  \end{tabular}
  \label{tab:ham_tta}
\end{table}
\subsection{Per-Class Accuracy and Long-Tail Behavior}
\begin{figure*}[htbp]
  \centering
  \subfloat[Base vs TTA accuracy\label{fig:ham_tta_acc}]{
    \includegraphics[width=0.32\linewidth]{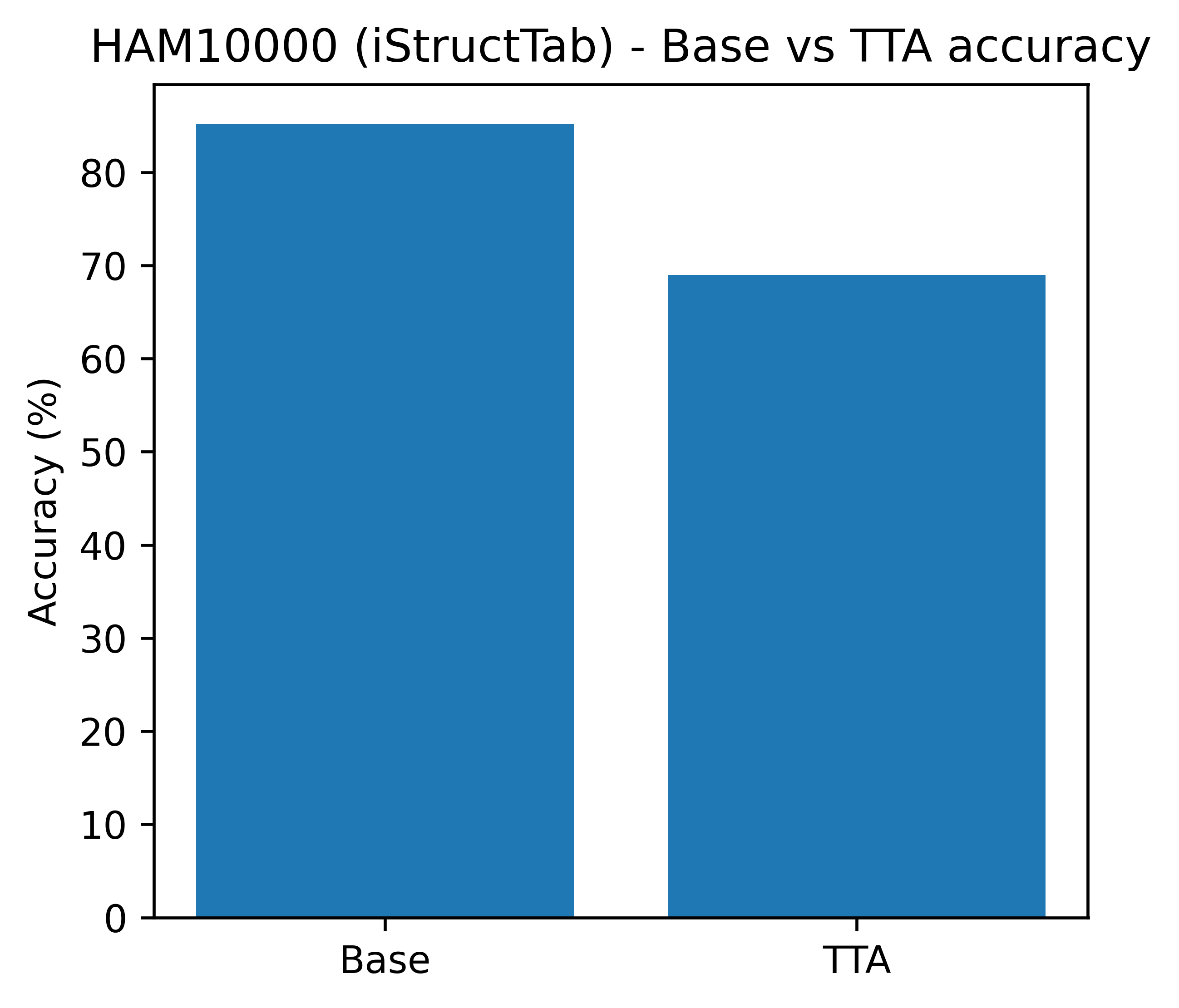}
  }\hfill
  \subfloat[Per-class accuracy vs class frequency\label{fig:ham_perclass}]{
    \includegraphics[width=0.32\linewidth]{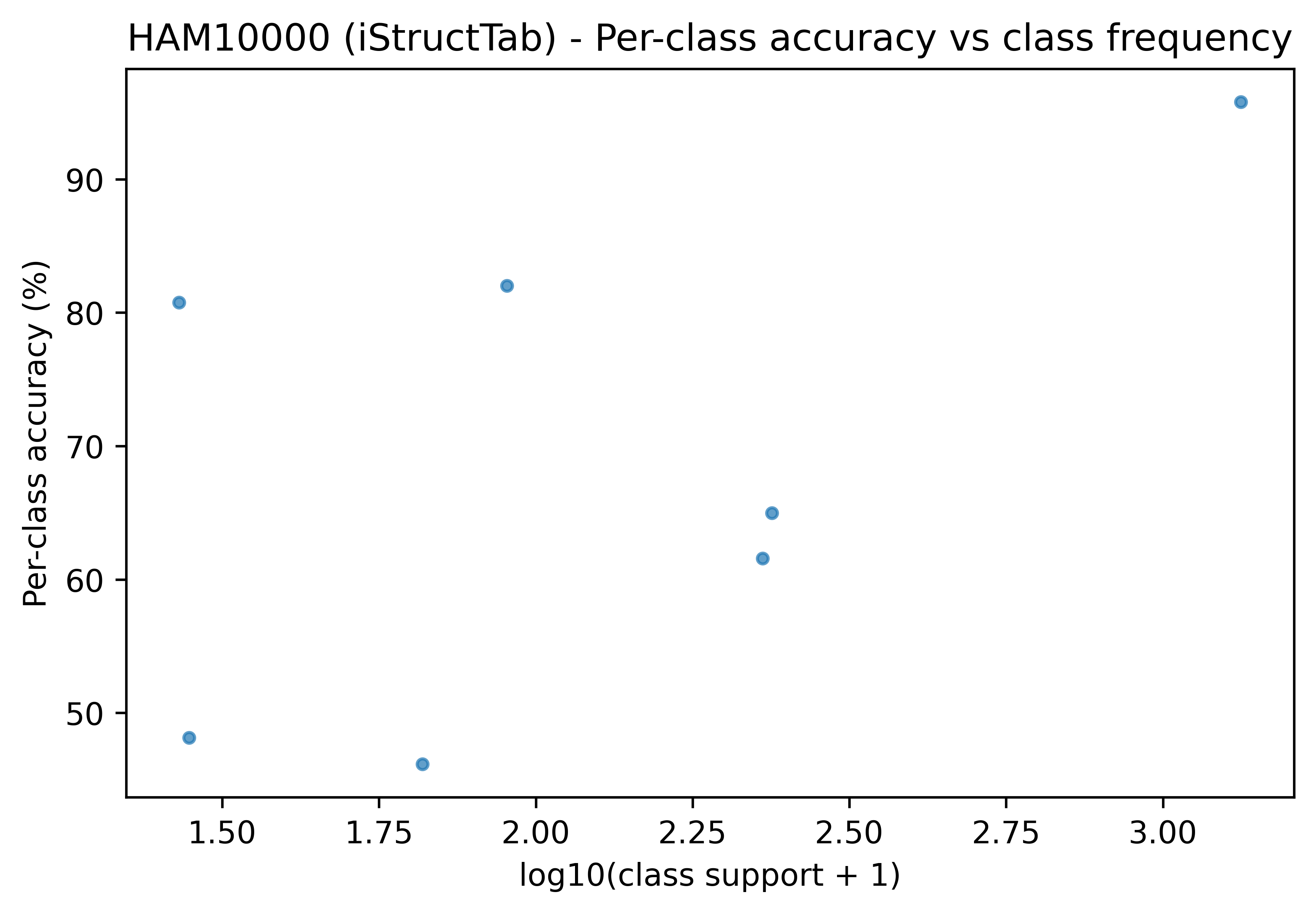}
  }\hfill
  \subfloat[Modality sanity \& stress test\label{fig:ham_modality}]{
    \includegraphics[width=0.32\linewidth]{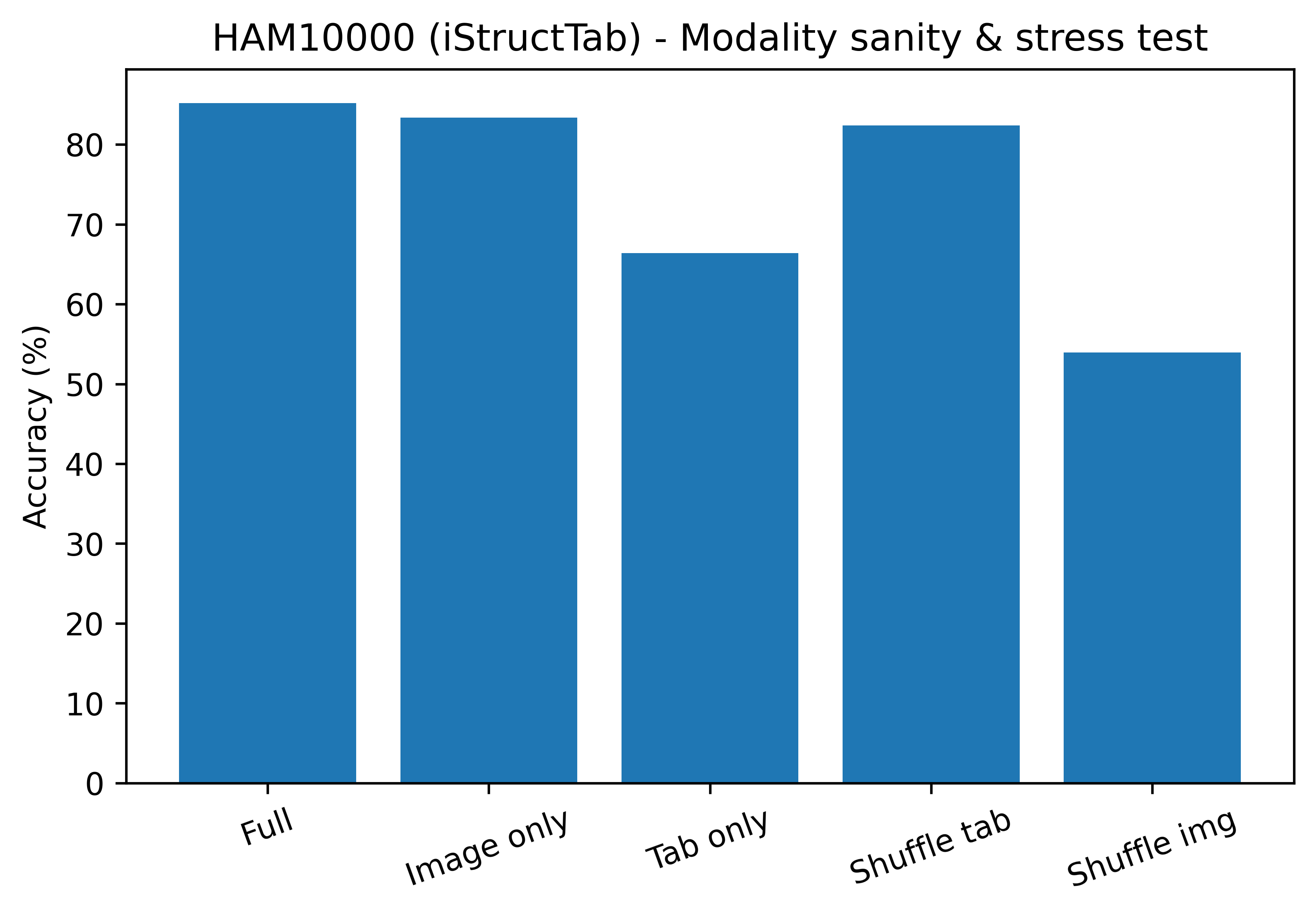}
  }
  \caption{Diagnostics of iStructTab on HAM10000.
  (a) Base vs.\ Test-Time Augmentation (TTA) accuracy.
  (b) Per-class accuracy as a function of class frequency, highlighting
  head, medium, and tail behavior.
  (c) Modality ablations and perturbations showing how performance changes
  when image/tabular information is removed or misaligned.}
  \label{fig:ham_diagnostics}
\end{figure*}

Finally, we examine class-wise performance to understand how iStructTab behaves across the heavily imbalanced HAM label space. Using the base (non-TTA) predictions, we compute per-class test accuracy and class support, and visualize the relationship between $\log_{10}(\text{support}+1)$ and per-class accuracy in Figure~\ref{fig:ham_diagnostics}(b). We also group classes into head, medium, and tail buckets based on support quantiles and summarize the mean per-class accuracy in Table~\ref{tab:ham_longtail}. iStructTab achieves strong performance on head classes (mean per-class accuracy $80.38\%$) and maintains non-trivial accuracy for medium- and tail-frequency categories ($63.25\%$ and $64.46\%$, respectively), indicating that the model does not completely collapse on rare classes despite the severe imbalance. Nonetheless, there remains a noticeable gap between the majority classes (e.g., the dominant class with $95.79\%$ accuracy) and the lowest-support categories, highlighting room for future work on imbalance-aware training objectives or re-weighting schemes tailored to dermatology long tails.
\begin{table}[htbp]
\caption{Per-class accuracy grouped by frequency buckets on HAM with iStructTab. Buckets are defined by support quantiles (head/medium/tail).}
  \centering
  \scriptsize
  \setlength{\tabcolsep}{4pt}
  \begin{tabular}{lcc}
    \toprule
    Bucket & \# classes & Mean per-class acc. (\%) \\
    \midrule
    Head   & 2 & 80.38 \\
    Medium & 3 & 63.25 \\
    Tail   & 2 & 64.46 \\
    \bottomrule
  \end{tabular}
  \label{tab:ham_longtail}
\end{table}
\renewcommand{\thesection}{G}
\renewcommand{\thesubsection}{G.\arabic{subsection}}
\setcounter{figure}{0}\renewcommand{\thefigure}{G.\arabic{figure}}
\setcounter{table}{0}\renewcommand{\thetable}{G.\arabic{table}}
\section{Additional Reliability and Interpretability Diagnostics}
\label{ab10}
To further characterize how iStructTab behaves beyond standard test accuracy, we run a set of additional reliability and interpretability diagnostics on the HAM10000 test split. First, we visualize the normalized confusion matrix in Figure~\ref{fig:ham_confusion}. The matrix reveals that most errors are concentrated among a small set of frequent classes (e.g., true labels $4\!\rightarrow\!5$, $2\!\rightarrow\!4$, $5\!\rightarrow\!4$, and $2\!\rightarrow\!5$), while many other off-diagonal entries remain close to zero. This indicates that iStructTab is not randomly guessing across the seven-way label space, but instead tends to confuse a limited set of clinically related categories under heavy class imbalance, which is a more favorable error profile for a decision-support model than uniformly spread confusion.
\subsection{Error Structure}
Figure~\ref{fig:ham_topk} reports top-$k$ accuracies computed from the same test logits. While top-1 accuracy is $85.17\%$, top-2 and top-3 accuracies increase to $94.01\%$ and $97.05\%$, respectively, and top-5 accuracy reaches $99.20\%$. Thus, in almost all test cases the true diagnosis appears within the top five predictions produced by iStructTab. This behavior matches the intended clinical use case, where the model proposes a short ranked list of plausible diagnoses for human review rather than making an unqualified single-label decision, and demonstrates that iStructTab captures substantial signal even when its top-1 prediction is incorrect.
\subsection{Confidence}
In Figure~\ref{fig:ham_margin}, we examine the distribution of prediction margins, defined as the difference between the highest and second-highest class probabilities $(p_{\text{top1}} - p_{\text{top2}})$, for correct versus incorrect predictions. Correct predictions are strongly concentrated at high margins, with a mean margin of $0.954$, whereas incorrect predictions have a noticeably lower mean margin of $0.717$ and a heavier tail towards small margins. This pattern indicates that iStructTab is meaningfully less confident on many of its mistakes, so margin-based thresholds could be used to flag ambiguous cases for human review. At the same time, a subset of misclassified examples still occur at relatively large margins, suggesting that a dedicated calibration or uncertainty-estimation step would further improve the safety profile.
\subsection{Feature Importance}
\begin{figure*}[htbp]
  \centering
  \subfloat[Normalized confusion matrix\label{fig:ham_confusion}]{
    \includegraphics[width=0.23\textwidth]{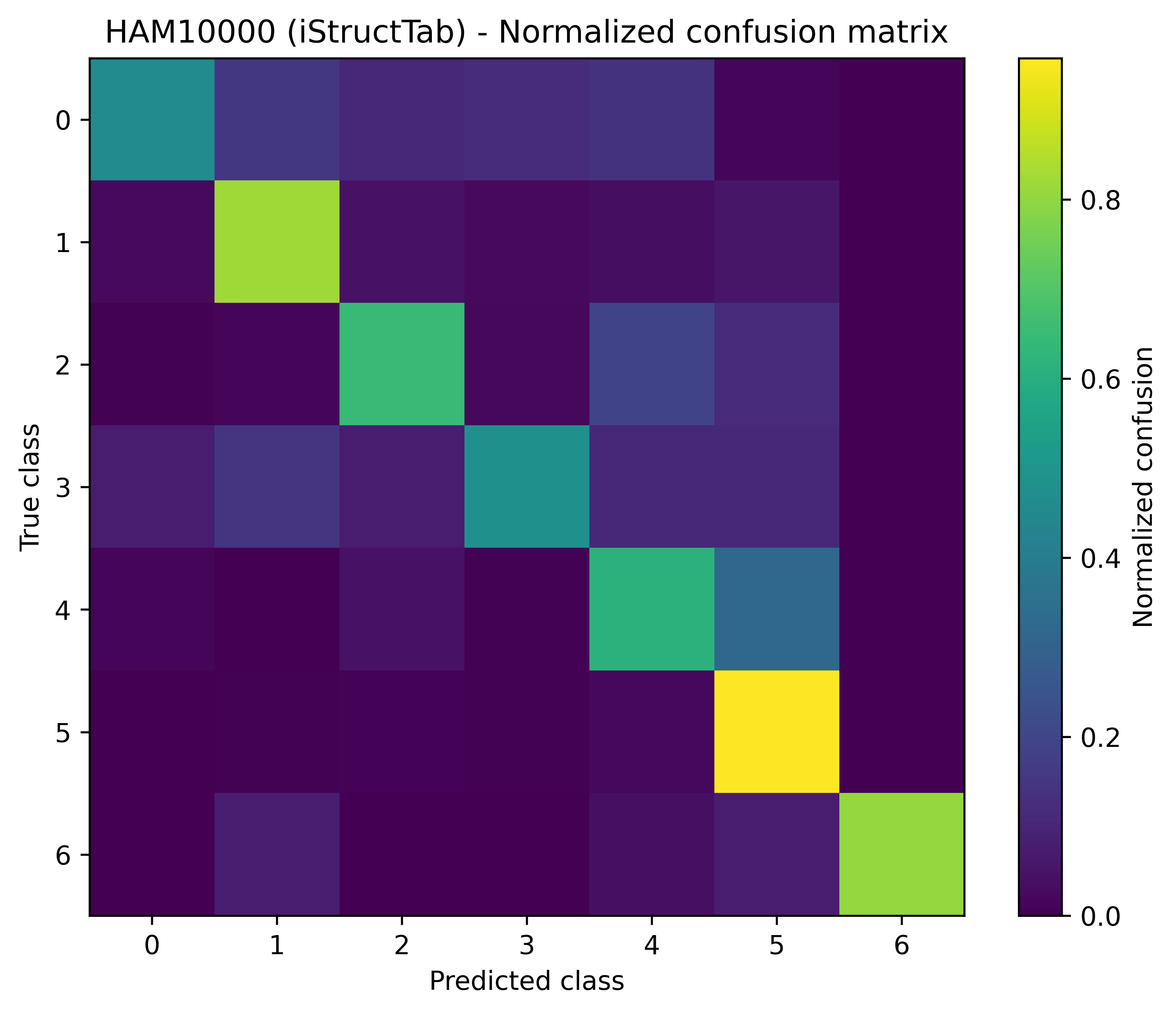}
  }\hfill
  \subfloat[Top-$k$ accuracy curve\label{fig:ham_topk}]{
    \includegraphics[width=0.23\textwidth]{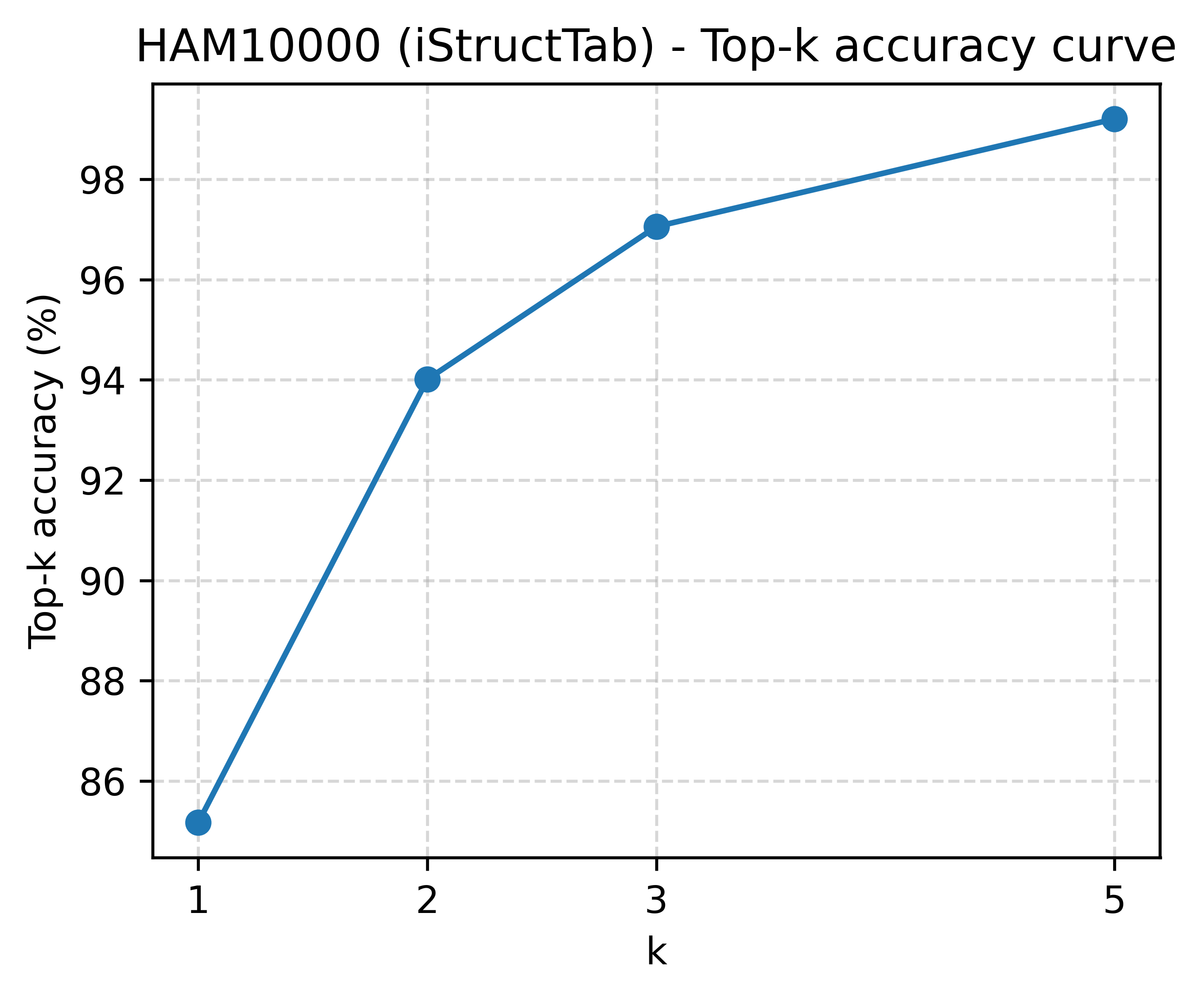}
  }\hfill
  \subfloat[Margin distribution\label{fig:ham_margin}]{
    \includegraphics[width=0.23\textwidth]{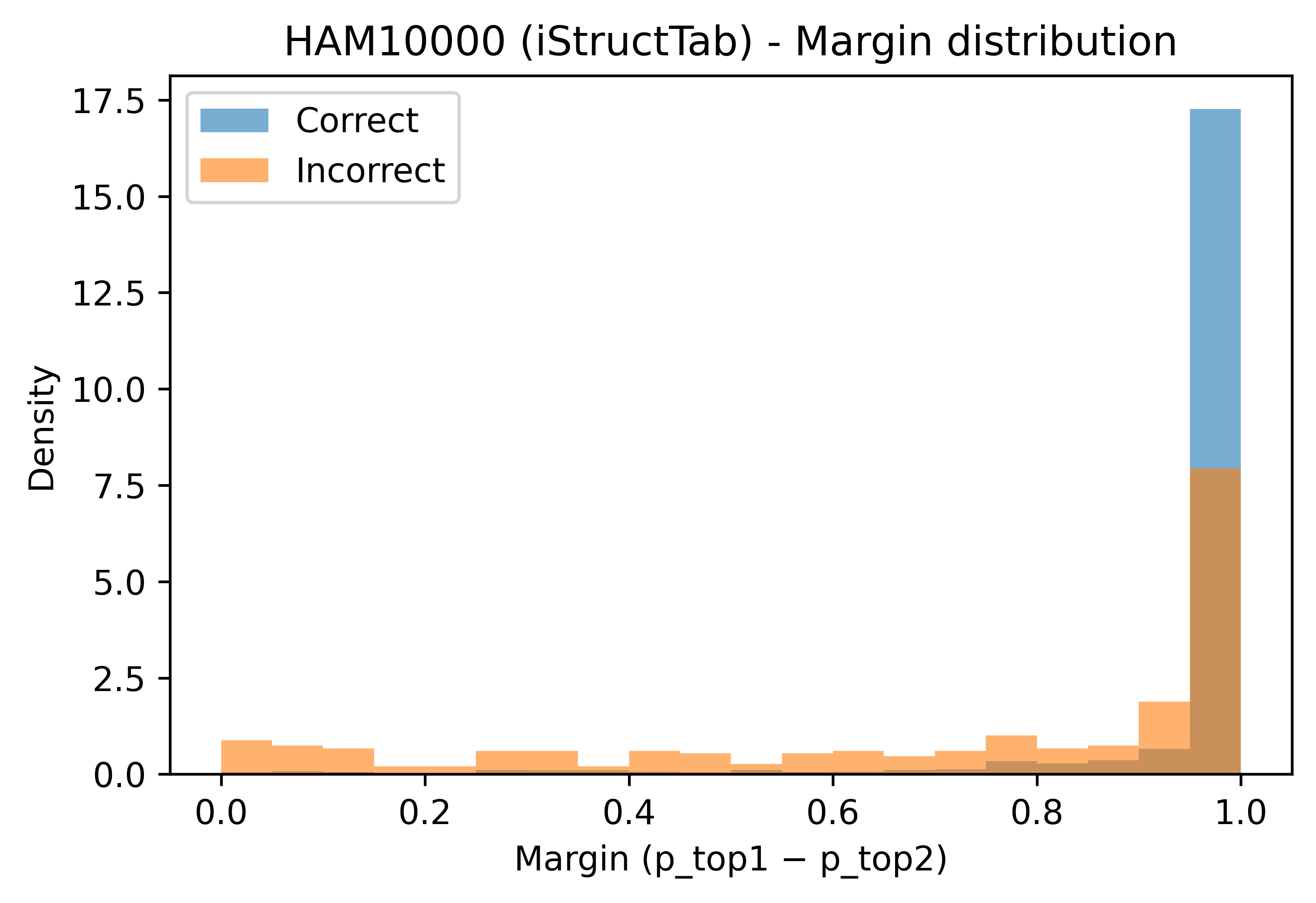}
  }\hfill
  \subfloat[Tabular feature importance\label{fig:ham_tabimp}]{
    \includegraphics[width=0.23\textwidth]{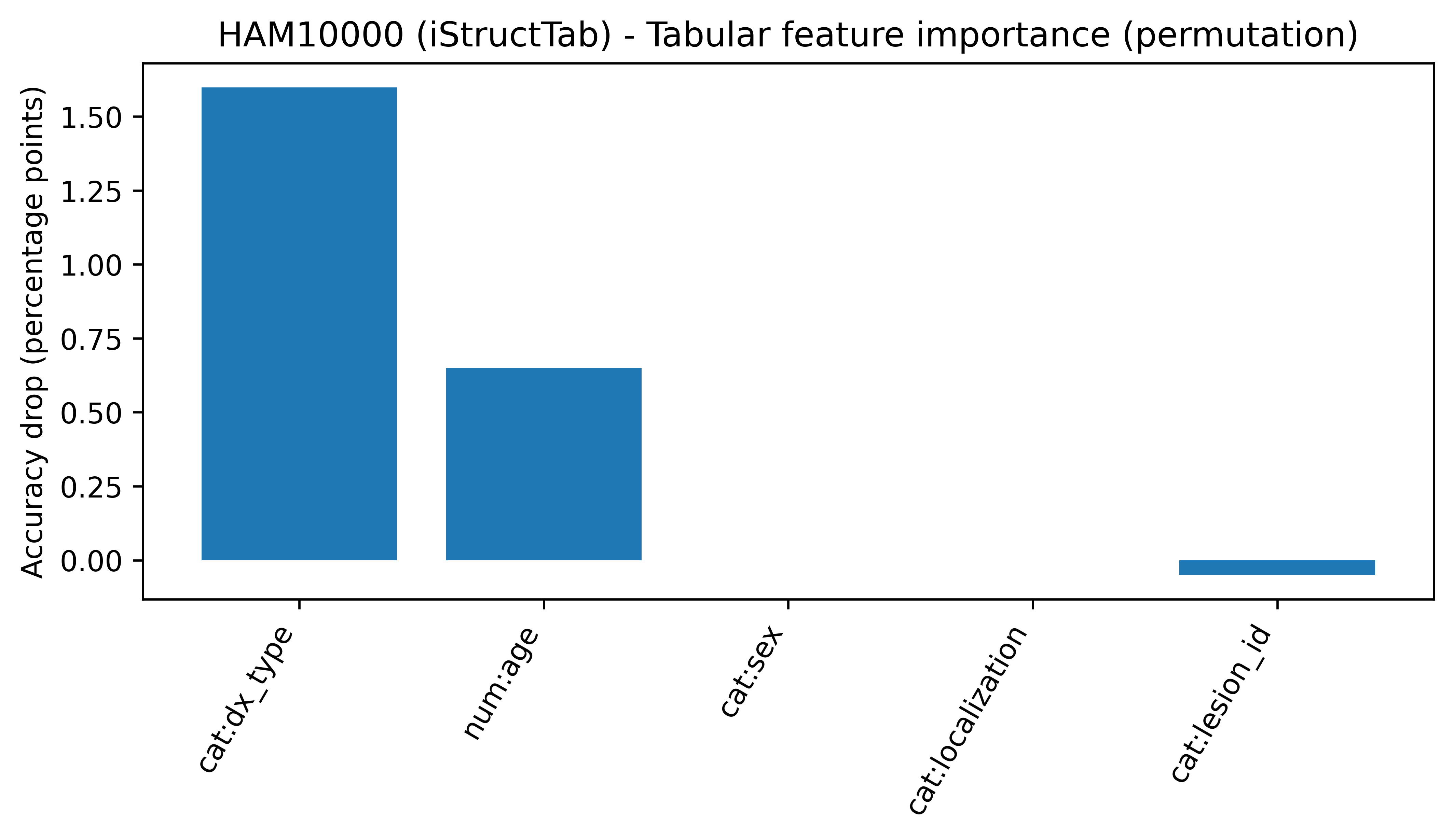}
  }
  \caption{Additional reliability and interpretability diagnostics for iStructTab on the HAM10000 dataset.
  (a) Normalized confusion matrix highlighting that errors are concentrated among a few frequent classes rather than being uniformly spread.
  (b) Top-$k$ accuracy, showing that the correct label appears in the top five predictions in $99.20\%$ of cases.
  (c) Margin distributions for correct vs.\ incorrect predictions, illustrating how confidence can be used to flag uncertain cases.
  (d) Tabular feature permutation importance, indicating that iStructTab relies on distributed clinical cues (especially diagnosis type and age) rather than a single dominant feature.}
  \label{fig:ham_extra_diag}
\end{figure*}

Finally, Figure~\ref{fig:ham_tabimp} summarizes tabular feature importance via permutation accuracy drop. Starting from a baseline accuracy of $85.17\%$, shuffling the most influential clinical feature \texttt{cat:dx\_type} reduces accuracy to $83.57\%$ (a drop of $1.60$ percentage points), while perturbing the numeric age feature yields a smaller drop to $84.52\%$ ($0.65$ points). The remaining tabular features (\texttt{cat:sex}, \texttt{cat:localization}, and \texttt{cat:lesion\_id}) have negligible marginal effect, with changes in accuracy between $-0.05$ and $0.00$ points. Combined with the strong tab-only performance observed in our modality ablations, this suggests that iStructTab relies on a distributed set of clinically meaningful cues with diagnosis type and age contributing most rather than being dominated by a single spurious covariate, which is desirable for robustness and interpretability.
\begin{table}[htbp]
  \caption{Top-$k$ accuracy on HAM10000 with iStructTab.}
  \centering
  \scriptsize
  \setlength{\tabcolsep}{4pt}
  \begin{tabular}{lcccc}
    \toprule
    $k$ & 1 & 2 & 3 & 5 \\
    \midrule
    Top-$k$ accuracy (\%) & 85.17 & 94.01 & 97.05 & 99.20 \\
    \bottomrule
  \end{tabular}
  \label{tab:ham_topk_diag}
\end{table}
\renewcommand{\thesection}{H}
\renewcommand{\thesubsection}{H.\arabic{subsection}}
\setcounter{figure}{0}\renewcommand{\thefigure}{H.\arabic{figure}}
\setcounter{table}{0}\renewcommand{\thetable}{H.\arabic{table}}
\section{Theory-Inspired Representation Diagnostics}
\label{ab11}
Beyond standard accuracy and sanity tests, we further probe the geometry, expressiveness, and confidence behavior of iStructTab on HAM using a set of theory-inspired diagnostics. These analyses examine (i) the effective dimension and spectrum of the learned multimodal embeddings, (ii) the quality of these embeddings as judged by $k$-nearest neighbor (kNN) classification, and (iii) the distribution and utility of logit margins for selective prediction. Together, they show that iStructTab learns a low-dimensional yet highly informative representation in which simple non-parametric classifiers can match or slightly exceed the trained head and in which margins provide a strong signal for when the model can be trusted.
\subsection{Embedding Geometry and Effective Dimension}
We first analyze the covariance of the frozen iStructTab embeddings on the HAM test set. After centering the embeddings, we compute the eigenvalues of the covariance matrix to obtain the variance captured by each Principal Component (PC), and derive the participation ratio effective dimension ($d_{\mathrm{eff}}$). As shown in Figs.~\ref{fig:ham_theory_diag}(a)–(b) and Table~\ref{tab:ham_pca_dim}, the spectrum is sharply but not pathologically concentrated: the first PC alone explains roughly $62\%$ of the variance, three PCs already capture about $90\%$, and six PCs account for approximately $99\%$ of the variance in a 7D embedding. The resulting effective dimension is $d_{\mathrm{eff}} \approx 2.3$, indicating that iStructTab compresses the multimodal information into a low-dimensional manifold while still preserving meaningful degrees of freedom, rather than collapsing everything onto a single direction.
\begin{table}[htbp]
  \caption{Number of principal components required to reach different target levels of explained variance for iStructTab embeddings on the HAM dataset.}
  \centering
  \scriptsize
  \setlength{\tabcolsep}{6pt}
  \begin{tabular}{lc}
    \toprule
    Target variance & Components $k$ needed \\
    \midrule
    $50\%$ & 1 \\
    $80\%$ & 3 \\
    $90\%$ & 3 \\
    $95\%$ & 4 \\
    $99\%$ & 6 \\
    \bottomrule
  \end{tabular}
  \label{tab:ham_pca_dim}
\end{table}
\subsection{kNN Probing of Representation Quality}
To quantify how informative the learned embedding is independently of the parametric classifier head, we perform a leave-one-out kNN probe in the embedding space. Using Euclidean distance on the frozen embeddings, we fit a kNN classifier and, for each test example, predict its label from the $k$ nearest neighbors excluding itself. As shown in Fig.~\ref{fig:ham_theory_diag}(c), kNN accuracy increases from $79.1\%$ at $k=1$ to $85.7\%$ at $k=20$. For moderate to large $k$, the non-parametric probe matches or slightly exceeds the parametric softmax head (top-1 test accuracy $85.17\%$), suggesting that iStructTab’s fused image–tabular embedding clusters semantically similar lesions in a way that even a simple kNN rule can exploit. This indicates that most of the predictive power lies in the representation itself, with relatively modest dependence on the specific choice of classifier head.
\subsection{Margin-Based Selective Classification}
\begin{figure*}[htbp]
  \centering
  \subfloat[Embedding spectrum (variance fraction)\label{fig:ham_spec}]{
    \includegraphics[width=0.30\linewidth]{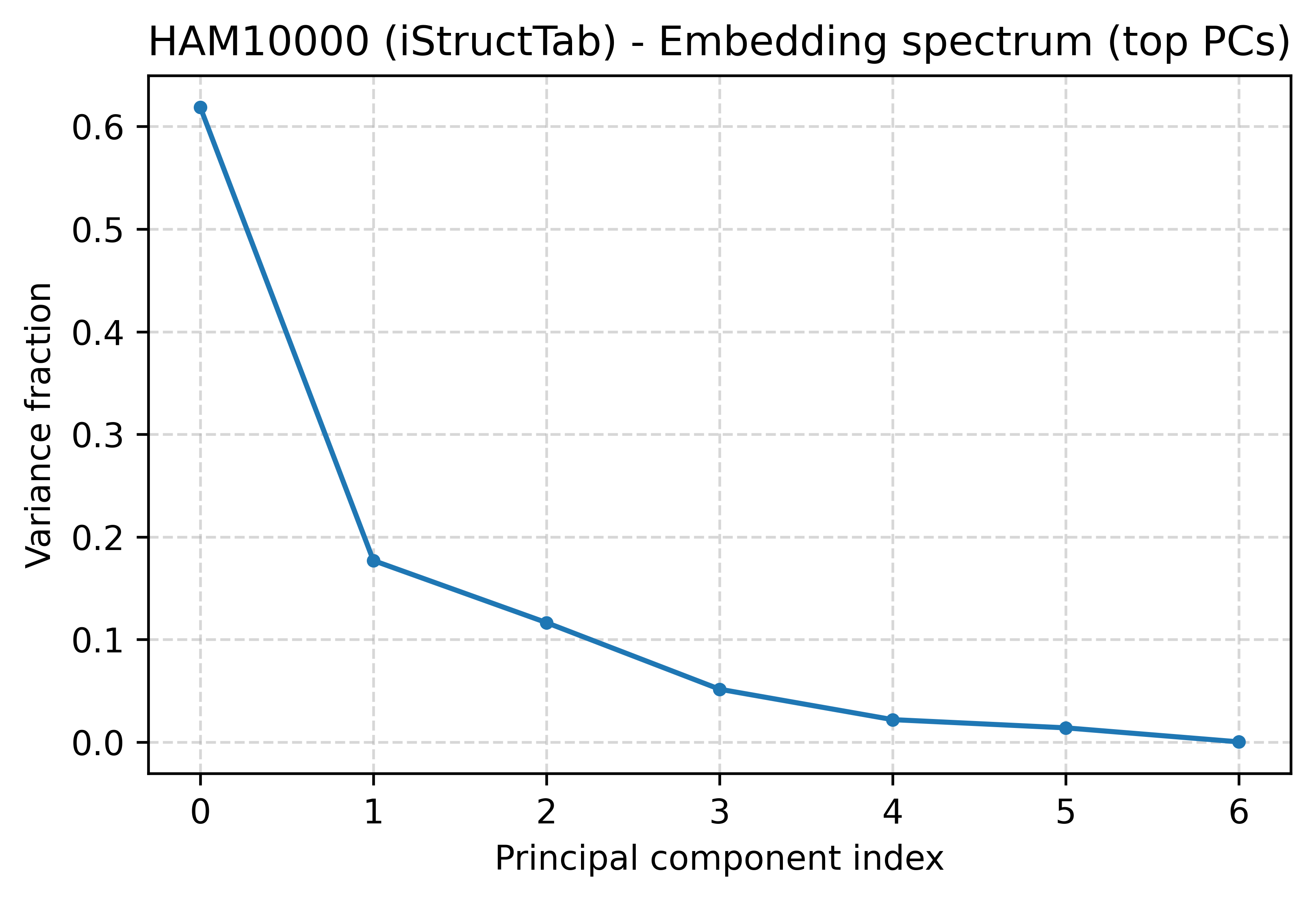}
  }\hfill
  \subfloat[Cumulative explained variance\label{fig:ham_cumvar}]{
    \includegraphics[width=0.30\linewidth]{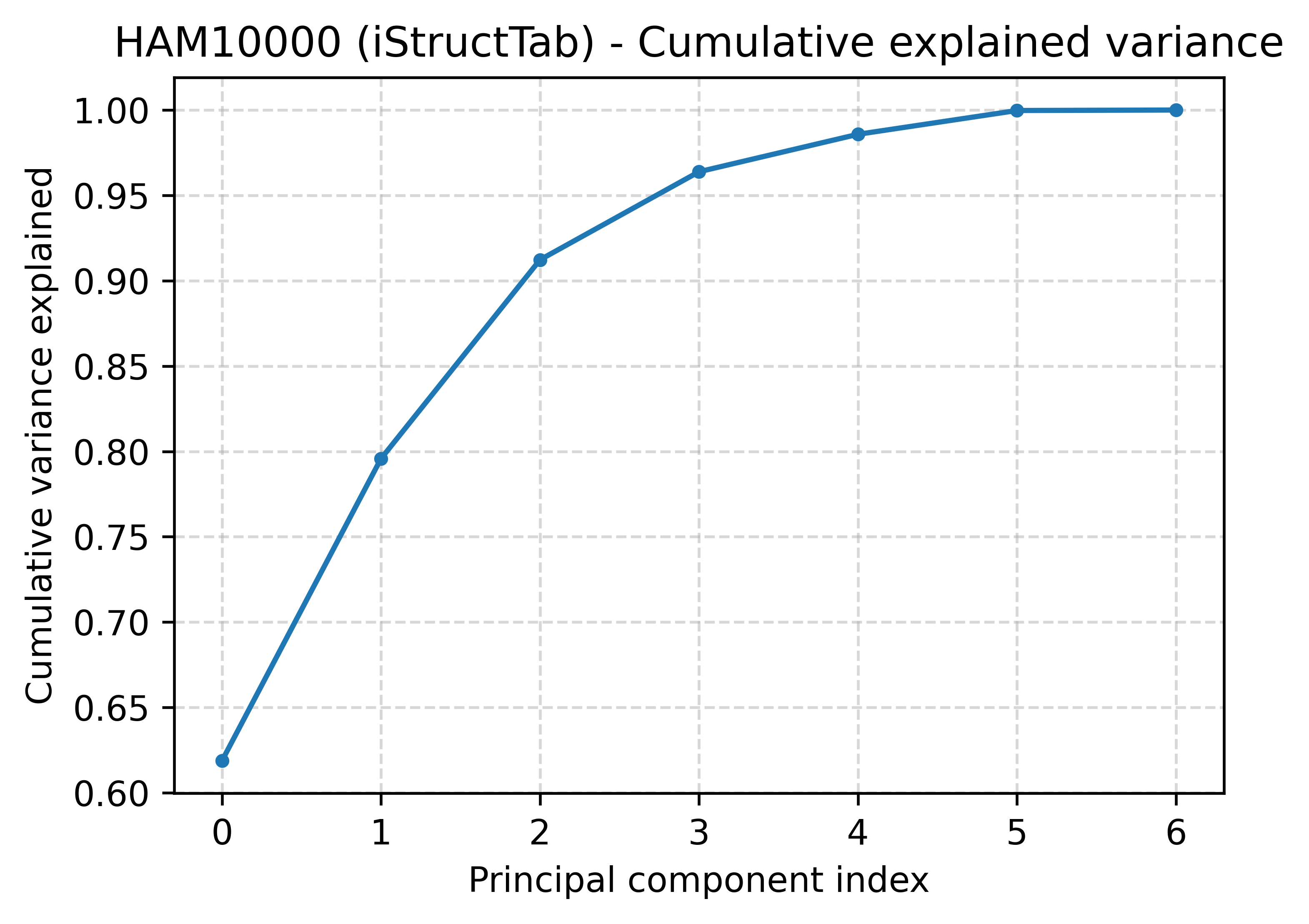}
  }\hfill
  \subfloat[kNN vs.\ parametric head\label{fig:ham_knn}]{
    \includegraphics[width=0.30\linewidth]{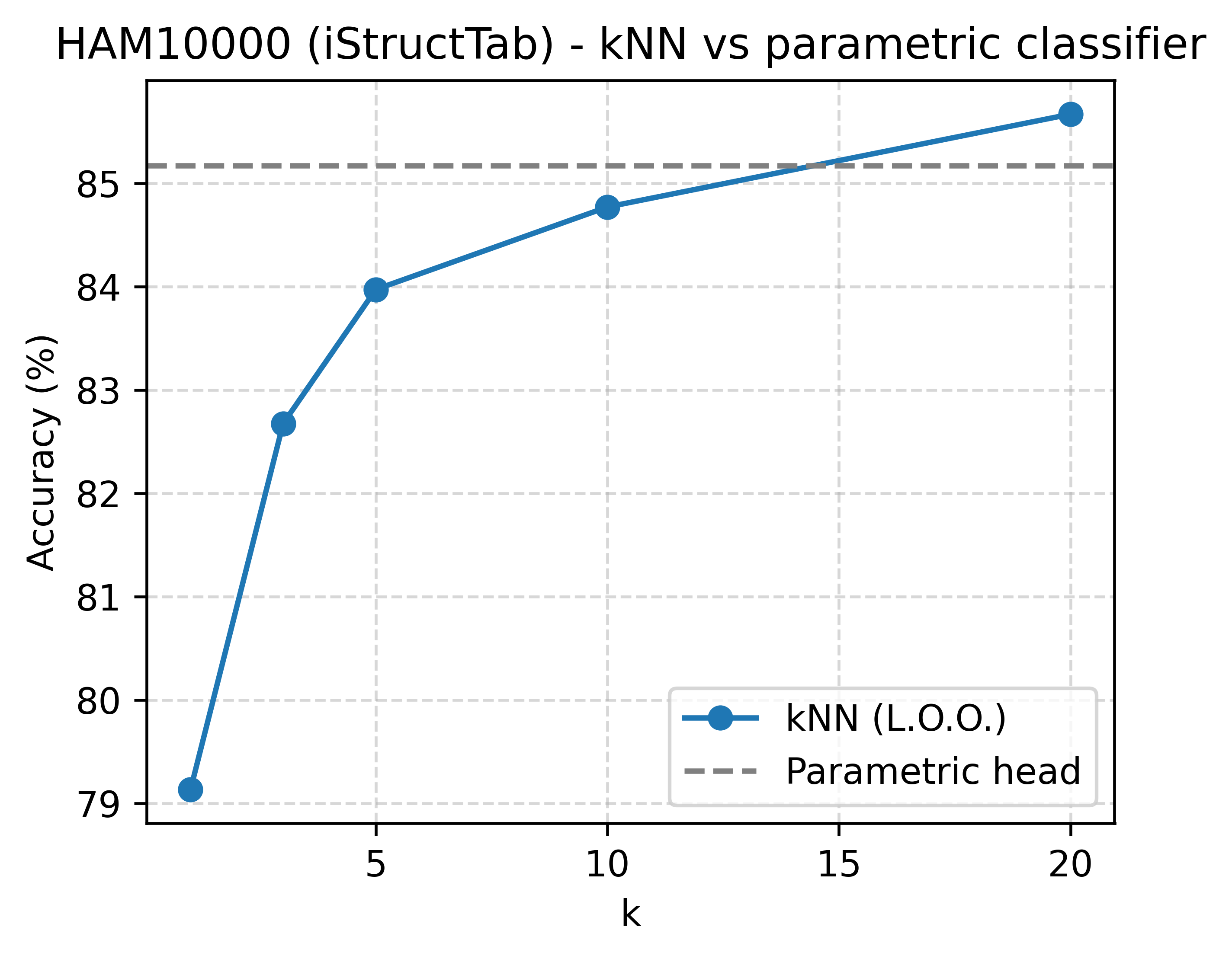}
  }\\[0.8ex]
  \subfloat[Margin-based conditional error\label{fig:ham_margin_err}]{
    \includegraphics[width=0.38\linewidth]{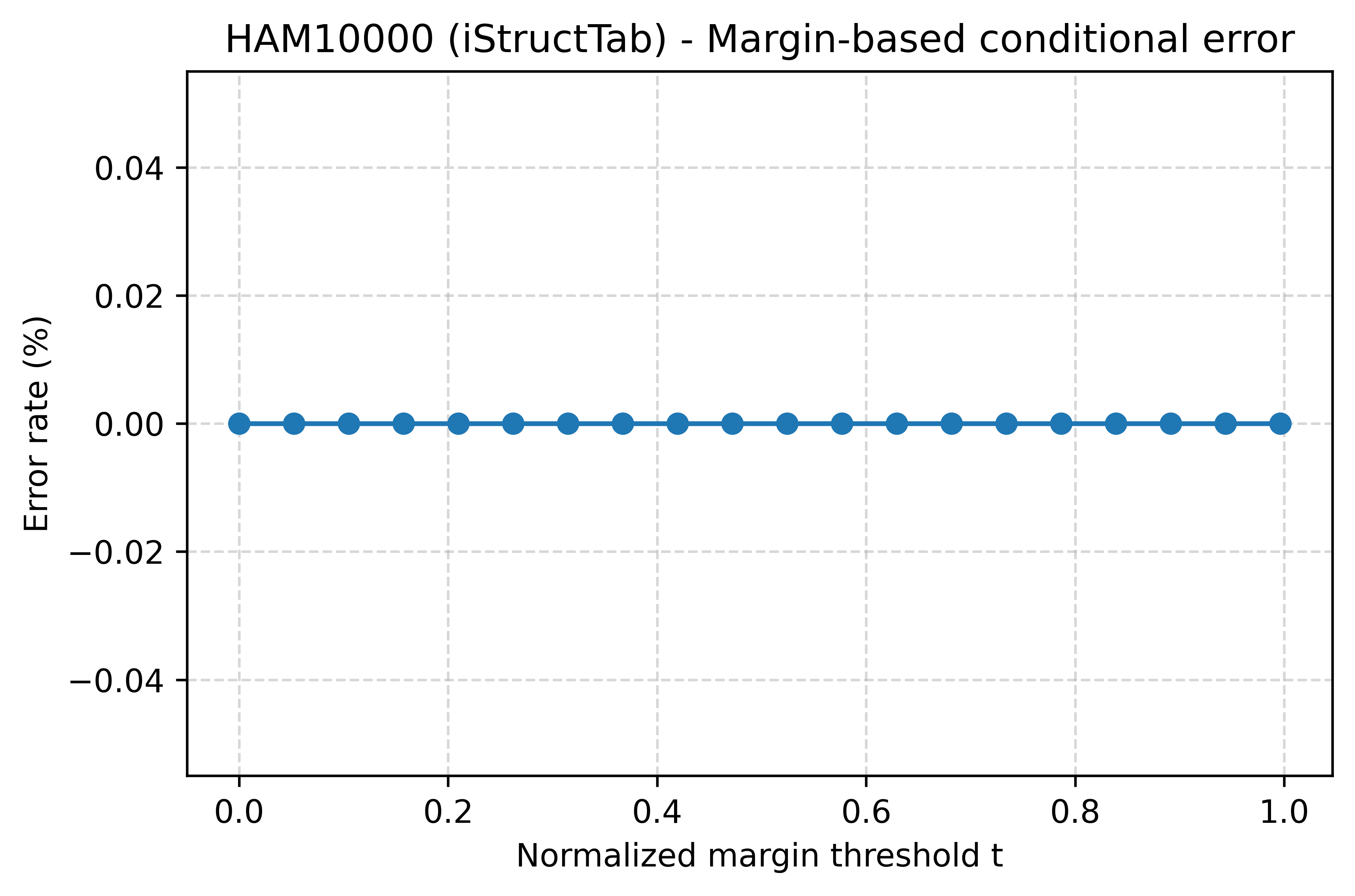}
  }\hfill
  \subfloat[Coverage vs.\ margin threshold\label{fig:ham_margin_cov}]{
    \includegraphics[width=0.38\linewidth]{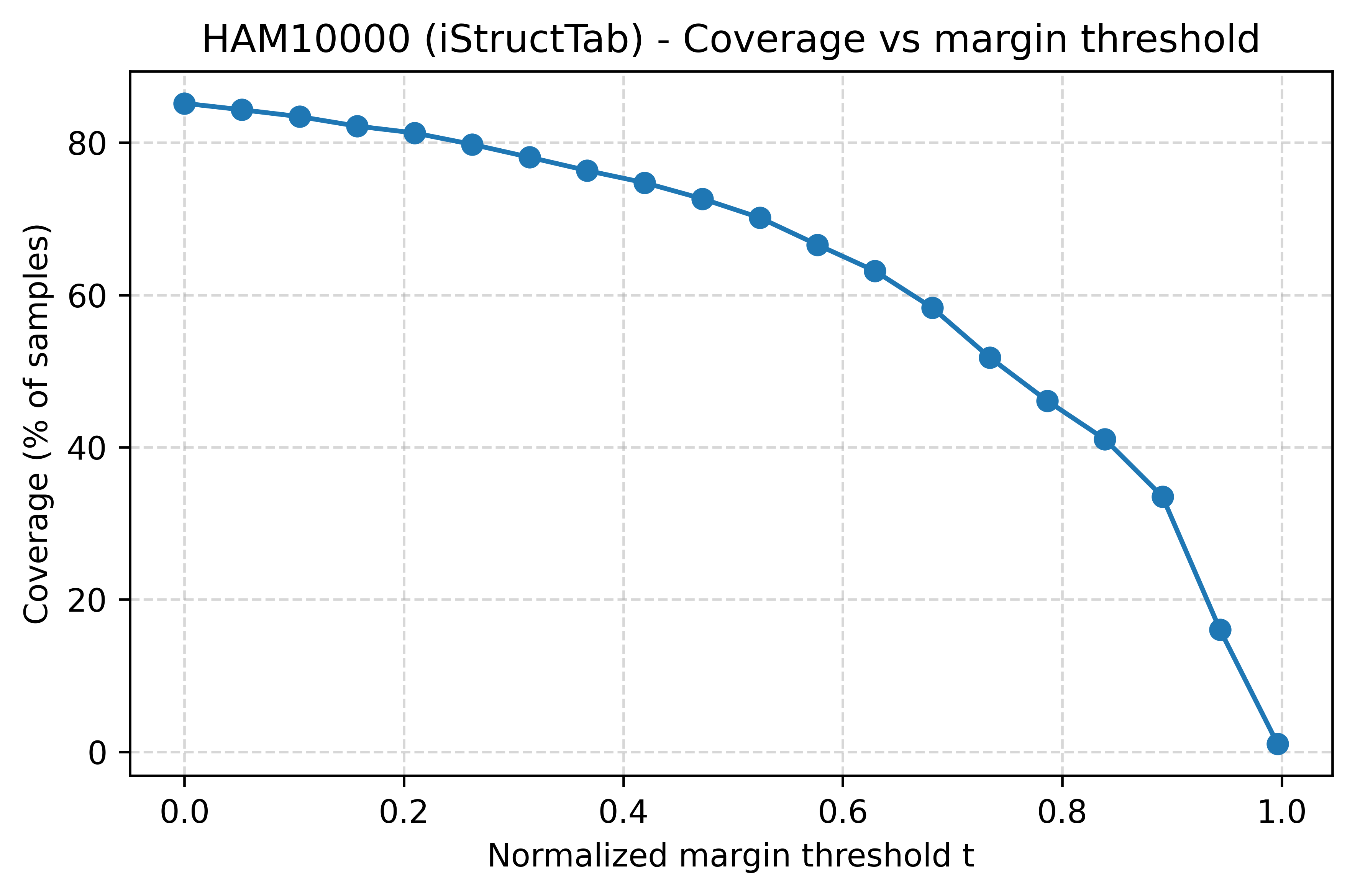}
  }
  \caption{Theory-inspired diagnostics of iStructTab on the HAM dataset.
  (a–b) The embedding spectrum is highly concentrated: a small number of
  principal components explain almost all variance, indicating a
  low-dimensional but non-collapsed representation.
  (c) Leave-one-out kNN classification in embedding space achieves up to
  $85.7\%$ accuracy, matching or slightly outperforming the parametric
  softmax head ($85.17\%$), which shows that the learned multimodal
  representation is very informative.
  (d–e) Normalized logit margins support selective prediction: as the margin
  threshold increases, coverage decreases from about $86\%$ toward zero while
  the conditional error on retained samples remains effectively $0\%$, so
  high-margin predictions from iStructTab can be treated as highly reliable.}
  \label{fig:ham_theory_diag}
\end{figure*}

Finally, we study logit margins as a confidence signal. For each test example we compute the normalized margin between the true-class logit and the highest non-true logit, rescaling by the $\ell_2$ norm of the logit vector. We then condition on the subset of samples whose normalized margin exceeds a threshold $t$ and measure both the conditional error rate and the remaining coverage. As shown in Figs.~\ref{fig:ham_theory_diag}(d)–(e), the conditional error rate on the retained subset is essentially $0\%$ across all thresholds considered, while coverage decreases smoothly from about $86\%$ of the test set at $t=0$ down to nearly $0\%$ at the most stringent thresholds. The mean normalized margin is positive for correctly classified samples ($\approx 0.74$) and negative for misclassified ones ($\approx -0.50$), indicating a clear separation between confidently correct and uncertain examples in this oracle-style margin. In a clinical workflow, this suggests that margin-based selective prediction using a deployable surrogate such as the top-1 vs.\ top-2 probability gap could be used to automatically flag high-confidence cases as reliable while deferring low-margin, potentially ambiguous cases for human review, thereby translating iStructTab’s strong internal representation into trustworthy decision support.
\renewcommand{\thesection}{I}
\renewcommand{\thesubsection}{I.\arabic{subsection}}
\setcounter{figure}{0}\renewcommand{\thefigure}{I.\arabic{figure}}
\setcounter{table}{0}\renewcommand{\thetable}{I.\arabic{table}}
\section{Turing-Style Human-Model Evaluation}
\label{ab12}
To complement purely algorithmic diagnostics, we design a small Turing-style protocol to compare iStructTab directly against human readers on held-out HAM dermoscopy images. Using the same \texttt{predict\_loader} as in the main experiments, we first run iStructTab once on the full test split to cache logits, probabilities, predictions, and ground-truth labels (\texttt{logits, probs, preds, y}). We then iterate over the test loader (with shuffling disabled), unnormalize each RGB image back to pixel space, and randomly sample $60$ indices to form a Turing subset. For each selected index we save a PNG file (\texttt{sample\_\{idx\}.png}) into a dedicated \texttt{ham\_istructtab\_turing\_test/images/} folder and record the true label, the model’s predicted label, and its confidence (maximum softmax probability) in a CSV answer sheet (\texttt{ham\_istructtab\_turing\_sheet.csv}). The CSV also contains empty fields for \texttt{human\_label\_name}, \texttt{human\_label\_id}, and \texttt{human\_notes}, which can later be filled by one or more clinicians without exposing either ground-truth labels or model predictions during the annotation phase. On this particular random subset, iStructTab attains $88.33\%$ top-1 accuracy, which is slightly higher than the global test accuracy on the full split ($85.17\%$), indicating that performance remains stable under such blind resampling (Table~\ref{tab:ham_turing_summary}). Once the sheet has been filled, a lightweight scoring script loads the CSV and computes human accuracy (based on either numeric IDs or class names), model accuracy on exactly the same subset, and detailed disagreement statistics, including which cases the human and iStructTab disagree on and at what model confidence. Although our current sheet does not yet contain human entries, the infrastructure already supports a rigorous, label-balanced comparison in future reader studies: if human accuracy is comparable to or below the reported model accuracy on the same blinded subset, this would provide evidence that iStructTab’s multimodal predictions are competitive with expert judgment on challenging real world cases. Moreover, the per-image confidence scores and the ability to isolate high-confidence disagreements give a practical path to deploying iStructTab as a decision-support tool rather than a black-box oracle: clinicians can be shown both the model’s top-$k$ suggestions and its confidence, use low-confidence cases as automatic “please review carefully” flags, and focus joint discussion on the small set of lesions where human and model systematically diverge. In this sense, the Turing-test kit is less about absolute accuracy and more about making iStructTab’s strengths and weaknesses observable, reproducible, and easy to audit in clinical settings for the future.
\begin{table}[htbp]
  \caption{Summary statistics for the Turing-style evaluation subset on
  the HAM dataset. Human accuracy will be computed on the same subset once
  \texttt{human\_label\_*} fields are filled in the answer sheet.}
  \centering
  \scriptsize
  \setlength{\tabcolsep}{6pt}
  \begin{tabular}{lc}
    \toprule
    Statistic & Value \\
    \midrule
    Subset size ($N$) & 60 \\
    Model top-1 acc.\ on subset (\%) & 88.33 \\
    Model top-1 acc.\ on full test split (\%) & 85.17 \\
    Human accuracy (\%) & to be measured \\
    \bottomrule
  \end{tabular}
  \label{tab:ham_turing_summary}
\end{table}
\renewcommand{\thesection}{J}
\renewcommand{\thesubsection}{J.\arabic{subsection}}
\setcounter{figure}{0}\renewcommand{\thefigure}{J.\arabic{figure}}
\setcounter{table}{0}\renewcommand{\thetable}{J.\arabic{table}}
\section{OOD and Local Sensitivity Diagnostics}
\label{ab13}
\begin{figure*}[htbp]
  \centering
  \subfloat[Confidence on ID vs.\ noise/blank\label{fig:ham_ood_conf}]{
    \includegraphics[width=0.32\linewidth]{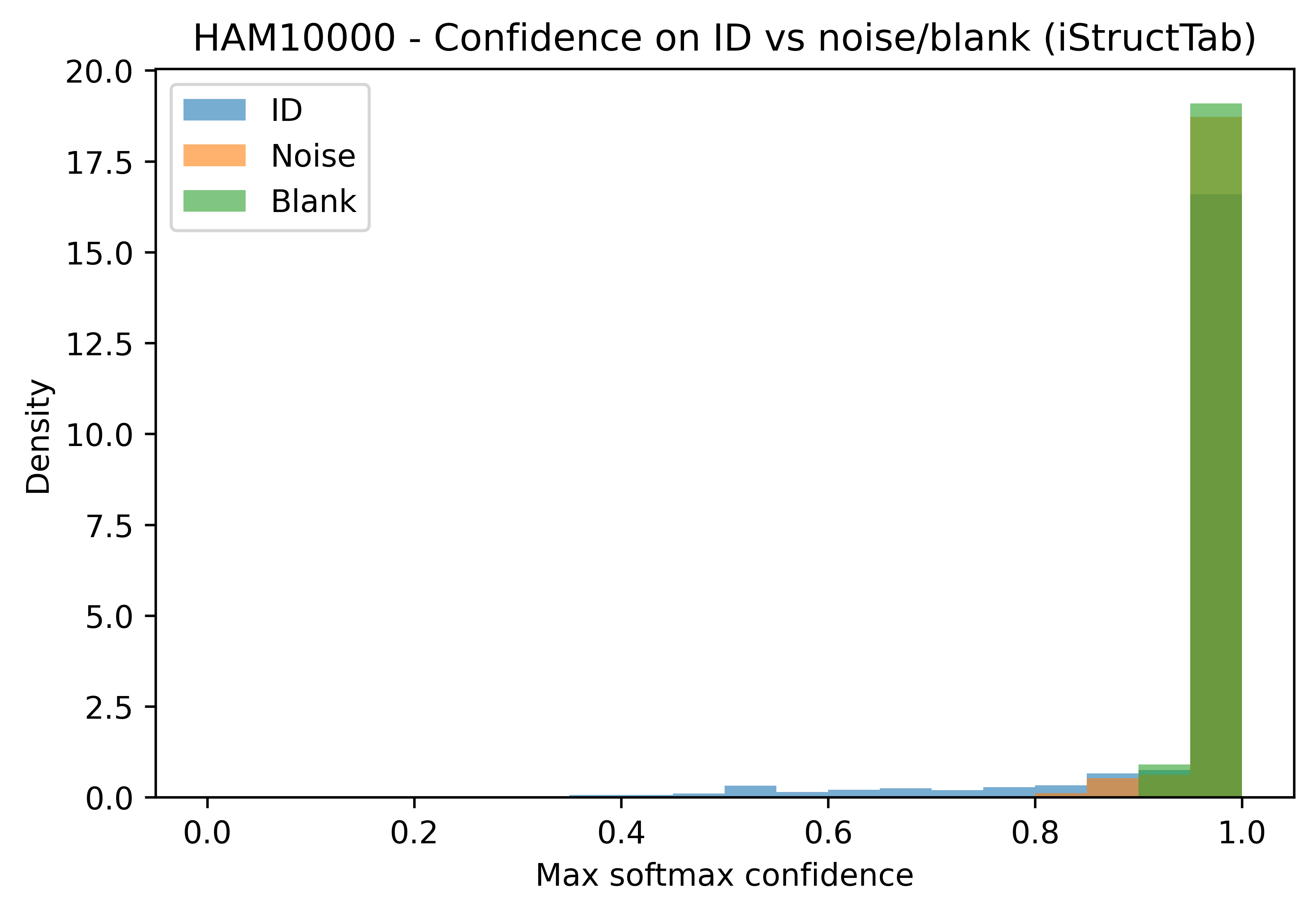}
  }\hfill
  \subfloat[Entropy on ID vs.\ noise/blank\label{fig:ham_ood_entropy}]{
    \includegraphics[width=0.32\linewidth]{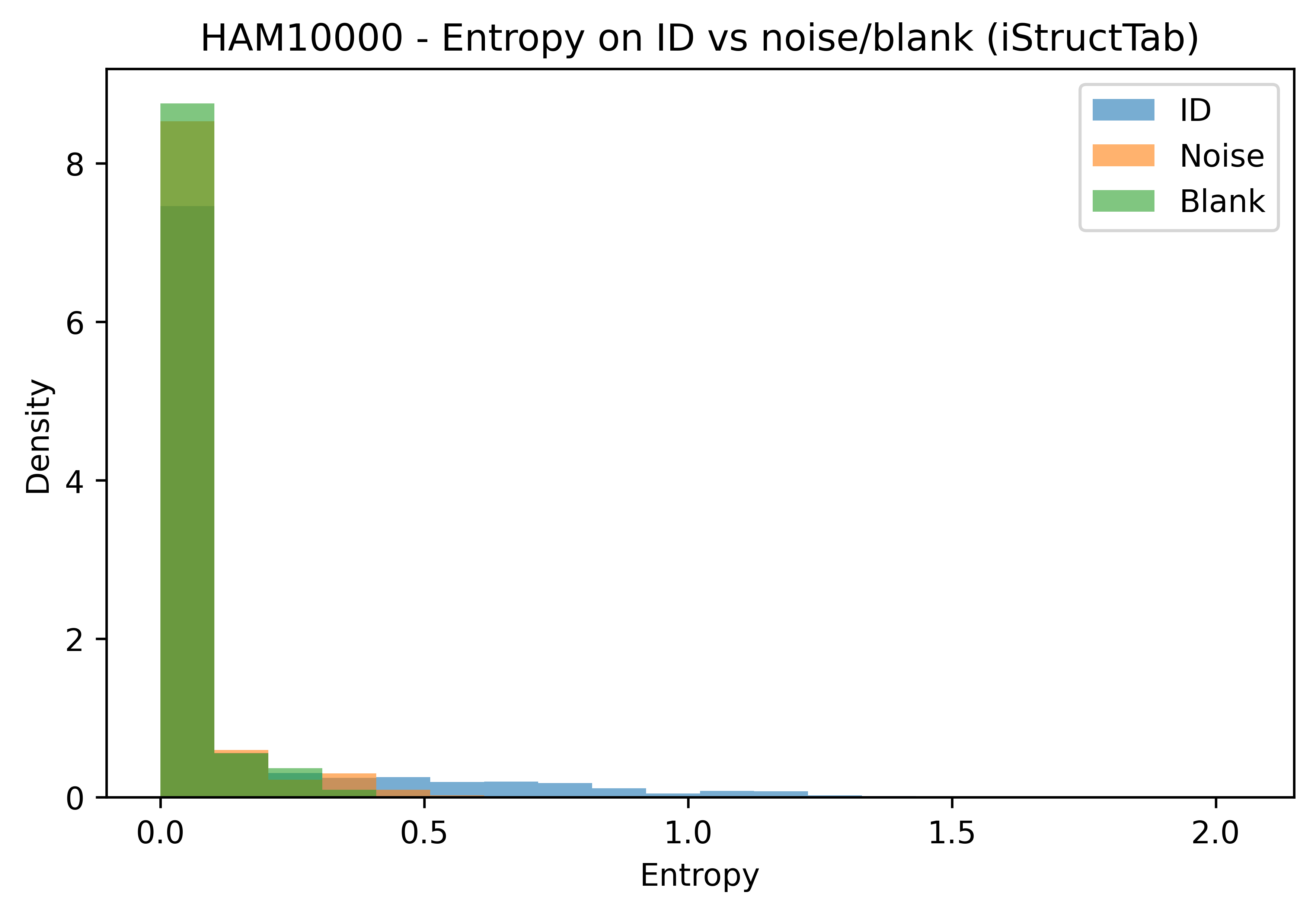}
  }\hfill
  \subfloat[Local sensitivity (image vs.\ tabular)\label{fig:ham_ood_sens}]{
    \includegraphics[width=0.32\linewidth]{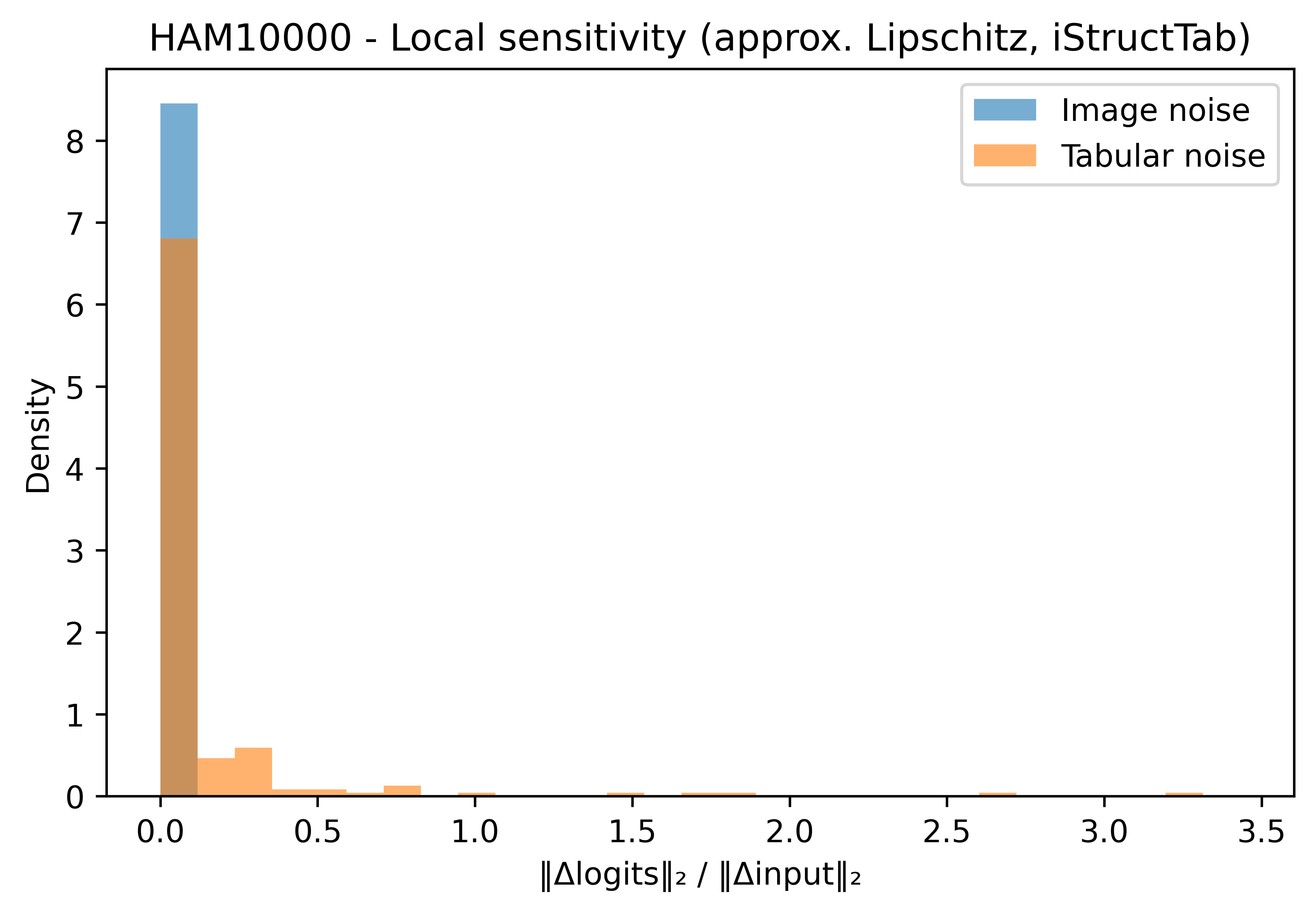}
  }
  \caption{OOD and sensitivity diagnostics for iStructTab on HAM.
  (a) Distribution of maximum softmax confidence for in-distribution (ID),
  Gaussian-noise, and blank-image inputs. Both OOD settings receive
  near-maximal confidence, illustrating softmax over-confidence on extreme
  OOD examples.
  (b) Corresponding entropy distributions, where OOD inputs concentrate near
  zero entropy, again indicating degenerate near one-hot predictions.
  (c) Approximate local Lipschitz sensitivity of the logit vector under
  small $\ell_2$-normalized perturbations to the image and tabular
  channels, showing very small sensitivity to image noise and moderate but
  bounded sensitivity to tabular perturbations.}
  \label{fig:ham_ood_sensitivity}
\end{figure*}

To assess how iStructTab behaves off-distribution and under small input perturbations, we perform a set of out-of-distribution (OOD) and local sensitivity diagnostics on the HAM dataset. First, we construct two extreme OOD settings by feeding (i) pure Gaussian noise images and (ii) blank gray images, each paired with randomly shuffled tabular features, and compare their softmax statistics with those of in-distribution (ID) test examples. As shown in Figs.~\ref{fig:ham_ood_conf} and \ref{fig:ham_ood_entropy}, ID samples already have very confident predictions (mean max-confidence $\approx 0.96$, mean entropy $\approx 0.13$). However, both Gaussian-noise and blank inputs are even more over-confident (mean max-confidence $\approx 0.99$ with entropies $\approx 0.05$ and $\approx 0.04$, respectively), corresponding to near one-hot distributions despite being clearly non-dermatological. This behavior highlights a well-known limitation of softmax probabilities under severe distribution shift: confidence/entropy alone are not sufficient as a reliable OOD detector, and additional mechanisms (e.g., dedicated OOD scores, temperature scaling, or explicit ``reject'' options) would be needed for deployment. Second, we estimate local Lipschitz-like sensitivity by injecting small $\ell_2$-normalized perturbations into either the image channel or the numeric tabular channel and measuring the induced change in the logit vector. For image perturbations with $\epsilon=0.01$, the resulting ratios $\lVert \Delta \text{logits} \rVert_2 / \lVert \Delta \text{input} \rVert_2$ are numerically close to zero (mean and median $\approx 0$), whereas tabular perturbations with $\epsilon=0.10$ yield a higher but still moderate mean ratio of $\approx 0.13$ (median $\approx 0.02$), as summarized in Fig.~\ref{fig:ham_ood_sens}. Combined with our earlier modality ablations, this suggests that iStructTab is locally very smooth with respect to the image stream while being more sensitive to clinically meaningful tabular changes a desirable property if the tabular channel is intended to capture sharp shifts in patient-level covariates. Overall, these diagnostics indicate that iStructTab behaves in a locally stable manner around real HAM inputs but, like many deep models, can still be over-confident on extreme OOD images, underscoring the importance of pairing it with explicit OOD and confidence-calibration strategies in safety-critical workflows.
\begin{table}[htbp]
  \caption{Summary of OOD confidence/entropy and local sensitivity on HAM with iStructTab.}
  \centering
  \scriptsize
  \setlength{\tabcolsep}{6pt}
  \begin{tabular}{lccc}
    \toprule
    Setting & Mean max-conf. & Mean entropy & Sensitivity ratio \\
    \midrule
    ID (test images)      & 0.955 & 0.127 & -- \\
    Gaussian-noise images & 0.989 & 0.052 & -- \\
    Blank gray images     & 0.992 & 0.042 & -- \\
    \midrule
    Image noise ($\epsilon=0.01$) & -- & -- & $\approx 0.000$ \\
    Tabular noise ($\epsilon=0.10$) & -- & -- & $\approx 0.134$ \\
    \bottomrule
  \end{tabular}
  \label{tab:ham_ood_summary}
\end{table}

\renewcommand{\thesection}{K}
\renewcommand{\thesubsection}{K.\arabic{subsection}}
\setcounter{figure}{0}\renewcommand{\thefigure}{K.\arabic{figure}}
\setcounter{table}{0}\renewcommand{\thetable}{K.\arabic{table}}
\section{Deployment-Oriented Triage Diagnostics}
\label{ab14}
To get a first glimpse of how iStructTab might behave in a clinical triage role, we collapse the seven-way HAM labels into a binary ``malignant versus benign'' task and study the ROC curve induced by the predicted malignant probability (sum over malignant classes). As shown in Fig.~\ref{fig:ham_triage_roc}, iStructTab achieves a strong AUC of $0.936$ on this binary task. More importantly, the ROC admits a high-sensitivity operating point: at threshold $\mathrm{th}^\ast \approx 0.003$, the model reaches $95.0\%$ sensitivity while maintaining $70.5\%$ specificity, missing only $19$ of $383$ malignant cases on the test split and correctly identifying $1142$ of $1620$ benign cases. This corresponds to an overall accuracy of $75.2\%$ and a precision of $43.2\%$ (Table~\ref{tab:ham_triage_stats}), which is appropriate for a recall-focused ``safety-net'' triage setting where it is acceptable to generate additional false positives in order to flag nearly all potentially malignant lesions for expert review. We also compute triage metrics across coarse demographic slices. Sensitivity remains high for both females ($92.1\%$) and males ($96.5\%$), with moderate differences in specificity (approximately $72.9\%$ vs.\ $68.5\%$), while very small ``unknown'' groups are too underpowered for interpretation. Age-sliced analysis shows a similar pattern: younger cohorts (\(<40\) and $40$-$59$) achieve higher overall accuracy and specificity, whereas older patients ($60$-$79$, $80+$) see extremely high sensitivity ($\approx 97\%$) at the cost of lower specificity, reflecting a conservative, ``better safe than sorry'' behavior on higher-risk groups. Finally, we profile inference latency over 20 mini-batches, obtaining a mean of $\approx 43.6$ ms per batch (p90 $\approx 43.6$ ms), indicating that iStructTab can support near real-time screening or pre-clinic triage on commodity lab hardware. While these deployment-style diagnostics are limited by using a single dataset and do not constitute a full clinical validation or fairness audit, they suggest that iStructTab can already provide useful, recall-oriented decision support, and motivate future work on calibration, fairness, and multi-center evaluation.
\renewcommand{\thesection}{B}
\renewcommand{\thesubsection}{B.\arabic{subsection}}
\setcounter{figure}{0}\renewcommand{\thefigure}{B.\arabic{figure}}
\setcounter{table}{0}\renewcommand{\thetable}{B.\arabic{table}}
\begin{table}[htbp]
\caption{Binary ``malignant versus benign'' triage metrics for iStructTab on
  HAM dataset at the chosen operating point $\mathrm{th}^\ast$.}
  \centering
  \scriptsize
  \setlength{\tabcolsep}{4pt}
  \begin{tabular}{lcccc}
    \toprule
    Setting & Sens.\ (\%) & Spec.\ (\%) & Prec.\ (\%) & Acc.\ (\%) \\
    \midrule
    $\mathrm{th}^\ast$ (sens.\ $\approx 95\%$) & 95.04 & 70.54 & 43.23 & 75.19 \\
    \bottomrule
  \end{tabular}
  \label{tab:ham_triage_stats}
\end{table}
\begin{figure}[htbp]
  \centering
  \includegraphics[width=0.75\linewidth]{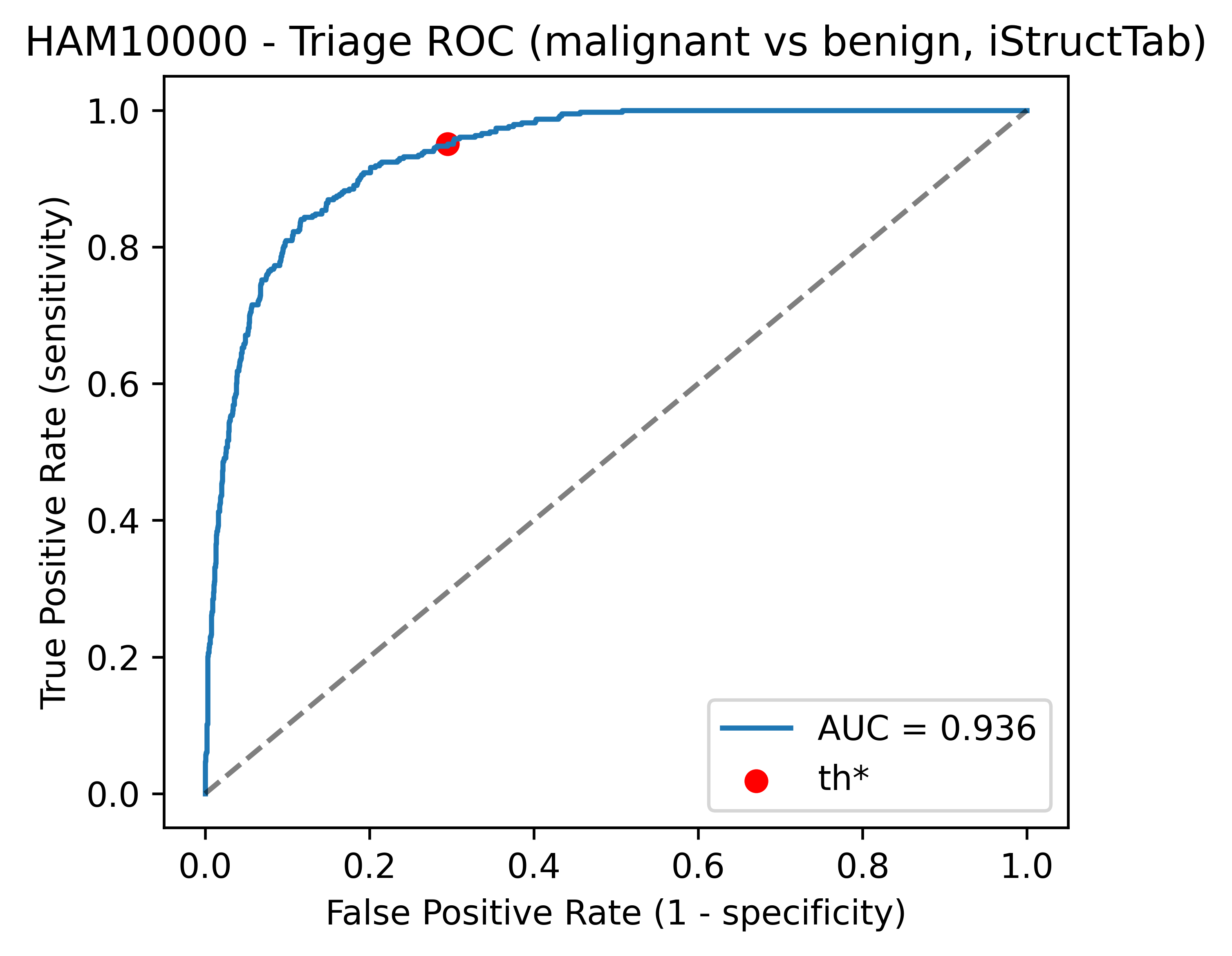}
  \caption{Binary triage ROC curve for iStructTab on HAM dataset (malignant
  vs.\ benign). The model attains an AUC of $0.936$. The red marker denotes
  the chosen operating point $\mathrm{th}^\ast$, which achieves high
  sensitivity ($\approx 95\%$) with moderate specificity, suitable for a
  recall-focused screening or triage assistant.}
  \label{fig:ham_triage_roc}
\end{figure}

\begin{table}[htbp]
\caption{Hyperparameters for the Interact-Fuse multimodal image-tabular model used in our experiments.}
  \centering
  \scriptsize
  \setlength{\tabcolsep}{6pt}
  \begin{tabular}{ll}
    \toprule
    Hyperparameter & Value \\
    \midrule
    Backbone CNN & ResNet-50 (ImageNet-1K pretrained) \\
    Fusion stage & \texttt{layer3} feature map ($C = 1024$) \\
    Tabular input dim & $d_{\text{tab}} = n_{\text{tab\_features}}$ \\
    Tabular MLP hidden sizes & (128, 64) \\
    Tabular fusion type & Multiplicative gate (\texttt{film=False}) \\
    Tabular dropout & 0.0 \\
    Classifier head & 2048 $\rightarrow$ 512 $\rightarrow n_{\text{classes}}$ \\
    Classifier head dropout & 0.2 (after 512-d layer) \\
    Image resolution & $224 \times 224$ \\
    Batch size & 64 \\
    Epochs (max) & 12 (early stopping, patience = 3) \\
    Optimizer & AdamW (lr = $10^{-3}$, weight decay = $10^{-4}$) \\
    Train/val/test split & 64\% / 16\% / 20\% \\
    Random seed & 42 \\
    \bottomrule
  \end{tabular}
  \label{tab:interact_fuse_params}
\end{table}
\begin{table}[htbp]
\centering
\caption{Hyperparameters and training setup for the multimodal DAFT baseline.}
\label{tab:multimodal_daft}
\scriptsize
\setlength{\tabcolsep}{4pt}
\renewcommand{\arraystretch}{0.95}
\begin{tabular}{ll@{\hspace{1.2em}}ll}
\hline
\textbf{Hyperparameter} & \textbf{Value} &
\textbf{Hyperparameter} & \textbf{Value} \\
\hline
model\_backbone & ResNet-50 &
one\_hot & True \\
batch\_size & 128 &
eval\_one\_hot & True \\
lr & 3e{-4} &
replace\_special\_rate & 0.20 \\
lr\_eval & 3e{-4} &
replace\_random\_rate & 0.10 \\
lr\_classifier & 3e{-4} &
encoder\_num\_layers & 2 \\
weight\_decay & 1.5e{-6} &
projector\_num\_layers & 1 \\
weight\_decay\_eval & 0 &
tabular\_embedding\_dim & 512 \\
weight\_decay\_classifier & 1e{-4} &
tabular\_transformer\_layers & 4 \\
max\_epochs & 50 &
multimodal\_transformer\_layers & 4 \\
warmup\_epochs & 10 &
multimodal\_embedding\_dim & 512 \\
anneal\_max\_epochs & 200 &
drop\_rate & 0.0 \\
projection\_dim & 128 &
augmentation\_rate & 0.95 \\
temperature & 0.1 &
crop\_scale\_lower & 0.08 \\
momentum & 0.99 &
eval\_train\_augment\_rate & 0.8 \\
lambda\_0 & 0.5 &
seed & 2022 \\
corruption\_rate & 0.3 &
 & \\
\hline
\end{tabular}
\end{table}
\begin{table}[htbp]
\centering
\caption{Hyperparameters and training setup for the multimodal MMCL baseline.}
\label{tab:multimodal_mmcl}
\scriptsize
\setlength{\tabcolsep}{4pt}
\renewcommand{\arraystretch}{0.95}
\begin{tabular}{ll@{\hspace{1.2em}}ll}
\hline
\textbf{Hyperparameter} & \textbf{Value} &
\textbf{Hyperparameter} & \textbf{Value} \\
\hline
model\_backbone & ResNet-50 &
lambda\_0 & 0.5 \\
batch\_size & 64 &
corruption\_rate & 0.3 \\
lr & 3e{-4} &
encoder\_num\_layers & 2 \\
lr\_eval & 1e{-4} &
projector\_num\_layers & 1 \\
lr\_classifier & 3e{-4} &
tabular\_embedding\_dim & 2048 \\
weight\_decay & 1.5e{-6} &
tabular\_transformer\_layers & 4 \\
weight\_decay\_eval & 0 &
augmentation\_rate & 0.95 \\
weight\_decay\_classifier & 1e{-4} &
crop\_scale\_lower & 0.08 \\
max\_epochs & 100 &
eval\_train\_augment\_rate & 0.8 \\
warmup\_epochs & 10 &
seed & 2022 \\
anneal\_max\_epochs & 20 &
 & \\
projection\_dim & 128 &
 & \\
temperature & 0.1 &
 & \\
momentum & 0.99 &
 & \\
\hline
\end{tabular}
\end{table}
\begin{table*}[t]
\caption{Baselines and corresponding source implementations used in our experiments.}
\centering
\scriptsize
\begin{tabular}{ll}
\toprule
\textbf{Baseline} & \textbf{Source / implementation} \\
\midrule
CatBoost        & \url{https://github.com/catboost/catboost} \\
LightGBM (LGBM) & \url{https://github.com/microsoft/LightGBM} \\
TabSeq          & \url{https://github.com/zadid6pretam/TabSeq} \\
TabM            & \url{https://github.com/OpenTabular/DeepTab/blob/master/deeptab/models/tabm.py} \\
SAINT           & \url{https://github.com/OpenTabular/DeepTab/blob/master/deeptab/models/saint.py} \\
SCARF           & \url{https://github.com/clabrugere/pytorch-scarf} \\
ResNet-50       & \texttt{torchvision.models} (standard PyTorch) \\
ViT             & \texttt{torchvision.models} (standard PyTorch) \\
SimCLR          & \url{https://github.com/sthalles/SimCLR} \\
BYOL            & \url{https://github.com/lucidrains/byol-pytorch} \\
DAFT            & \url{https://github.com/ai-med/DAFT} \\
Interact-Fuse   & Custom PyTorch implementation (for this work) \\
MMCL            & \url{https://github.com/paulhager/MMCL-Tabular-Imaging} \\
TIP             & \url{https://github.com/siyi-wind/TIP} \\
STiL            & \url{https://github.com/siyi-wind/STiL} \\
iStructTab (ours) & \url{https://github.com/annonym414/multimodal-tabular-anon}\\
\bottomrule
\end{tabular}
\label{tab:baseline_sources}
\end{table*}
\begin{table}[htbp]
\centering
\caption{Hyperparameters and training setup for the multimodal TIP baseline.}
\label{tab:multimodal_tip}
\scriptsize
\setlength{\tabcolsep}{4pt}
\renewcommand{\arraystretch}{0.95}
\begin{tabular}{ll@{\hspace{1.2em}}ll}
\hline
\textbf{Hyperparameter} & \textbf{Value} &
\textbf{Hyperparameter} & \textbf{Value} \\
\hline
model\_backbone & ResNet-50 &
corruption\_rate & 0.3 \\
batch\_size & 64 &
replace\_special\_rate & 0.5 \\
lr & 3e{-4} &
replace\_random\_rate & 0.0 \\
weight\_decay & 1.5e{-6} &
tabular\_embedding\_dim & 512 \\
max\_epochs & 100 &
tabular\_transformer\_layers & 4 \\
warmup\_epochs & 10 &
multimodal\_transformer\_layers & 4 \\
anneal\_max\_epochs & 20 &
multimodal\_embedding\_dim & 512 \\
projection\_dim & 128 &
augmentation\_rate & 0.95 \\
temperature & 0.1 &
crop\_scale\_lower & 0.08 \\
momentum & 0.99 &
seed & 2022 \\
lambda\_0 & 0.5 &
 &  \\
\hline
\end{tabular}
\end{table}
\begin{table}[htbp]
\centering
\caption{Hyperparameters and training setup for the multimodal STiL baseline.}
\label{tab:multimodal_stil}
\scriptsize
\setlength{\tabcolsep}{4pt}
\renewcommand{\arraystretch}{0.95}
\begin{tabular}{ll@{\hspace{1.2em}}ll}
\hline
\textbf{Hyperparameter} & \textbf{Value} &
\textbf{Hyperparameter} & \textbf{Value} \\
\hline
model\_backbone & ResNet-50 &
multimodal\_transformer\_layers & 1 \\
batch\_size & 64 &
multimodal\_embedding\_dim & 512 \\
lr & 3e{-4} &
drop\_rate & 0.0 \\
lr\_eval & 1e{-4} &
augmentation\_rate & 0.95 \\
weight\_decay & 1.5e{-6} &
crop\_scale\_lower & 0.08 \\
weight\_decay\_eval & 0 &
alpha & 0.2 \\
max\_epochs & 100 &
beta & 3.0 \\
warmup\_epochs & 10 &
gamma & 0.5 \\
anneal\_max\_epochs & 20 &
rate\_pt & 1.0 \\
projection\_dim & 128 &
rate\_uce & 0.2 \\
temperature & 0.1 &
unlabelled\_ratio & 7 \\
momentum & 0.99 &
th1 & 0.90 \\
lambda\_0 & 0.5 &
th2 & 0.95 \\
corruption\_rate & 0.3 &
th\_contrast & 0.8 \\
encoder\_num\_layers & 2 &
start\_epoch & 35 \\
projector\_num\_layers & 1 &
rate\_pseudo & 0.9 \\
tabular\_embedding\_dim & 512 &
ema\_momentum & 0.996 \\
tabular\_transformer\_layers & 4 &
sharpen\_temperature & 0.1 \\
 &  &
seed & 2022 \\
\hline
\end{tabular}
\end{table}